\documentclass[letterpaper, 10 pt, journal, twoside]{IEEEtran} 

\usepackage{tcolorbox}
\usepackage[ruled,linesnumbered]{algorithm2e}
\usepackage{bm}
\usepackage{stackrel}
\usepackage{subcaption}
\usepackage{epsf}
\usepackage{amsmath,amssymb}
\usepackage{graphicx}
\usepackage{color}
\usepackage{cite}
\usepackage{multirow,tabularx}
\usepackage{ifthen} 
\usepackage{epstopdf}
\usepackage{mathtools}
\usepackage{float}
\usepackage{times}
\usepackage{tikz}

\usepackage{booktabs} %
\usepackage{capt-of}
\usepackage{microtype}
\usepackage{hyperref}
\usepackage{xr}
\usepackage{amsthm}
\usepackage{soul}
\usepackage{enumerate}
\usepackage{lipsum}
\usepackage{wrapfig}

\usepackage[capitalise]{cleveref}
\newtheorem{theorem}{Theorem}
\newtheorem{problem}{Problem}

\newtheorem{lemma}[theorem]{Lemma}
\newtheorem{assumption}[theorem]{Assumption}
\newtheorem{definition}[theorem]{Definition}

\newtheorem{remark}[theorem]{Remark}

\newcommand{\cf}{\emph{cf.}\xspace}

\newcommand{\bdmath}{\begin{dmath}}
\newcommand{\edmath}{\end{dmath}}
\newcommand{\beq}{\begin{equation}}
\newcommand{\eeq}{\end{equation}}
\newcommand{\bdm}{\begin{displaymath}}
\newcommand{\edm}{\end{displaymath}}
\newcommand{\bea}{\begin{eqnarray}}
\newcommand{\eea}{\end{eqnarray}}
\newcommand{\beal}{\beq \begin{array}{ll}}
\newcommand{\eeal}{\end{array} \eeq}
\newcommand{\beas}{\begin{eqnarray*}}
\newcommand{\eeas}{\end{eqnarray*}}
\newcommand{\ba}{\begin{array}}
\newcommand{\ea}{\end{array}}
\newcommand{\bit}{\begin{itemize}}
\newcommand{\eit}{\end{itemize}}
\newcommand{\ben}{\begin{enumerate}}
\newcommand{\een}{\end{enumerate}}

\newcommand{\calD}{{\cal D}}

\newcommand{\calG}{{\cal G}}
\newcommand{\calH}{{\cal H}}

\newcommand{\calL}{{\cal L}}
\newcommand{\calM}{{\cal M}}

\newcommand{\calP}{{\cal P}}

\newcommand{\calS}{{\cal S}}

\newcommand{\setG}{\textsf{G}\xspace}

\newcommand{\setK}{\textsf{K}\xspace}

\newcommand{\smallheading}[1]{\textit{#1}: }

\newcommand{\etal}{\emph{et~al.}\xspace}
\newcommand{\setal}{~\emph{et~al.}\xspace}
\newcommand{\eg}{\emph{e.g.,}\xspace}
\newcommand{\ie}{\emph{i.e.,}\xspace}
\newcommand{\myParagraph}[1]{{\bf #1.}\xspace}
\newcommand{\M}[1]{{\bm #1}} %
\renewcommand{\boldsymbol}[1]{{\bm #1}}

\newcommand{\LC}[1]{{\color{red} \textbf{LC}: #1}}

\newcommand{\hide}[1]{}
\newcommand{\wrt}{w.r.t.\xspace}

\newcommand{\hiddenText}{{\color{gray} hidden text.}}
\newcommand{\hideWithText}[1]{\hiddenText}

\newcommand{\SecLong}{Section}

\DeclareMathOperator*{\argmin}{arg\,min}

\newcommand{\tran}{^{\mathsf{T}}}

\newcommand{\inv}{^{-1}}

\newcommand{\ones}{{\mathbf 1}}
\newcommand{\zero}{{\mathbf 0}}

\newcommand{\matTwo}[1]{\left[\begin{array}{cc}  #1  \end{array}\right]}

\newcommand{\Real}[1]{ { {\mathbb R}^{#1} } }

\newcommand{\SEthree}{\ensuremath{\mathrm{SE}(3)}\xspace}

\newcommand{\SOthree}{\ensuremath{\mathrm{SO}(3)}\xspace}

\newcommand{\MA}{\M{A}}
\newcommand{\MB}{\M{B}}
\newcommand{\MC}{\M{C}}

\newcommand{\MU}{\M{U}}
\newcommand{\MR}{\M{R}}

\newcommand{\MI}{\M{I}}
\newcommand{\MV}{\M{V}}

\newcommand{\MT}{\M{T}}
\newcommand{\MX}{\M{X}}

\newcommand{\MZ}{\M{Z}}

\newcommand{\vb}{\boldsymbol{b}}

\newcommand{\vn}{\boldsymbol{n}}

\newcommand{\vt}{\boldsymbol{t}}
\newcommand{\vxx}{\boldsymbol{x}}
\newcommand{\vy}{\boldsymbol{y}}

\newcommand{\scenario}[1]{{\smaller \sf#1}\xspace}

\newcommand{\name}{\scenario{C-3PO}}
\newcommand{\nameLong}{Certifiable 3D POse\xspace}
\newcommand{\selfsupervised}{self-supervised\xspace}
\newcommand{\Selfsupervised}{Self-supervised\xspace}

\newcommand{\blue}[1]{{\color{blue}#1}}

\newcommand{\red}[1]{{\color{red}#1}}

\newcommand{\linkToPdf}[1]{\href{#1}{\blue{(pdf)}}}
\newcommand{\linkToPpt}[1]{\href{#1}{\blue{(ppt)}}}
\newcommand{\linkToCode}[1]{\href{#1}{\blue{(code)}}}
\newcommand{\linkToWeb}[1]{\href{#1}{\blue{(web)}}}
\newcommand{\linkToVideo}[1]{\href{#1}{\blue{(video)}}}
\newcommand{\linkToMedia}[1]{\href{#1}{\blue{(media)}}}
\newcommand{\award}[1]{\xspace} %

 \graphicspath{{./figures/}}
\usepackage[compact]{titlesec}

\newcommand{\arxivomit}[1]{}
\newcommand{\arxivadd}[1]{#1}

\newcommand{\newpart}[1]{\blue{#1}}

\newcommand{\Sec}{\SecLong}

\IEEEoverridecommandlockouts

\begin{document}

\title{Certifiable Object Pose Estimation: \\ {Foundations, Learning Models, and Self-Training} }
\author{Rajat Talak, Lisa Peng, and Luca Carlone
\thanks{The authors are with the Laboratory of Information and Decision Systems (LIDS), Massachusetts Institute of Technology, Cambridge, MA 02139, USA. Corresponding author: Rajat Talak (email: {\tt talak@mit.edu}) }
\thanks{This work was partially funded by ARL DCIST CRA W911NF-17-2-0181, ONR RAIDER N00014-18-1-2828, and NSF CAREER award ``Certifiable Perception for Autonomous Cyber-Physical Systems''.}
}

\IEEEaftertitletext{\vspace{-0.6\baselineskip}}

\newcommand{\nrKeypoints}{\ensuremath{N}\xspace}
\newcommand{\nrPoints}{\ensuremath{n}\xspace}
\newcommand{\nrModelPoints}{\ensuremath{m}\xspace}
\newcommand{\objectKeypoints}{\ensuremath{\vb}\xspace}
\newcommand{\objectModel}{\ensuremath{\MB}\xspace}
\newcommand{\objectModelOcc}{\ensuremath{\objectModel^{\text{occ}}}\xspace}
\newcommand{\objectModelPC}{\ensuremath{\hat{\MB}}\xspace}

\newcommand{\inputPC}{\ensuremath{\MX}\xspace}
\newcommand{\outputPC}{\outputA}
\newcommand{\inputPCfull}{\ensuremath{\MX}^*\xspace}
\newcommand{\noisePC}{\ensuremath{\vn_w}\xspace}
\newcommand{\observableCorrectness}{observable correctness\xspace}
\newcommand{\ObservableCorrectness}{Observable correctness\xspace}

\newcommand{\kpCorrection}{\ensuremath{\Delta \vy}\xspace}
\newcommand{\kpOptCorrection}{\ensuremath{\Delta \vy^{\ast}}\xspace}
\newcommand{\kpDetected}{\ensuremath{\tilde{\vy}}\xspace}
\newcommand{\kpCorrected}{\ensuremath{\vy}\xspace}

\newcommand{\predTransform}{\ensuremath{\hat{\MT}}\xspace}
\newcommand{\gtTransform}{\ensuremath{\MT^{\ast}}\xspace}
\newcommand{\aTransform}{\ensuremath{\MT}\xspace}
\newcommand{\aRotation}{\ensuremath{\MR}\xspace}
\newcommand{\aTranslation}{\ensuremath{\vt}\xspace}

\newcommand{\predModelKP}{\ensuremath{\hat{\vy}}\xspace}
\newcommand{\predModel}{\ensuremath{\hat{\MX}}\xspace}
\newcommand{\predRotation}{\ensuremath{\hat{\MR}}\xspace}
\newcommand{\predTranslation}{\ensuremath{\hat{\vt}}\xspace}
\newcommand{\indicator}[1]{\ensuremath{\mathbb{I}\left\{ #1 \right\}}}

\newcommand{\Model}{\ensuremath{\calM}\xspace}

\newcommand{\KeyPo}{\text{KeyPO}\xspace} 
\newcommand{\SDF}[2]{\ensuremath{\text{SDF}_{#1}\left(#2\right)\xspace}}

\newcommand{\ndspace}{\ensuremath{\mathcal{ND}^{\ast}}\xspace}
\newcommand\ringring[1]{%
	{%
		\mathop{\kern0pt #1}\limits^{%
			\vbox to-1.85ex{
				\kern-2ex %
				\hbox to 0pt{\hss\normalfont\kern.1em \r{}\kern-.45em \r{}\hss}%
				\vss %
			}%
		}%
	}%
}

\newcommand{\posedkpA}{\ensuremath{\vy'}\xspace}
\newcommand{\posedkpB}{\ensuremath{\vy''}\xspace}
\newcommand{\posedModelA}{\ensuremath{\MX'}\xspace}
\newcommand{\posedModelB}{\ensuremath{\MX''}\xspace}
\newcommand{\posedOutputA}{\ensuremath{(\posedkpA, \posedModelA)}\xspace}
\newcommand{\posedOutputB}{\ensuremath{(\posedkpB, \posedModelB)}\xspace}
\newcommand{\posedOutputPred}{\ensuremath{(\predModelKP, \predModel)}\xspace}

\newcommand{\distHausdorff}[2]{\ensuremath{d_{H}(#1, #2)}}
\newcommand{\occFunction}[1]{\ensuremath{\Theta\left( #1\right)\xspace}}
\newcommand{\occFunctionTwoX}{\ensuremath{\Theta'}\xspace}
\newcommand{\occFunctionTwo}[1]{\ensuremath{\Theta'\left( #1 \right)}\xspace}

\newcommand{\ocx}{\ensuremath{\texttt{oc}}\xspace}
\newcommand{\oc}[1]{\ensuremath{\texttt{oc}(#1)}\xspace}
\newcommand{\cfitx}{\ensuremath{\texttt{cf}}\xspace}
\newcommand{\cfit}[1]{\ensuremath{\texttt{cf}(#1)}\xspace}
\newcommand{\ndx}{\ensuremath{\texttt{nd}}\xspace}
\newcommand{\nd}[1]{\ensuremath{\texttt{nd}(#1)}\xspace}

\newcommand{\distSurf}[2]{\ensuremath{D(#1, #2)}\xspace}
\newcommand{\distAB}[2]{\ensuremath{\overrightarrow{d}(#1, #2)}}

\newcommand{\inputSpace}{\ensuremath{\mathbb{\MX}}\xspace}
\newcommand{\outputSpace}{\ensuremath{\mathbb{\MZ}}\xspace}
\newcommand{\outputSpaceSampled}{\ensuremath{\hat{\outputSpace}}\xspace}
\newcommand{\instProblem}[1]{\ensuremath{\calP(#1)}\xspace}
\newcommand{\solSpace}[1]{\ensuremath{\calS(#1)}\xspace}
\newcommand{\diam}[1]{\ensuremath{\text{Diam}[#1]}\xspace}
\newcommand{\distOutSpace}[2]{\ensuremath{d_{\outputSpace}(#1, #2)}\xspace}
\newcommand{\outputA}{\ensuremath{\MZ}\xspace}
\newcommand{\outputB}{\ensuremath{\MZ'}\xspace}
\newcommand{\outputC}{\ensuremath{\MZ''}\xspace}
\newcommand{\outputGT}{\ensuremath{\MZ^{\ast}}\xspace}
\newcommand{\outputPred}{\ensuremath{\hat{\MZ}}\xspace}

\newcommand{\idealObsCorrect}{\ensuremath{\texttt{ObsCorrect}[\Model, \inputPC]}\xspace}
\newcommand{\idealObsCorrectx}{\ensuremath{\texttt{ObsCorrect}}\xspace}
\newcommand{\idealNonDegen}{\ensuremath{\texttt{NonDegeneracy}[\Model, \inputPC]}\xspace}
\newcommand{\idealNonDegenx}{\ensuremath{\texttt{NonDegeneracy}}\xspace}
\newcommand{\genFun}{\ensuremath{\phi}\xspace}
\newcommand{\certifiablyCorrect}[1]{\ensuremath{#1}-\text{certifiably correct}\xspace}

\newcommand{\solSpaceOuter}[1]{\ensuremath{\overline{\calS}(#1)}\xspace}
\newcommand{\problemSet}{\ensuremath{\mathbb{P}(\inputSpace, \outputSpace)}\xspace}
\newcommand{\consisCollection}{\ensuremath{\calH}\xspace}
\newcommand{\consisFun}{\ensuremath{h}\xspace}

\newcommand{\ocB}[1]{\ensuremath{\texttt{oc}'(#1)}\xspace}
\newcommand{\noisePCb}{\ensuremath{\vn^{'}_w}\xspace}
\newcommand{\occFunctionB}[1]{\ensuremath{\Theta'[#1]}\xspace}
\newcommand{\outputPredCT}{\ensuremath{\tilde{z}}\xspace}
\newcommand{\predModelCT}{\ensuremath{\tilde{\MX}}\xspace}

\newcommand{\objectModelSubset}{\ensuremath{\MA}\xspace}

\newcommand{\omittext}[1]{}

\newcommand{\zetaCorrectness}{$\zeta$-correctness\xspace}
\newcommand{\zetaCorrect}{$\zeta$-correct\xspace}

\newcommand{\predModelKPCorrection}{\ensuremath{\predModelKP(\kpCorrection)}\xspace}
\newcommand{\predModelCorrection}{\ensuremath{\predModel(\kpCorrection)}\xspace}

\newcommand{\deltaIndicator}{\ensuremath{\delta_{ind}}\xspace} 
\makeatletter
\let\@oldmaketitle\@maketitle%
\renewcommand{\@maketitle}{
	\@oldmaketitle%
	\vspace{5mm}
	\begin{minipage}{\textwidth}
		\begin{center}
			\begin{tabular}{cc}
				 \hfill 
				 \includegraphics[trim=-110 -50 -120 0,clip,width=0.18\linewidth]{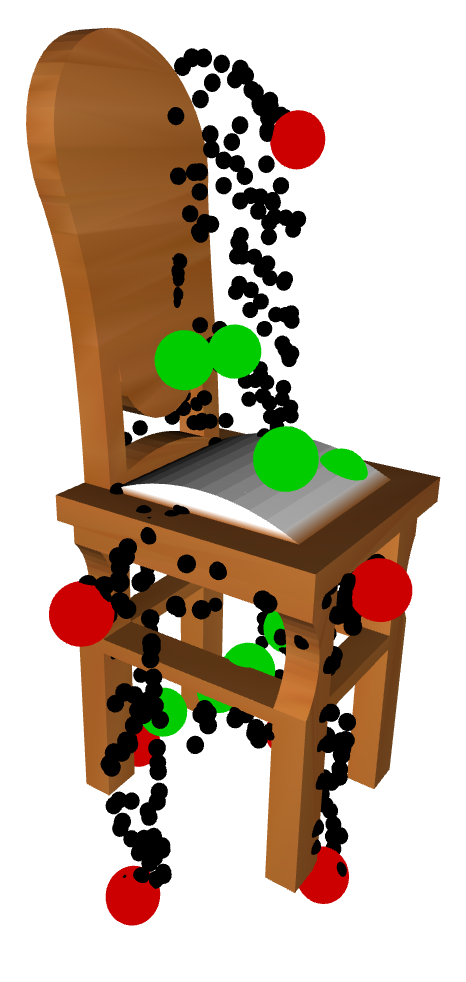}  
				 \vspace{-4mm}
				 \hfill 
				 \includegraphics[trim=-90 -120 -90 0,clip,width=0.18\linewidth]{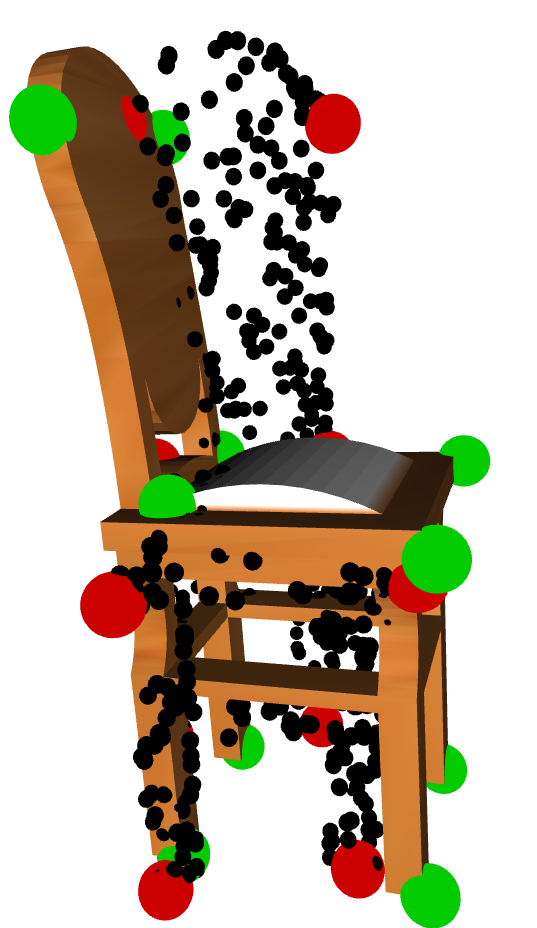}
				 \hfill
				 \includegraphics[trim=-90 -70 -90 0,clip,width=0.18\linewidth]{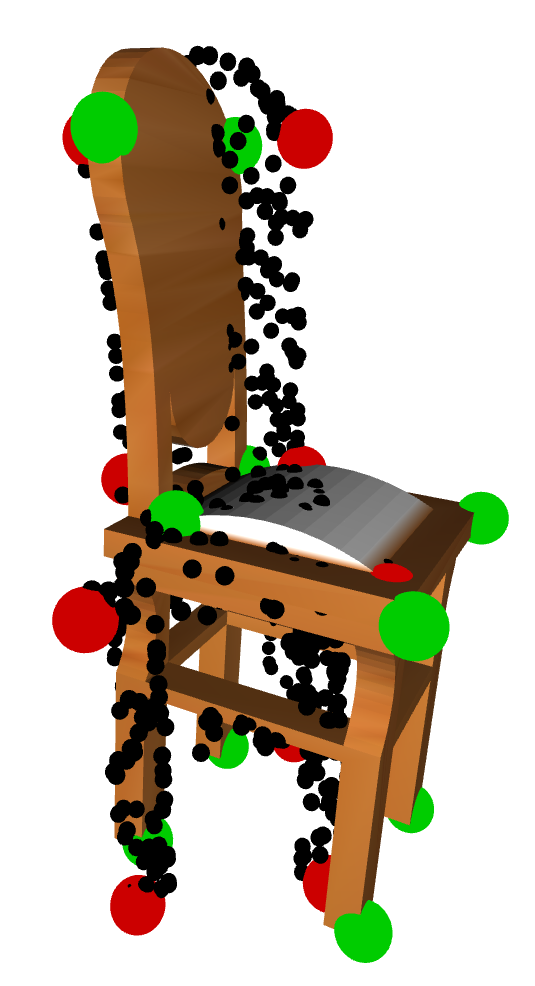}
				 \hfill
				 \includegraphics[trim=-130 -50 -130 0,clip,width=0.18\linewidth]{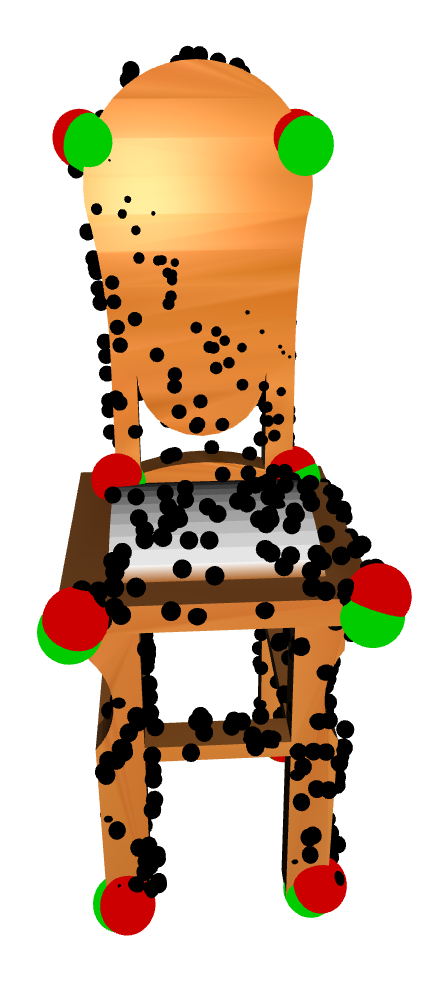}
				 \hfill 
				 \includegraphics[trim=-140 -20 -140 0,clip,width=0.18\linewidth]{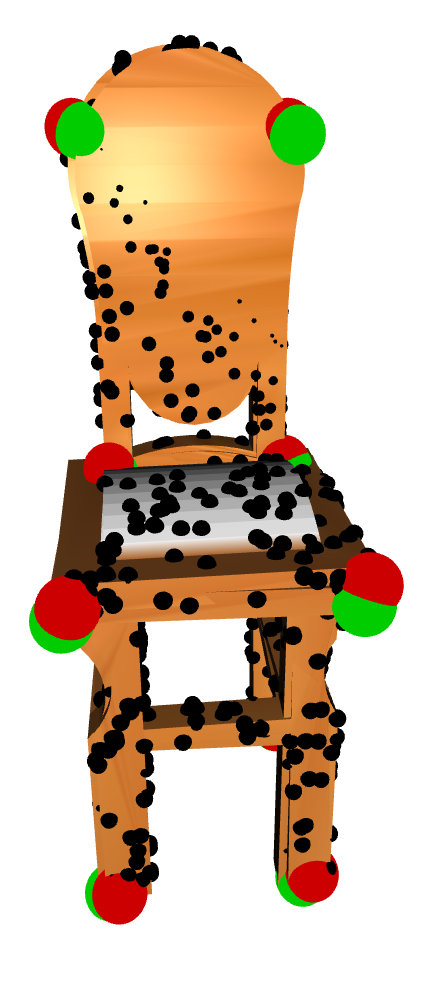}   
				 \hfill
				\\
				\vspace{-3mm}
				\hfill 
				(a) \red{$\ocx = 0$}, $\ndx=1$
				\hfill 
				(b) \red{$\ocx = 0$}, $\ndx=1$
				\hfill 
				(c) \red{$\ocx = 0$}, $\ndx=1$
				\hfill 
				(d) $\ocx = 1$, $\ndx=1$
				\hfill 
				(e) GT vs Estimate 
				\hfill 
				\\
				\vspace{-6mm}
				\hfill 
				\includegraphics[trim=-160 -180 -160 0,clip,width=0.18\linewidth]{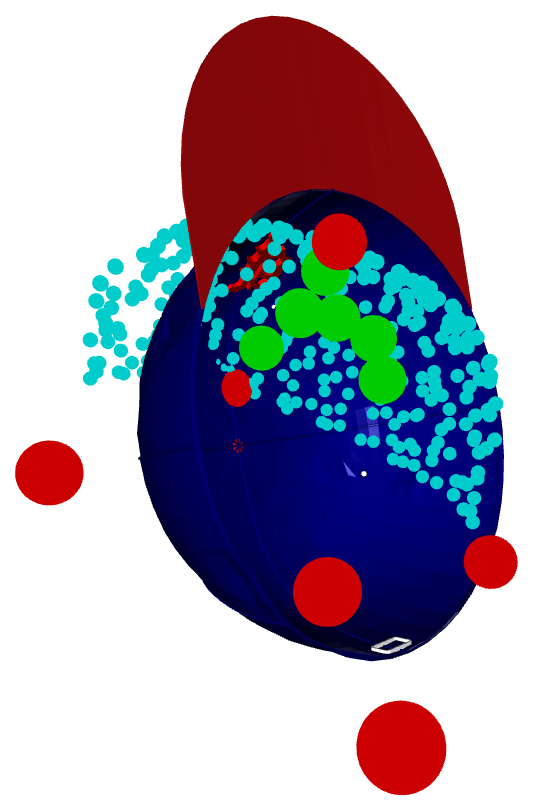} 
				\hfill 
				\includegraphics[trim=-140 -220 -160 -100,clip,width=0.18\linewidth]{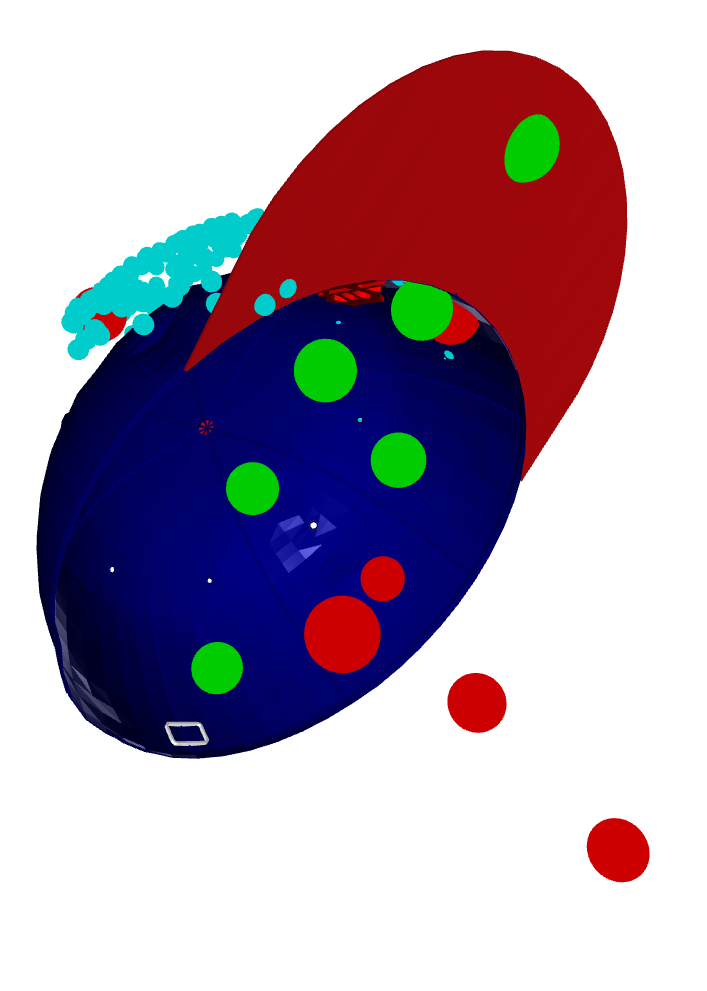}  
				\hfill 
				\includegraphics[trim=-140 -200 -160 0,clip,width=0.18\linewidth]{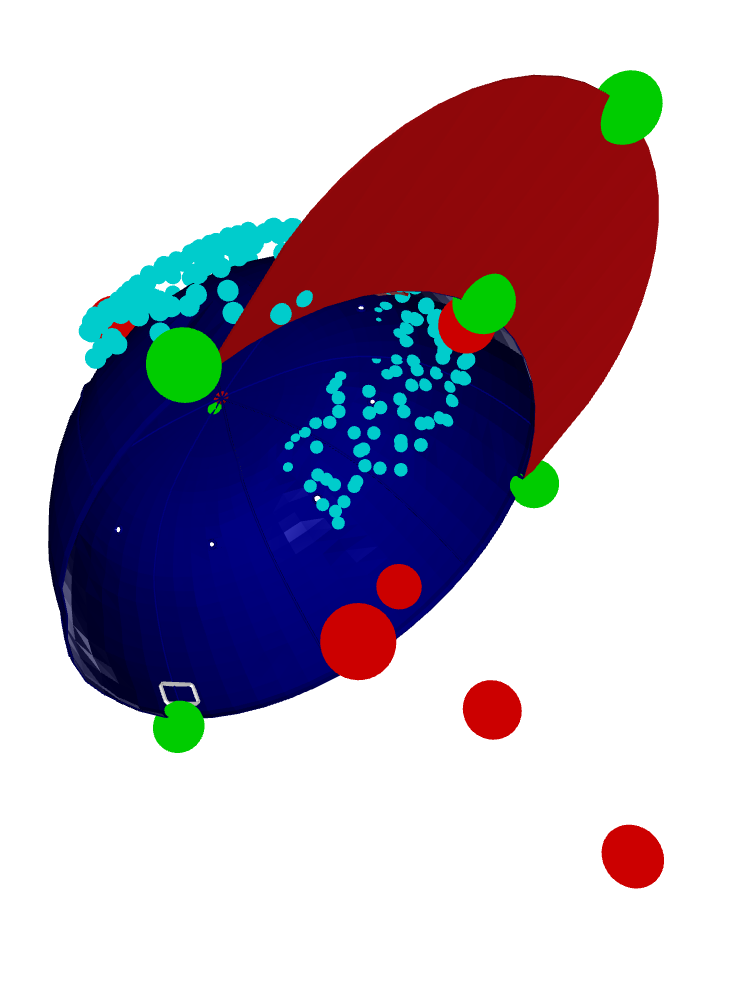}  
				\hfill 
				\includegraphics[trim=-140 -210 -160 0,clip,width=0.18\linewidth]{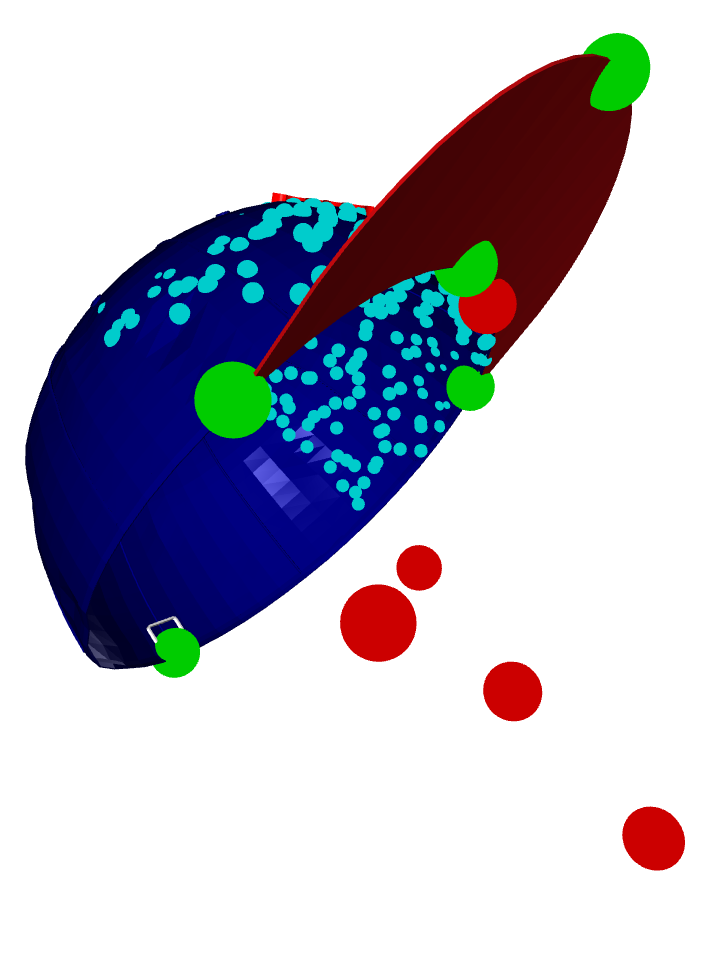} 
				\hfill 
			    \includegraphics[trim=-140 -230 -160 -100,clip,width=0.18\linewidth]{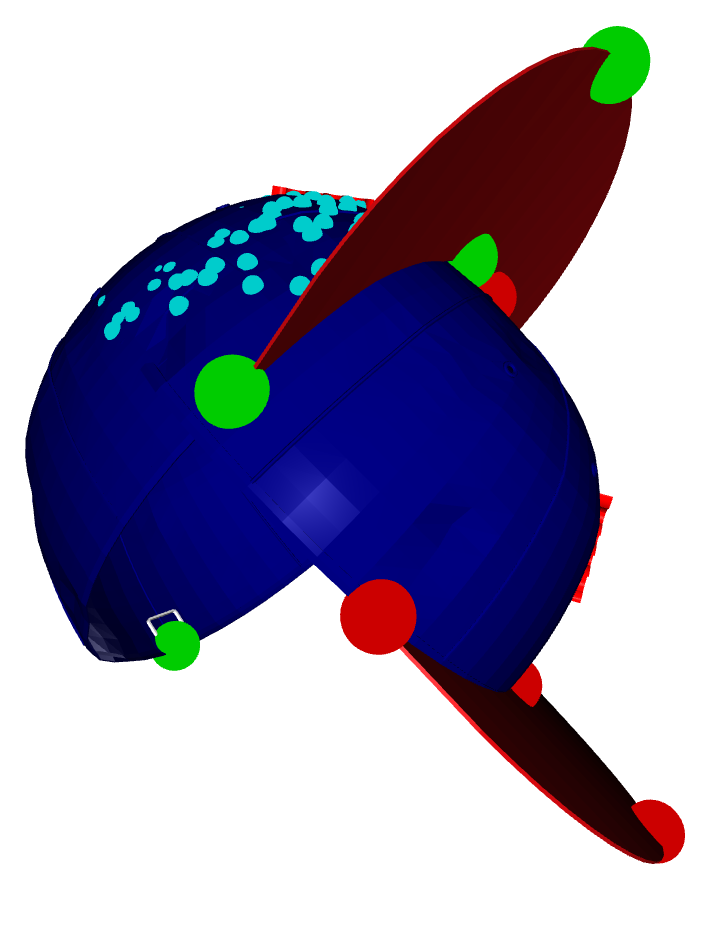}  
			    \hfill 
				\\
				\hfill 
				(f) \red{$\ocx = 0$}, \red{$\ndx=0$}
				\hfill 
				(g) \red{$\ocx = 0$}, \red{$\ndx=0$}
				\hfill 
				(h) \red{$\ocx = 0$}, \red{$\ndx=0$}
				\hfill 
				(i) $\ocx = 1$, \red{$\ndx=0$}
				\hfill 
				(j) GT vs Estimate 
				\hfill 
			\\
			\end{tabular}
		\end{center}
		\vspace{-2mm}
	\end{minipage}
	\\
	
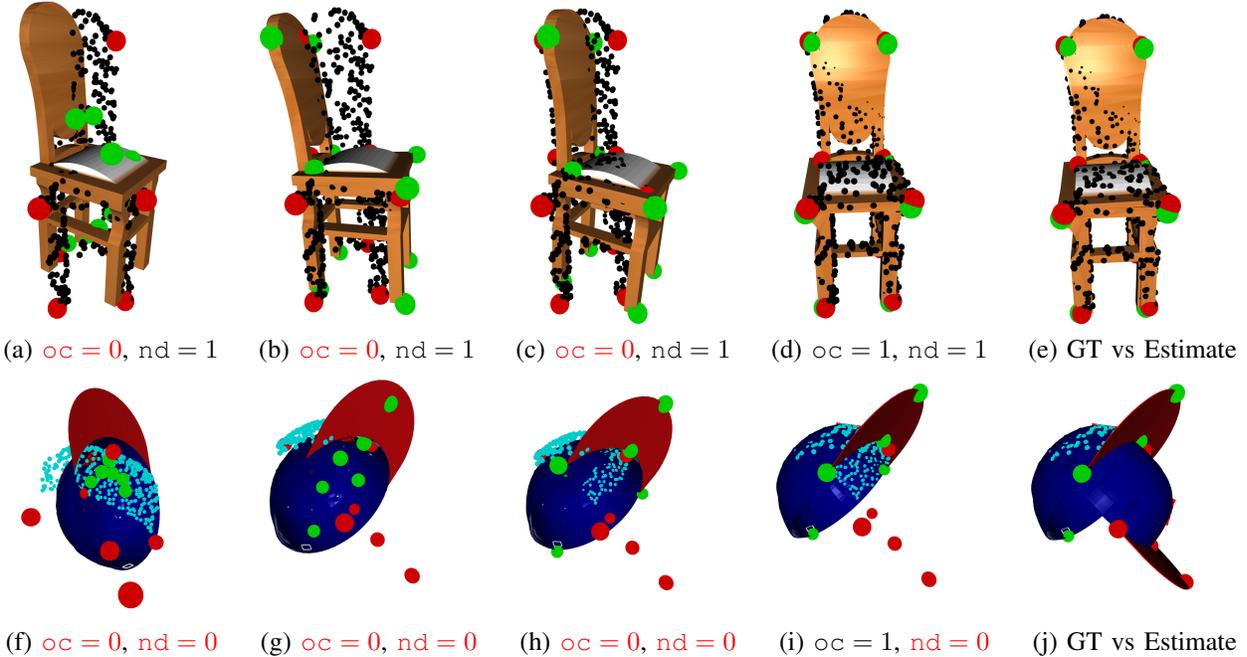
\captionof{figure}{
	We propose \name, a certifiable and self-supervised pose estimation model, that estimates the pose of an object from partial point cloud observations and computes a certificate of correctness for the resulting outputs. 
	\name detects semantic keypoints, corrects them using a differentiable optimization layer, and computes two certificates of \observableCorrectness (\ocx) and non-degeneracy (\ndx). 
	When $\ocx =1 $ and $\ndx=1$, the \name output is guaranteed to be correct.
	The first row shows \name processing an input point cloud of a \emph{chair} (each subfigure also shows the keypoints (green) and the object CAD model arranged according to the pose computed at different iterations of the \name corrector; ground-truth keypoints are shown in red): (a) detected keypoints, (b)-(c) \name corrector result at iterations 4, 10, and 300, 
	and (e) ground-truth vs. final \name estimate. 
	The second row shows \name processing the point cloud of a \emph{cap}: (f) detected keypoints, (g)-(i) \name corrector result at iterations 5, 10, 100, %
	and (j) ground-truth vs. final estimate. The non-degeneracy certificate $\ndx = 0$ for the second row indicating that the input point cloud does not contain sufficient information to uniquely estimate the object pose.
		\label{fig:method-overview}
		\vspace{-5mm}}
}
\makeatother

\maketitle
\arxivadd{
\begin{tikzpicture}[overlay, remember picture]
\path (current page.north east) ++(-4.0,-0.4) node[below left] {
	This paper has been accepted for publication in IEEE Trans. Robotics 2023. Please cite the paper as:
};
\end{tikzpicture}
\begin{tikzpicture}[overlay, remember picture]
\path (current page.north east) ++(-2.0,-0.9) node[below left] {
	R. Talak, L. Peng, and L. Carlone, ``Certifiable Object Pose Estimation: Foundations, Learning Models, and Self-Training’’,
};
\end{tikzpicture}
\begin{tikzpicture}[overlay, remember picture]
\path (current page.north east) ++(-8.5,-1.4) node[below left] {\emph{IEEE Trans. Robotics}, 2023.
};
\end{tikzpicture}
}

\vspace*{-\baselineskip}
\let\subparagraph\llncssubparagraph
\titlespacing{\section}{0pt}{5mm}{2mm}
\titlespacing{\subsection}{0pt}{3mm}{2mm}

\begin{abstract} 
We consider a \emph{certifiable} object pose estimation problem, where 
---given a partial point cloud of an object--- the goal is to not only estimate the object pose, but also to provide a certificate of correctness for the resulting estimate. 
Our first contribution is a general theory of certification for end-to-end perception models. 
In particular, we introduce the notion of \emph{\zetaCorrectness}, which bounds the  distance between an estimate and the ground truth.
We then show that \zetaCorrectness can be assessed by implementing two certificates: (i)~a certificate of \emph{\observableCorrectness}, that asserts if the model output is consistent with the input data and prior information,
(ii)~a certificate of \emph{non-degeneracy}, that asserts whether the input data is sufficient to compute a unique estimate.
Our second contribution is to apply this theory and design a new learning-based certifiable pose estimator.
In particular, we propose \name, a semantic-keypoint-based pose estimation model, augmented with the two certificates, to solve the certifiable pose estimation problem. 
\name also includes a \emph{keypoint corrector}, implemented as a differentiable optimization layer, that can correct large detection errors (\eg due to the sim-to-real gap). %
Our third contribution is a novel self-supervised training approach that uses our certificate of \observableCorrectness to provide the supervisory signal to \name during training. 
In it, the model trains only on the observably correct input-output pairs produced in each batch and at each iteration. As training progresses, we see that the observably correct input-output pairs grow, eventually reaching near $100\%$ in many cases.
We conduct extensive experiments to evaluate the performance of the corrector, the certification, and the proposed \selfsupervised training using the ShapeNet and YCB datasets. 
The experiments show that (i) 
standard semantic-keypoint-based methods (which constitute the backbone of \name) outperform more recent alternatives in challenging problem instances, (ii) \name further improves performance and significantly outperforms all the baselines, (iii) \name's certificates are able to discern correct pose estimates. 
We release the implementation and an interactive visualization of all the results presented in this paper at: \url{https://github.com/MIT-SPARK/C-3PO} and \url{https://github.com/MIT-SPARK/pose-baselines}.

\end{abstract} 
\IEEEpeerreviewmaketitle

\section{Introduction}
\label{sec:intro}

%
%
Object pose estimation is a crucial prerequisite for robots to understand and interact with their surroundings.
In this paper, we consider the problem of object pose estimation using depth or partial point cloud data, which arises when the robot is equipped with depth sensors (\eg lidar, RBG-D cameras) or has to postprocess a 3D point cloud (\eg resulting from a 3D reconstruction pipeline).
Local-feature-based registration methods can be re-purposed to solve the object pose estimation problem, see, \eg~\cite{Yang20tro-teaser}. Recent work, however, argues that either a neural model that mimics the ICP algorithm~\cite{Aoki19cvpr-pointnetlk, Li21cvpr-PointNetLKRevisited}, or approaches based on learned global representations~\cite{Yuan20eccv-DeepGMRLearning, Li21nips-LeveragingSE} yield better performance. 
A known drawback of these methods is that they are not expressive enough to yield high pose estimation accuracy %
when the object point cloud suffers from occlusions. %
Object pose estimation from partial or depth point clouds, therefore, remains an open problem~\cite{Yuan20eccv-DeepGMRLearning, Li21cvpr-PointNetLKRevisited, Li21nips-LeveragingSE}.

%
%

%
%
When using a learning-based approach for pose estimation,
another major challenge is to guarantee ---or at least assess--- the trustworthiness of 
the predictions of neural-network-based models. Robot perception requires models that can be relied upon for the robot to act safely in the world. 
Perception failures can lead to catastrophic consequences including loss of life~\cite{uberAccident}. %
Recent progress has been made towards certifying geometric optimization problems~\cite{Yang22pami-certifiablePerception, Yang20tro-teaser, Garcia21IVC-certifiablerelativepose}, which form the back-end of common perception pipelines. 
However, these approaches only produce certificates of optimality for the back-end optimization, 
but might still fail without notice if the perception front-end (\eg the feature detector) produces largely incorrect results.
Here we argue that we need to go beyond certifiable optimization and need to certify the \emph{entire} perception pipeline, \ie we need to produce certificates that can determine if the output produced by an end-to-end (\eg opaque box) perception pipeline is correct or not.
To date, there is no theory developed for this purpose for general perception tasks, let alone for the case of object pose estimation. 
The current literature only offers statistical guarantees for specific problems~\cite{Yang22rsswk-ConformalSemantic} or is restricted to data near the training regime~\cite{Shafer08jmlr-TutorialConformal, Angelopoulos22arxiv-GentleIntroduction, Xu21iclr-FastComplete, Wang21nips-BetaCROWNEfficient, Albarghouthi21arxiv-IntroductionNeural}.
Finally, most of the recent progress on pose estimation has focused on supervised learning and relies on large annotated datasets~\cite{Gumeli22cvpr-ROCARobust, park19iccv-Pix2PosePixelWise, Xiang17rss-posecnn, Wang19cvpr-DenseFusion6D,  Kundu18-3dRCNN}. However, 
obtaining human-labeled data is not only costly, but also 
robot-dependent: 
a change in
placement of the sensor/camera (\eg from the top of a self-driving car to the head of a quadruped robot) is likely to distort the predictions produced by the trained model. Though photo-realistic simulators provide a partial answer to this problem~\cite{Tremblay18corl-DeepObject, hofer21survey-Sim2RealRobotics, li22ral-SimtoRealObject}, a sim-to-real gap is likely going to remain and needs to be tackled. 
Here, we argue that a scalable pose estimation method must rely on self-supervision rather than human-labeled data.
Very few works have tackled pose estimation in a \selfsupervised manner~\cite{Wang20eccv-Self6DSelfSupervised, Wang22pami-OcclusionAwareSelfSupervised, Zakharov20cvpr-Autolabeling3D, Deng20icra-Selfsupervised6D, Chen22arxiv-SimtoReal6D, Yang21cvpr-sgp, Li21nips-LeveragingSE}.
%
%

\subsection{Contributions}

We address the twin challenges of certifiability and self-supervised training by noting that certification can provide a supervision signal for self-training (\ie if we can identify correct predictions, we can learn from them).
More specifically, we establish the \emph{foundations} of certifiable object pose estimation using learning-based models, by defining a notion of correctness and developing certificates that can assert correctness in practice. We then propose a new \emph{learning-based model}, named \name (\nameLong), for certifiable pose estimation from depth or point cloud data. Finally, we show how to use the proposed certificates to enable \emph{self-supervised training} of \name. We unpack each contribution below.

\begin{figure}[t]
	\vspace{-0.2cm}
	\centering
	\begin{subfigure}{0.29\columnwidth}
		\centering
		\vspace{7mm}
		\includegraphics[trim=10 0 10 10,clip,width=\linewidth]{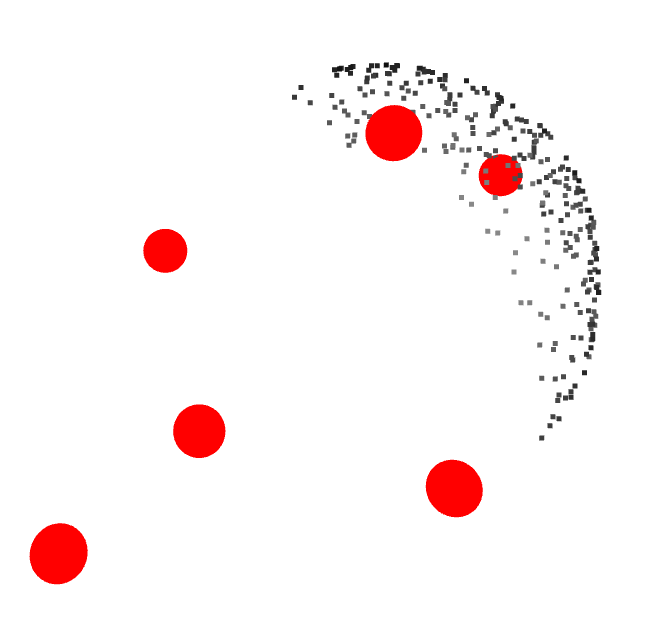}
		\vspace{-5.8mm}
		\caption{}
	\end{subfigure}
	\begin{subfigure}{0.29\columnwidth}
		\centering
		\includegraphics[trim=0 0 0 0,clip,width=\linewidth]{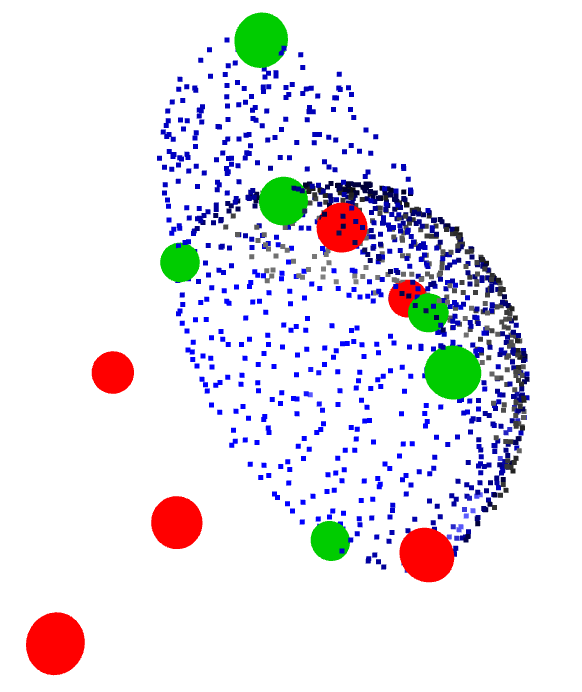}
		\caption{}
	\end{subfigure}
	\begin{subfigure}{0.29\columnwidth}
		\centering
		\includegraphics[trim=-10 -10 -10 -10,clip,width=\linewidth]{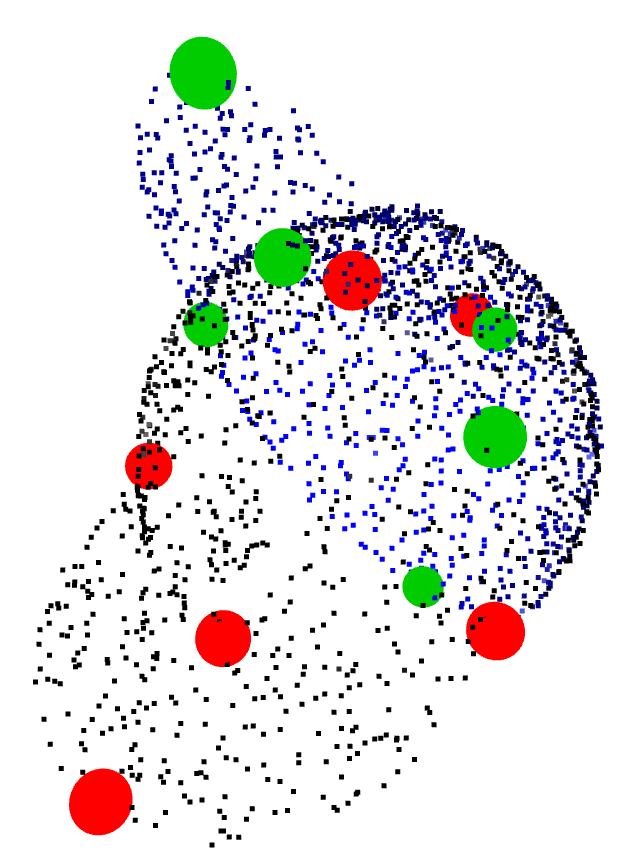}
		\vspace{-8mm}
		\caption{}
	\end{subfigure}
	\vspace{-2mm}
	\caption{Example of degenerate input for pose estimation. {(a) Input: partial point cloud of an object (gray), namely the ``cap'' in ShapeNet~\cite{Chang15arxiv-shapenet}; (b) Estimated object pose shown as a posed object CAD model (blue);} (c) Estimated (blue) and ground-truth (gray) pose overlaid as two posed point clouds. %
	Ground truth semantic keypoints are in red, detected keypoints in green.}
	\label{fig:expt_shapenet_degeneracy}
	\vspace{-0.5cm}
\end{figure}

\myParagraph{(I) Theory of Certifiable Models and Certifiable Object Pose Estimation} 
We start by defining the notion of \emph{\zetaCorrectness} that discerns whether an estimate is correct (\ie matches the ground truth object) or not (\cref{def:ep-cert-correct-gen}). 
Using this, we develop a general theory of certifying learning-based models (\cref{sec:cert-model}).
We show that by implementing two certificates, namely, a certificate of \emph{\observableCorrectness}, and a certificate of \emph{non-degeneracy}, we can infer \zetaCorrectness (\cref{thm:epsilon-cert-general}). 
The \observableCorrectness certificate determines whether an output produced by a model is 
consistent with the sensor data and prior knowledge (\eg knowledge of the object shape).
Non-degeneracy, on the other hand, ensures that for the provided input there exists a unique, correct solution. A case of degeneracy arises often, for instance, in object pose estimation when the partial point cloud is so occluded that there is more than one correct pose fitting the data; see \cref{fig:expt_shapenet_degeneracy}. 
Therefore, a \emph{certifiable model} is a learning-based model that implements these two certificates 
---of \observableCorrectness and non-degeneracy--- so that for every output produced by the model, its \zetaCorrectness may be checked. 

We use this theory to introduce the problem of \emph{certifiable object pose estimation} (\cref{prob:cert-pe}), where ---given a partial point cloud of an object--- we aim to estimate the pose of the object in the point cloud and provide a certificate that guarantees that the pose estimate is \zetaCorrect (\cref{sec:problem}).

\myParagraph{(II) C-3PO: a Certifiable Object Pose Estimation Model} 
We propose \name (\nameLong) to solve the certifiable object pose estimation problem (\cref{sec:architecture}). 
\name first  detects semantic keypoints from the input, partial point cloud using a trainable regression model. %
To correct for potential keypoint-detection errors (\eg induced by a sim-to-real gap),
\name adds a keypoint \emph{corrector} module that corrects some of the keypoint-detection errors.
The corrector module is implemented as a differentiable optimization module~\cite{Gould19arXiv-deepDeclarativeNetworks, Agrawal19NIPS-diffConvex} (\cref{sec:corrector}): it takes in the detected keypoints and produces a correction, by solving a non-linear, non-convex optimization problem. Although the corrector optimization problem is hard to analyze, we show that it can be differentiated through very easily. We provide an exact analytical expression of the gradient of the output (the correction), given the input (detected keypoints) to the corrector (\cref{thm:corrector_bkprop}), which we use to implement back-propagation. We also implement a \emph{batch gradient descent} that solves a batch of corrector optimization problems in parallel on GPU, thereby speeding up the forward pass.
Experimental analysis shows that the corrector is able to correct high-variance perturbations to the semantic keypoints (\cref{sec:expt_corrector_analysis}). 

Finally, using our theory of certifiable models, we derive two certificates, namely, a certificate of \observableCorrectness and a certificate of non-degeneracy for \name (\cref{sec:pose-certificates}). We prove that this enables \name to check \zetaCorrectness in the forward pass (\cref{thm:pose-cert-guarantee}). We also empirically observe that the implemented certificates 
are able to identify correct predictions in practice (\cref{sec:expt_shapenet,sec:expt_ycb}).

\myParagraph{(III) Certificate-based Self-supervised Training}
\name allows addressing the twin problems of certification 
and self-supervision 
by noting the following: if at test time we can use the proposed certificates to distinguish \zetaCorrect outputs, we can use the latter for further training. 
This forms the basis of our new self-supervised training, which
  proceeds by training only on the observably correct instances (\ie input-output pairs), in each batch and in each training iteration. As the training progresses, the number of observably correct instances increases, eventually resulting in a fully self-trained model.

We test the \selfsupervised training and certification on depth-point-cloud dataset generated using ShapeNet objects~\cite{Chang15arxiv-shapenet} and on the depth point clouds extracted from RGB-D images in the YCB dataset~\cite{calli15-YCBobject} (\cref{sec:expt_shapenet,sec:expt_ycb}). In both cases, we observe that while the simulation-trained \name performs poorly (indicating a large sim-to-real gap), our \selfsupervised training is able to successfully train \name with self-supervision and significantly outperforms existing baselines.
As a further experiment, we show that the %
proposed \selfsupervised training
 can be even used to self-train \name when the object category labels are not available, for instance \name is able to learn to detect a ``chair'' from a large unlabeled dataset containing chairs, tables, cars, airplanes, etc. (Section~\ref{sec:expt_fullss}).

An interactive visualization of all the results presented in this paper is made available at: \url{https://github.com/MIT-SPARK/C-3PO} and \url{https://github.com/MIT-SPARK/pose-baselines}.

\section{Related Work}
\label{sec:lit}

 Here we provide a non-exhaustive review about object pose estimation, including recent image-based estimation methods.

\myParagraph{Object Pose Estimation from RGB and RGB-D Inputs}
Recent progress in object pose estimation has been fueled by the availability of pose-annotated datasets~\cite{Hodan18eccv-BOPBenchmark, Hodan20eccvw-BOPChallenge, Calli15ram-BenchmarkingManipulation, Hodan17-TLESSRGBD}.  
State-of-the-art methods have evolved from correspondence-based methods (which extract correspondences between the sensor data and the object CAD model, and use them to register the model to the data) and template-based methods (which attempt to match the sensor data to templates consisting of the CAD model of the object rendered at various poses) to direct regression, augmented with pose refinement. 
Xiang\setal	\cite{Xiang17rss-posecnn} regress the object pose using a convolutional neural network (CNN) backbone. 
Wang\setal~\cite{Wang21cvpr-GDRNetGeometryGuided} propose GDR-Net, which instead extracts a $6$-dimensional rotation representation, thus, avoiding discontinuity issues~\cite{Zhou19cvpr-ContinuityRotation}. 
Nguyen\setal \cite{Nguyen22cvpr-Templates3D} revive template-based methods by showing their generalization power.
Li\setal~\cite{Li18eccv-DeepIMDeep} propose DeepIM, which uses a pre-trained optical flow detection model to guide and train object pose refinement. 
Labbe\setal~\cite{Labbe20eccv-CosyPose} propose CosyPose, which showed state-of-the-art performance on the BOP'20 pose estimation challenge~\cite{Hodan20eccvw-BOPChallenge}, relying on the pose regression network in~\cite{Xiang17rss-posecnn}, followed by the pose refinement proposed in~\cite{Li18eccv-DeepIMDeep}. The state-of-the-art performance in the BOP'22 challenge was also attained by a model that uses a pose regression network~\cite{Liu22eccvw-gdrnppBOP, Wang21cvpr-GDRNetGeometryGuided}. Recent work also attempts object pose estimation, without the availability of the object CAD model~\cite{He22nips-oneposepp}.

We note two gaps in the pose estimation literature. 
First, most of the approaches above are RGB or RGB-D based, and investigating object pose estimation from only depth or partial point clouds has received little attention. 
Second, these methods rely on large pose-annotated datasets, 
with only few attempts to develop self-supervised approaches~\cite{Wang20eccv-Self6DSelfSupervised, Wang22pami-OcclusionAwareSelfSupervised}. 

We next review different approaches for \emph{point-cloud-only} pose estimation, including
approaches based on local feature detection/description, global features, and semantic keypoints.

\myParagraph{Local Feature Detectors and Descriptors}
Local-feature-based methods extract local features (potentially with descriptors), determine point-to-point correspondences, and solve a robust registration problem, 
which estimates the object pose while filtering out
incorrect correspondences.
Significant progress has been made in recent years towards (i) better features to extract  point-to-point correspondences~\cite{Rusu09icra-fast3Dkeypoints, Yew18eccv-3dfeatnet, Deng18cvpr-PPFNetGlobal, Deng18eccv-PPFFoldNetUnsupervised, Choy19iccv-FCGF, Zeng17cvpr-3dmatch}, (ii) robust registration algorithms~\cite{Yang16pami-goicp, Yang22pami-certifiablePerception, Yang20tro-teaser}, as well as (iii) end-to-end registration pipelines that simultaneously learn correspondences and obtain the relative poses~\cite{Zhou16eccv-fastGlobalRegistration, Sarode19arxiv-PCRNetPoint, Wang19nips-prnet, Choy20cvpr-deepGlobalRegistration, Huang21cvpr-PREDATORRegistration, Yew20cvpr-RPMNetRobust, Gojcic19cvpr-3Dsmoothnet}.
Modern robust registration methods have been shown to compute accurate estimates even in the face of 99\% random outliers~\cite{Yang20tro-teaser}. %
Several end-to-end learning-based registration approaches have been proposed%
, but only a few can be trained in a self-supervised manner. 
A weakly-supervised approach, using a triplet loss, is developed in~\cite{Yew18eccv-3dfeatnet}.
Self-supervised partial-to-partial scene registration is considered in~\cite{Liu22sensors-SelfsupervisedPoint, Banani21cvpr-UnsupervisedRUnsupervised}.
In particular, Banani\setal~\cite{Banani21cvpr-UnsupervisedRUnsupervised} use the fact that two successive frames inherit some geometric and photometric consistency, and leverage differentiable rendering to implement the training loss. 
A teacher-student-verifier framework for joint feature learning and registration is considered in~\cite{Yang21cvpr-sgp}.

\myParagraph{Global Features}
Methods that extract and use global features have been proposed for point cloud alignment and object pose estimation. 
Sarode~\etal~\cite{Sarode19arxiv-PCRNetPoint} propose an architecture that uses PointNet encoding to extract global features. %
Huang~\etal \cite{Huang20cvpr-FeatureMetricRegistration} extract global features from the point clouds 
and then estimate the pose by minimizing a feature-metric projection error.  
Li~\etal \cite{Li21cvpr-PointNetLKRevisited} propose a similar approach, but uses Lucas-Kanade-like iterations for minimizing the feature-metric error.
Yuan~\etal \cite{Yuan20eccv-DeepGMRLearning} propose a probabilistic registration paradigm, which extracts pose-invariant point correspondences, as latent Gaussian mixture model.
Zhu~\etal~\cite{Zhu22corl-CorrespondenceFreePoint} addresses point cloud registration by extracting \SOthree-equivariant features.
Sun~\etal~\cite{Sun21nips-CanonicalCapsules} use a capsule network encoder.
Among these approaches PointNetLK~\cite{Li21cvpr-PointNetLKRevisited} and DeepGMR~\cite{Yuan20eccv-DeepGMRLearning} show the best results for object pose estimation, albeit in different scenarios.
Li~\etal~\cite{Li21nips-LeveragingSE} tackle category-level object pose estimation from partial or complete object point clouds, by extracting an \SEthree-invariant shape feature and a \SEthree-equivariant pose feature. 

\myParagraph{Semantic Keypoints}
Semantic keypoint detectors extract keypoints that correspond to specific points on the object to detect (\eg the wheels or headlights of a car), hence circumventing the needs to compute local descriptors. 
Semantic keypoint-based methods have been studied in the context of human pose estimation~\cite{Yang21iccv-TransPoseKeypoint, Liu22cs-RecentAdvances, Pavlakos18cvpr-humanPoseAndShape}.
Pavlakos\setal \cite{Pavlakos17icra-semanticKeypoints, Schmeckpeper22arxiv-singleRGBpose} use class-specific semantic keypoints, extracted from CNN features, to estimate object pose from an RGB image.
Lin\setal \cite{Lin22ral-E2EKEndtoEnd} propose an end-to-end, differentiable pose estimation architecture using RGB images. 
Shi\setal \cite{Shi21rss-pace} propose an active shape model using semantic keypoints, and solve the joint pose and shape estimation problem assuming detected semantic keypoints.  
Yang and Pavone~\cite{Yang22rsswk-ConformalSemantic} obtain performance guarantees on keypoint detections and pose estimates using conformal prediction.
Zhou\setal \cite{Zhou18eccv-StarMapCategoryAgnostic} propose category-agnostic keypoint detection, and argue that using a fixed number of keypoints per-category can be limiting (\eg chair with many legs). 
Vecerik\setal \cite{Vecerik21corl-S3KSelfSupervised} advocate semantic keypoints to be the right visual representation for object manipulation, and provide an efficient training approach to learn instance and category-level keypoints using a small number of annotated images.

{Contrary to the works above, \name does not rely on pose or keypoint annotations for training. Instead, it only assumes the availability of the object CAD model annotated with semantic keypoints.
You\setal \cite{You20cvpr-KeypointNetLargescale} provide a large-scale keypoint annotated dataset on ShapeNet CAD models. Suwajanakorn\setal \cite{Suwajanakorn18nips-DiscoveryLatent} show that semantic keypoint annotations can be learned in a self-supervised manner, from just the object CAD models.}

\myParagraph{Self-Supervised Pose Estimation} 
Few works have tackled self-supervised pose estimation.
In~\cite{Wang20eccv-Self6DSelfSupervised, Wang22pami-OcclusionAwareSelfSupervised}, a pose estimation model is first trained on synthetic RGB-D data, and then refined further with self-supervised training on real, unannotated data; differentiable rendering provides the required supervision signal. Student-teacher iterative schemes are proposed in~\cite{Chen22arxiv-SimtoReal6D,Yang21cvpr-sgp} to bridge the domain gap. %
Another approach is to extract a pose-invariant feature, thereby canonizing only its shape~\cite{Wang19-normalizedCoordinate}, and using it for supervision. 
Zakharov\setal~\cite{Zakharov20cvpr-Autolabeling3D} utilize differentiable rendering of signed distance fields of objects, along with normalized object coordinate spaces~\cite{Wang19-normalizedCoordinate}, to learn 9D cuboids in a self-supervised manner.
Li\setal~\cite{Li21nips-LeveragingSE} extract an \SEthree-invariant feature, which works as a canonical object, and uses it to supervise training with a Chamfer loss.  Sun\setal~\cite{Sun21nips-CanonicalCapsules} tackles point cloud alignment in a self-supervised manner by extracting features using capsule network. 
Deng\setal~\cite{Deng20icra-Selfsupervised6D} propose a way to self-supervise pose estimation by interacting with the objects in the environment; the model gets trained on the data collected autonomously by 
a
manipulator. 

\myParagraph{Certification} Certifying correctness of the model output is crucial for 
safety-critical and high-integrity applications.
Certifiably optimal algorithms return a solution and also provide a certificate of optimality~\cite{Bandeira15arxiv, Yang22pami-certifiablePerception, Yang20tro-teaser} of the solution. Estimation contracts~\cite{Yang20tro-teaser,Carlone22arxiv-estimationContracts}, on the other hand, determine if the input is reasonable enough for the optimal solution to be indeed correct, \ie close to the ground-truth.
Certifiably optimal algorithms have been devised for several geometric perception problems\cite{Yang22pami-certifiablePerception, Yang20tro-teaser, Garcia21IVC-certifiablerelativepose}, %
where the duality gap serves as the certificate of optimality. 
These notions of certification can be applied to optimization problems, which form the back-end of any perception pipeline, but do not apply to the entire end-to-end pipeline or learning-based models.
In this work, we extend the notion of certification to learning-based models. 
Recent works have attempted to address the question of robustness and uncertainty-quantification of learning-based models. Neural network verification attempts to ensure that a learning-based model is robust to small deviations in the input space~\cite{Xu21iclr-FastComplete, Wang21nips-BetaCROWNEfficient, Albarghouthi21arxiv-IntroductionNeural}. Conformal prediction, on the other hand, uses training data to safeguard a learning-based model~\cite{Shafer08jmlr-TutorialConformal, Angelopoulos22arxiv-GentleIntroduction, Papadopoulos02ecml-InductiveConfidence, Lei14jrstats-Distributionfreeprediction,Yang22rsswk-ConformalSemantic}. Given a user-specified probability, it enables a trained model to predict uncertainty sets/intervals, instead of a point estimate. The predicted set is then provably correct, with the specified probability. Conformal prediction works under the assumption that the test set and the dataset used to derive thresholds for set-predictions have the same distribution. Providing probabilistic guarantees under sim-to-real gap or covariate shifts is an active research area~\cite{Tibshirani19nips-ConformalPrediction, Fannjiang22pnas-Conformalprediction}.

\myParagraph{Differentiable Optimization}
Learning-based models have traditionally relied on simple feedfoward functions (\eg linear, ReLU). Recent work has proposed to embed differentiable optimization as a layer in learning-based models~\cite{Agrawal19NIPS-diffConvex, Amos17ICML-optnet, Gould22pami-DeepDeclarative}. A differentiable optimization layer is an optimization problem for which the gradient of the optimal solution, with respect to the input parameters, can be computed and back-propagated. 
Embedding an optimization problem as a differentiable optimization layer enables a learning-based model to explicitly take into account various geometric and physical constraints.
 Gould\setal \cite{Gould22pami-DeepDeclarative} provide generic expressions to differentiate any non-linear optimization problem. Agrawal\setal \cite{Agrawal19NIPS-diffConvex} provide a way to differentiate through convex programs in standard form. 
Amos\setal \cite{Amos17ICML-optnet} focus on quadratic optimization problems. 
Differentiating combinatorial optimization problems is considered in~\cite{Pogancic20iclr-DifferentiationBlackbox, Paulus21icml-CombOptNetFit}, and a differentiable MAXSAT solver is proposed in~\cite{Wang19icml-SATNetBridging}.
Donti\setal~\cite{Donti17nips-TaskbasedEndtoend} propose a way to differentiate a stochastic optimization problem. Teed\setal \cite{Teed21nips-DROID-SLAM} implement differentiable bundle adjustment.

While developing generic techniques to implement derivatives of various classes of optimization problems is useful, there is much scope for efficiency if the problem structure can be exploited to implement simpler differentiation rules~\cite{Amos17ICML-optnet, Jiang203dv-JointUnsupervised, Campbell20eccv-SolvingBlind}. 
\name implements a corrector module, which is a differentiable optimization layer. It solves a non-linear, non-convex optimization problem, but also exploits its specific structure efficiently to compute its derivative.

\section{A Theory of Certifiable Models}
\label{sec:cert-model}
\label{sec:framework}
A learning-based model cannot guarantee \emph{correctness} of the output, in terms of its closeness to the ground-truth. We first provide a general theory to show that implementing two binary-valued certificates can determine the correctness of the model estimate viz-a-viz the ground-truth. Our theory will introduce the notions of \zetaCorrectness, certificates of \emph{\observableCorrectness} and \emph{non-degeneracy}, and \emph{certifiable models}.
We apply this theory to specify and solve the certifiable object pose estimation problem.

\myParagraph{Setup}
Let \inputSpace and \outputSpace be the space of all inputs and outputs, for a problem set that we are solving. We assume the output space \outputSpace is a metric space with a distance metric $\distOutSpace{\cdot}{\cdot}$. 
Moreover, we assume any input $\inputPC \in \inputSpace$ is generated according to a given \emph{generative model}, \ie $\inputPC = \phi(\outputPC^*)$, for some unknown $\outputPC^* \in\outputSpace$; hence, the goal is to estimate $\outputPC^*$ given $\inputPC$. 
We make the following definition of correctness: 
\begin{definition}[\zetaCorrectness] 
\label{def:ep-cert-correct-gen}
We say that the output \outputPC produced by a model, for input $\inputPC$, is \zetaCorrect if 
\begin{equation}
\distOutSpace{\outputPC}{\outputGT} \leq \zeta.
\end{equation}
\end{definition}

A learning-based model \Model maps inputs $\inputPC \in \inputSpace$ to outputs $\outputPC \in \outputSpace$, and is trained to solve a problem \instProblem{\inputPC}; instantiated by the input \inputPC.
Given this, we can define the solution space $\solSpace{\inputPC}$ of the problem. It will play a pivotal role in the definition and analysis of our certificates. 

\begin{definition}[Solution Space] 
Given an input $\inputPC$, the \emph{solution space} is the set of all outputs \outputA that can generate the input \inputPC
using the generative model $\phi(\cdot)$:
\begin{equation}
\label{eq:solspace-from-genFun}
\solSpace{\inputPC} \triangleq \left\{ \outputA \in \outputSpace~|~\genFun(\outputA) = \inputPC \right\}.	
\end{equation}
\end{definition}
The intuition behind our certification approach is as follows: the ground truth $\outputPC^*$ is in the solution space since it generated the input data, hence if we can (i)  produce an estimate in the solution space, and (ii) prove that the solution space is not large, we can conclude that the estimate must be close to the ground truth. Our \emph{certificates} formalize these intuitions.

\myParagraph{Certificates} 
We define 
two certificates to check (i) the \emph{observable correctness} of the output (\ie whether the output \outputA is in the solution space \solSpace{\inputPC} or not) and (ii) the \emph{non-degeneracy} (\ie whether \solSpace{\inputPC} is small, or not); then we prove that the two 
certificates imply correctness.

\begin{definition}[Observable Correctness and Non-Degeneracy]
	\label{def:cert-correctness}
	For model \Model and input \inputPC, the \emph{certificate of observable correctness} is a Boolean condition defined as 
	\begin{equation}\label{eq:correctness}
	\idealObsCorrect = \indicator{\outputA = \Model(\inputPC) \in \solSpace{\inputPC}}.
	\end{equation}
	A \emph{certificate of non-degeneracy} is defined as
	\begin{equation}\label{eq:nondegeneracy}
	\idealNonDegen = \indicator{\diam{\solSpace{\inputPC}} < \delta},
	\end{equation}  
	where $\diam{\solSpace{\inputPC}} = \max_{\outputA, \outputB \in \solSpace{\inputPC}} \distOutSpace{\outputA}{\outputB}$ denotes the diameter of the solution space \solSpace{\inputPC}, and $\delta$ is a small constant. 
\end{definition} 

We now show that the two certificates enable us to determine when a model produces an \zetaCorrect output. 
\arxivadd{The proof is given in Appendix~\ref{app:sec:certifiable-model}.}
\begin{theorem}
\label{thm:epsilon-cert-general}
For any input \inputPC, the output produced by the model $\outputA = \Model(\inputPC)$ is \certifiablyCorrect{\zeta}, \ie 
$\distOutSpace{\outputA}{\outputGT} < \zeta$,
if the certificate of observable correctness~\eqref{eq:correctness} and non-degeneracy~\eqref{eq:nondegeneracy} are both $1$, and $\delta \leq \zeta$ in~\eqref{eq:nondegeneracy}.
\end{theorem}

\myParagraph{Implementing Certificates}
Theorem~\ref{thm:epsilon-cert-general} shows the importance of the certificates in~\eqref{eq:correctness}-\eqref{eq:nondegeneracy}. 
However, in general 
it may be hard to implement~\eqref{eq:correctness}-\eqref{eq:nondegeneracy} as the solution space \solSpace{\inputPC} might be hard to characterize. 
We next prove that using an outer approximation of \solSpace{\inputPC} (\ie $\solSpace{\inputPC} \subseteq \solSpaceOuter{\inputPC}$), which is often easier to obtain in practice, retains the correctness guarantees.
\begin{theorem}
\label{thm:epsilon-cert-general-outer}
Define the certificates $\ocx$ and $\ndx$ as:
\begin{align}
\oc{\Model, \inputPC} &= \indicator{\outputA = \Model(\inputPC) \in \solSpaceOuter{\inputPC}},~\text{and} \\
\nd{\Model, \inputPC} &= \indicator{\diam{\solSpaceOuter{\inputPC}} < \delta}, \label{eq:qqq}
\end{align}
where $\solSpace{\inputPC} \subseteq \solSpaceOuter{\inputPC}$. Then, the output $\outputA = \Model(\inputPC)$ is \certifiablyCorrect{\zeta}, \ie 
$\distOutSpace{\outputA}{\outputGT} < \zeta$, 
if $\oc{\Model, \inputPC}=1$ and $\nd{\Model, \inputPC}=1$, and $\delta \leq \zeta$ in~\eqref{eq:qqq}.
\end{theorem}

In~\cref{sec:architecture}, 
we provide practical implementations of the certificates in~\cref{thm:epsilon-cert-general-outer} for the pose estimation problem.

\myParagraph{Certifiable Model} We define a \emph{certifiable model} to be a triplet $(\Model, \ocx, \ndx)$, where \Model is a learning-based model, and \ocx and \ndx are two certificates that (conservatively) approximate the certificate of observable correctness~\eqref{eq:correctness} and non-degeneracy~\eqref{eq:nondegeneracy}, in order to establish \zetaCorrectness of every output produced. In other words, a certifiable model should always allow for a theorem, that is as follows:
\begin{theorem}[Meta-Theorem for Certifiable Models]
\label{thm:meta-thm}
The output $\outputA = \Model(\inputPC)$ is \certifiablyCorrect{\zeta}, \ie $\distOutSpace{\outputA}{\outputGT} < \zeta$, 
if $\oc{\Model, \inputPC}=1$ and $\nd{\Model, \inputPC}=1$.
\end{theorem}

\begin{remark}[Relation to Certifiable Algorithms]
A parallel can be drawn between the two certificates in~\eqref{eq:correctness}-\eqref{eq:nondegeneracy}, and the \emph{certificate of optimality} and the notion of \emph{estimation contracts} in related work~\cite{Yang22pami-certifiablePerception, Yang20tro-teaser, Carlone22arxiv-estimationContracts}. 
A certificate of optimality indicates whether the solver to optimal estimation problem returns the optimal solution or not.
Estimation contracts, on the other hand, ensure that the input is ``reasonable'' enough for the optimal solution to be indeed correct, \ie close to the ground-truth. 
While the former is analogous to our certificate of observable correctness (which indicates whether the output of a learning-based model is in the solution space or not), the latter is similar in spirit to the certificate of non-degeneracy. 
The difference is that our approach, unlike~\cite{Yang22pami-certifiablePerception, Yang20tro-teaser, Carlone22arxiv-estimationContracts}, aims at guaranteeing correctness of  learning-based models, rather than optimization-based estimators. 
\end{remark}

We apply this general theory to specify and solve the certifiable object pose estimation problem. In Section~\ref{sec:problem}, we see how the definition of \zetaCorrectness translates to the object pose estimation problem, thereby helping us to specify \emph{correctness} property. In Section~\ref{sec:architecture}, we show how the two certificates~\eqref{eq:correctness}-\eqref{eq:nondegeneracy} can be implemented to establish \zetaCorrectness. Finally, in Section~\ref{sec:training}, we will see how we use these certificates to provide the supervision in self-training.

\section{Certifiable and Self-Supervised \\ Object Pose Estimation}
\label{sec:problem}

\myParagraph{Certifiable Object Pose Estimation}   
Let \objectModel be the CAD model of an object, represented as the set of all points (expressed in homogeneous coordinates)
on the surface of the object.
Let $\inputPCfull$ be the corresponding \emph{posed} model, \ie the CAD model transformed according to its
ground-truth pose $\gtTransform \in \SEthree$:
\begin{equation}
\label{eq:gt}
\inputPCfull = \gtTransform \cdot \objectModel.
\end{equation}
Due to occlusions and sensor measurement noise, 
we only observe a partial and noisy point cloud \inputPC. This can be written as a function of the ground-truth posed model $\inputPCfull$:
\begin{equation}
\label{eq:generative-model}
\inputPC = \occFunction{\inputPCfull} + \noisePC,
\end{equation} 
where $\occFunction{\inputPCfull}$ 
 denotes the sampling of $\nrPoints$ points on the object surface and deletion of occluded parts, and $\noisePC$ is the measurement noise. 

The goal of certifiable pose estimation is to estimate the pose of the object, given the partial point cloud \inputPC, and also provide \emph{correctness guarantees} of the resulting pose estimate ---in terms of its closeness to the ground-truth pose \gtTransform. Inspired by the theory of certifiable models (Section~\ref{sec:framework}), we would like to implement a certifiable model $(\Model, \ocx, \ndx)$ that (i) estimates object pose $\predTransform = \Model(\inputPC)$, and (ii) determines if it is \zetaCorrect by checking whether $\oc{\Model, \inputPC} = \nd{\Model, \inputPC} = 1$.

In order to do this, we first instantiate the general definition of \zetaCorrectness to the problem of object pose estimation. Rather than defining the output space to be the space of all poses $\predTransform \in \SEthree$, we consider the output space to be the space of all posed CAD models, namely 
\begin{equation}
\outputSpace = \left\{ \MT \cdot \objectModel ~\Big|~\MT \in \SEthree \right\}.
\end{equation}
We define the distance between any two points in \outputSpace to be the Hausdorff distance:
\begin{equation}
\label{eq:hd}
\distHausdorff{\MU}{\MV} = \max\left\{ \sup_{\vxx \in \MU} D(\vxx, \MV), \sup_{\vxx \in \MV} D(\vxx, \MU)\right\},
\end{equation}
where $\MU, \MV \in \outputSpace$ and $D(\vxx, \MU)$ denotes the minimum distance from point $\vxx$ to the set $\MU$. Hausdorff distance is known to be a valid distance metric~\cite{Barich11project-hausdorffDist}. Thus, the results in Section~\ref{sec:framework} hold when determining correctness of the pose estimate \predTransform. The \zetaCorrectness (Definition~\ref{def:cert-correctness}) translates to the following definition for the object pose estimation problem:  
\begin{definition}[\zetaCorrectness for Object Pose Estimation]
\label{def:ep-cert-correct}
The pose estimate $\predTransform$, produced by a model, is \zetaCorrect if
\begin{equation}
\label{eq:def:ep-cert-correct}
\distHausdorff{\predTransform \cdot \objectModel}{\gtTransform \cdot \objectModel} \leq \zeta,
\end{equation}
where \distHausdorff{\cdot}{\cdot} is the Hausdorff distance.
\end{definition}

\begin{remark}[Hausdorff Distance \& Handling Symmetric Objects]
Intuitively, \zetaCorrectness of the estimate $\predTransform$ ensures that every point on the surface of the posed model $\predTransform \cdot \objectModel$ is at most $\zeta$ distance away from the surface of the ground-truth posed model $\gtTransform \cdot\objectModel$. 
Unlike rotation and translation error, the metric $\distHausdorff{\predTransform \cdot \objectModel}{\gtTransform \cdot \objectModel}$ operates directly on the CAD model surfaces. This obviates the need to handle symmetric objects separately: for instance, while two poses corresponding to symmetries of an object would produce different rotation and translation errors, the posed model associated to both poses will be the same, leading to the same Hausdorff distance for both. %
The ADD-S metric~\cite{Wang19cvpr-DenseFusion6D} has also been proposed to obviate the need for separately handling symmetric objects at evaluation. However, it operates on sampled point clouds and cannot satisfy the properties of a valid distance metric. In~\cref{app:sec:problem-statement} we discuss additional reasons for using Hausdorff distance as opposed to pose error or ADD-S metric~\cite{Wang19cvpr-DenseFusion6D} for theoretical analysis.
\end{remark}

The \zetaCorrectness of a pose estimate cannot be determined by directly computing~\eqref{eq:def:ep-cert-correct}, as we do not have access to the ground-truth pose in practice. A model solving the certifiable pose estimation problem should find a way to assert if~\eqref{eq:def:ep-cert-correct} is satisfied, without a genie-access to the ground truth, leading to the following problem statement.

\begin{problem}[Certifiable Object Pose Estimation]
\label{prob:cert-pe}
Propose a pose estimation model that, along with producing an estimate, can also 
provide a binary ($0/1$) \emph{certificate}, such that whenever a certificate is $1$, the 
estimated pose is \zetaCorrect (def.~\ref{def:ep-cert-correct}).
\end{problem}

\myParagraph{Unannotated Data and Self-Supervision} 
It is easy to realize that certification is tightly coupled with self-supervision: if we are able to discern ---at test time--- correct outputs, we can then use them for further training.
Therefore, in this paper we also study the twin problem of self-supervising pose estimation models.
In particular, we investigate learning models that can self-train using an unannotated real-world dataset, which consists of partial point clouds of objects, segmented from a scene; no pose or any other annotation is assumed on the real-world dataset.
We start by noting that annotated simulation data is broadly available  and can be used to initialize weights of any learning-based pose estimation model. %
Therefore, we consider the practical case where  we are given a 
sim-trained 
learning-based model, and our %
 goal is to modify and further train the model 
on the unannotated real-world data --- especially, when the sim-to-real gap is significant.
\begin{problem}[Self-supervised Pose Estimation]
	\label{prob:ss}
	Propose a method to modify a sim-trained 
	pose estimation model, 
	and train it with self-supervision  
	on unannotated real-world data.  
\end{problem}

\section{\name: a Certifiable Pose Estimation Model}
\label{sec:architecture}

This section provides a certifiable pose estimation model, named \name (\nameLong), that includes a novel architecture (mixing learning-based and optimization-based modules), and a practical implementation of the certificates  introduced in the previous section. Then, in~\cref{sec:training} we show that the use of the proposed certificates naturally enables  \name to self-train on unannotated data.

\begin{figure}[t]
	\centering
	\includegraphics[trim=170 140 185 138,clip,width=0.95\linewidth]{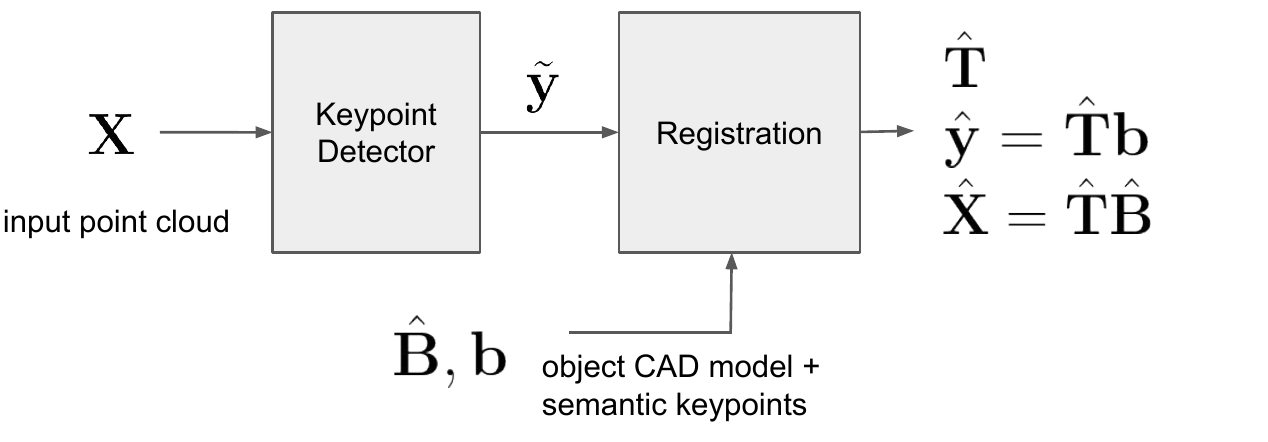}\vspace{-2mm}
	\caption{\KeyPo: Semantic keypoint-based pose estimation.}
	\label{fig:basic-architecture}
	\vspace{-0.4cm}
\end{figure}

\begin{figure}[t]
	\centering
	\includegraphics[trim=15 10 15 10,clip,width=0.85\linewidth]{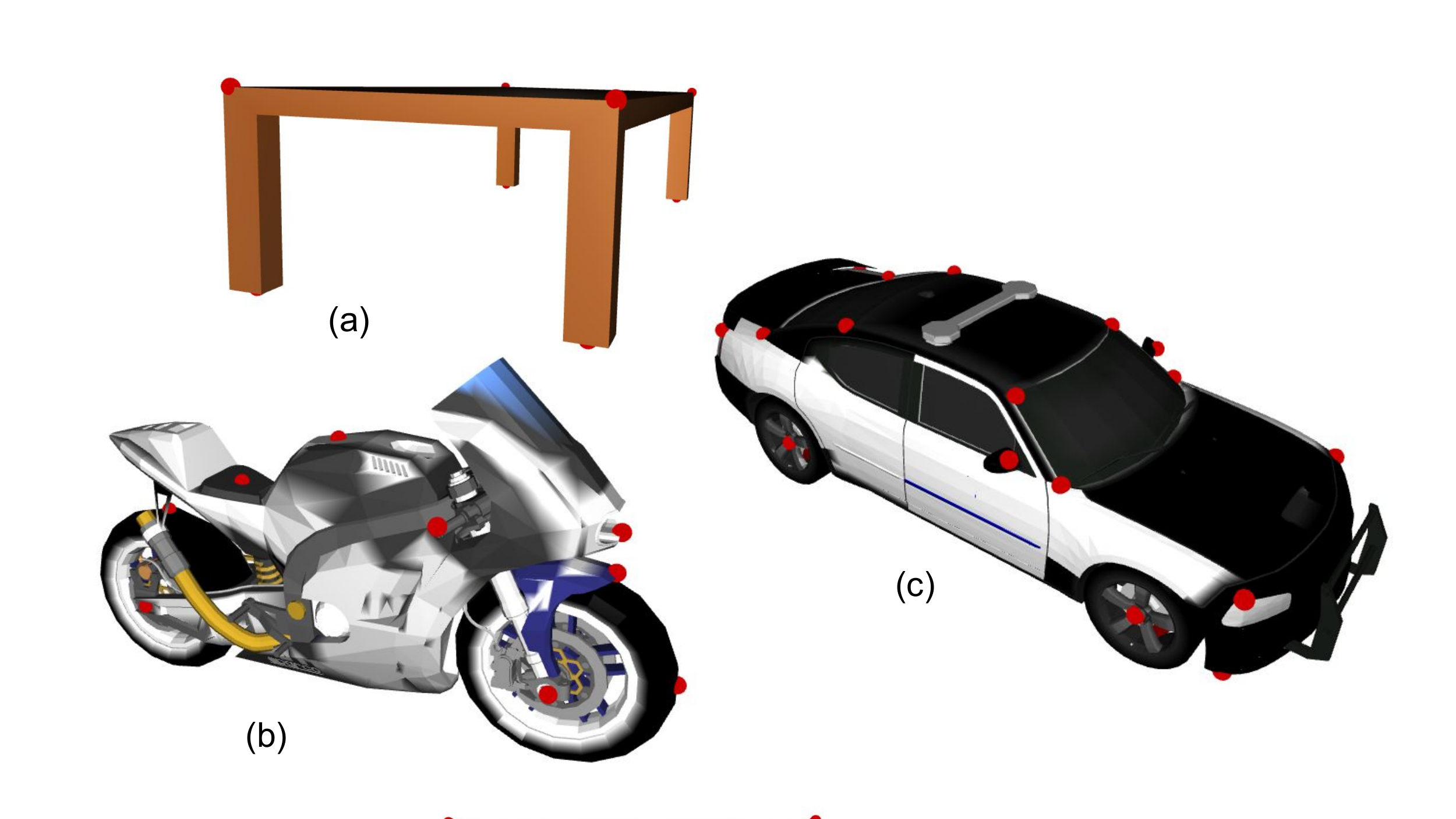}
	\caption{Annotation of semantic keypoints (red) on CAD models of (a) table, (b) motorcycle, and (c) car in KeypointNet~\cite{You20cvpr-KeypointNetLargescale}.}
	\label{fig:cad-models}
\vspace{-0.6cm}
\end{figure}

\subsection{Preliminaries: Semantic-Keypoint-based Pose Estimation (\KeyPo)}

We start by briefly reviewing a standard pose estimation method based on semantic keypoints (\cref{fig:basic-architecture}), 
which constitutes the backbone of \name.
 Semantic keypoints $\objectKeypoints[i], i=1,\ldots,\nrKeypoints$, are specific points annotated on the CAD model \objectModel, %
 and are fixed in number for a given object (see, \eg~\cref{fig:cad-models}).
 {
A standard semantic-keypoint-based pose estimator (\KeyPo) first detects semantic keypoints $\kpDetected[i], i=1,\ldots,\nrKeypoints$, from the input point cloud  using a neural network.
 Then, it estimates the object pose by solving an outlier-free registration problem:
\begin{equation}
\begin{aligned}
\predTransform =& \argmin_{\MT \in \SEthree} %
& & \textstyle\sum_{i=1}^{\nrKeypoints}~~\lVert \kpDetected[i] - \MT \cdot \objectKeypoints[i] \rVert_{2}^{2}, \label{eq:pose_est}
\end{aligned}
\end{equation}
that computes the object pose $\MT \triangleq \matTwo{\MR & \vt \\ \zero & 1}$ 
(where $\MR$ is the object rotation and $\vt$ is its translation) by aligning the 
detected keypoints $\kpDetected[i]$ and the model keypoints $\objectKeypoints[i]$, annotated on the CAD model. 
}
After computing the pose estimate $\predTransform$, we can also compute the \emph{posed} CAD model and keypoints, namely
 \begin{equation}
 \label{eq:predModel}
 (\predModelKP, \predModel) = ( \predTransform\cdot \objectKeypoints, \predTransform\cdot \objectModelPC)
 \end{equation}
  where 
  {$\objectModelPC$ denotes a point cloud obtained by densely sampling the CAD model~\objectModel.} %

\begin{remark}[Limitations of Standard Semantic-Keypoints-based Models and Novel Insights]
In the presence of keypoint detection errors (\eg caused by sim-to-real gap), 
the pose computed by \KeyPo might be inaccurate. Moreover, using only a sparse number of keypoints (typically in the order of tens) for pose estimation might lead to less accurate results.
For this reason, modern approaches either resort to neural models that mimic ICP~\cite{Aoki19cvpr-pointnetlk, Li21cvpr-PointNetLKRevisited}, approaches based on dense correspondences~\cite{Zhou16eccv-fastGlobalRegistration, Wang19ICCV-DeepClosestPoint, Wang19nips-prnet}, or approaches based on learned global representations~\cite{Yuan20eccv-DeepGMRLearning, Li21nips-LeveragingSE}. 
In this paper we provide two novel insights: (i) despite common beliefs, standard semantic-keypoint-based methods still outperform newer alternatives in the presence of occlusions and large object displacements (\cref{sec:keypo});
(ii) we can correct detection errors and factor dense information into the registration phase of semantic-keypoint methods by introducing a \emph{corrector} module, as discussed in the following section.
\end{remark}

\subsection{Overview of \name}
\begin{figure}
	\centering 
	\includegraphics[trim=110 170 165 90,clip,width=0.95\linewidth]{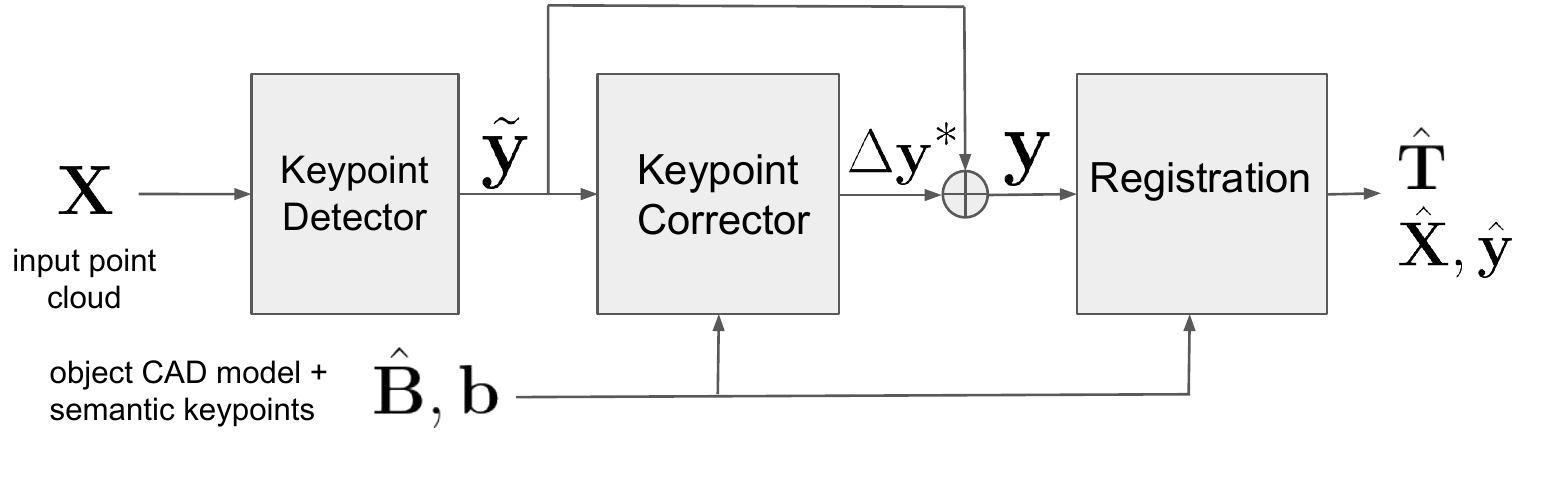}\vspace{-2mm}
	\caption{{\name: Proposed semantic-keypoint-based pose estimation and model fitting architecture with 
		\emph{corrector}.}
		\label{fig:proposed-model-corr} 
		\vspace{-5mm}}
\end{figure}

\begin{figure}
	\centering 
	\includegraphics[trim=110 160 165 100,clip,width=0.95\linewidth]{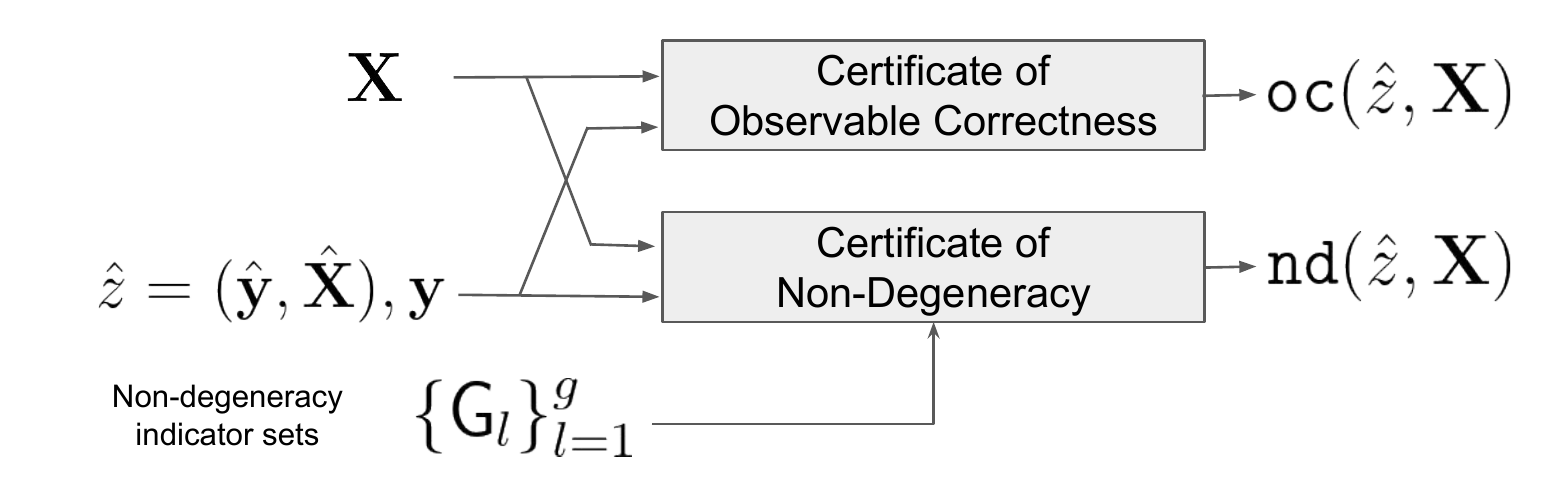}
	\caption{\name: Certificate of \observableCorrectness and non-Degeneracy. Here, $\left\{ \setG_l \right\}_{l=1}^{g}$ denote the collection of \emph{indicator sets} that are used to determine non-degeneracy; see~\eqref{eq:nondegeneracy}. \label{fig:proposed-model-cert} 
		\vspace{-4mm}
	}
\end{figure}

\name takes a partial or occluded object point cloud $\inputPC$ as input and
 estimates the object pose \predTransform, as well as a posed point cloud $\predModel$ and keypoints $\predModelKP$ as in~\eqref{eq:predModel}.
\name uses a standard semantic keypoint-based pose estimation architecture as the backbone, which first uses a neural network to detect semantic keypoints, and then estimates the 3D pose via registration to the corresponding CAD model. 
However, contrary to standard semantic keypoint-based methods, 
\name adds a keypoint \emph{corrector module} that corrects some of the keypoint-detection errors (\cf~\cref{fig:basic-architecture,fig:proposed-model-corr}).
Finally, \name implements two certificates to check \zetaCorrectness of the estimate (\cref{fig:proposed-model-cert}). %
Below, we provide an overview of the key components of \name, while we postpone the details to 
the following sections.

\myParagraph{Semantic Keypoint-based Pose Estimation} 
A semantic keypoint detection network first detects the semantic keypoints $\kpDetected[i], i=1,\ldots,\nrKeypoints$,  from the input point cloud $\inputPC$ (\cref{fig:proposed-model-corr}). 
We implement the keypoint detector in \KeyPo as a trainable regression model;
in our tests, we use PointNet++~\cite{Qi17nips-pointnet++} or point transformer~\cite{zhao20arxiv-PointTransformer} 
as a neural architecture that operates on point clouds. 
A regression model enables detecting semantic keypoints even in the occluded regions of the input \inputPC.

\myParagraph{Corrector} 
The semantic keypoint detector might produce perturbed keypoints when tested on real data 
(\eg if the detector is trained on a simulation dataset and there is a sim-to-real gap).
In \name, we add a \emph{corrector} module that takes the estimated semantic keypoints $\kpDetected$ ---produced by the keypoint detector--- as input, and outputs a correction term $\kpOptCorrection$ to the keypoints. 
The corrected keypoints become: %
\begin{equation}
\kpCorrected = \kpDetected + \kpOptCorrection.
\end{equation}
The resulting architecture is shown in~\cref{fig:proposed-model-corr}.
The correction term $\kpOptCorrection$ is obtained as a solution to the \emph{corrector optimization problem}, whose details are given in \cref{sec:corrector}. We implement the corrector as a differentiable optimization module, by explicitly computing its gradient. For faster training, we implement batch gradient descent that solves a batch of corrector optimization problems, in parallel, on GPU.

\myParagraph{Certification} \name implements the two certificates introduced in~\cref{sec:cert-model}, 
a certificate of \observableCorrectness and a certificate of non-degeneracy,
which together determine \zetaCorrectness of a pose estimate. While Theorem~\ref{thm:epsilon-cert-general-outer} provides a mathematical justification of implementing the two certificates, we describe what they intuitively mean in the case of object pose estimation.
The certificate of \observableCorrectness assesses whether the 
output of \name fits the input data; if it does then it is equivalent to saying that the ``output is in the solution space''.
The non-degeneracy certificate, on the other hand, ensures that the input point cloud $\MX$ provides sufficient information on the pose of the object to allow us to unambiguously estimate it.
\Cref{fig:expt_shapenet_degeneracy} shows an instance in which using just the input point cloud (shown in (a)), it is impossible to estimate the ground-truth pose.  
The key intuition in \name is that we can use the \emph{keypoints} to infer non-degeneracy: if we choose keypoints to cover the different parts of the object (\eg in~\cref{fig:expt_shapenet_degeneracy}, the keypoints capture the visor, the top button, and other parts of the cap), then as long as we detect specific subsets of keypoints on $\inputPC$ (\eg including the ones on the visor for the cap in~\cref{fig:expt_shapenet_degeneracy}), we can unambiguously estimate the object pose.
Since we design the two certificates to be conservative, we then use~\cref{thm:epsilon-cert-general-outer} to conclude that, when the certificates are $1$, the resulting estimate is \zetaCorrect.

\subsection{Keypoint Corrector}
\label{sec:corrector}

We now discuss the semantic keypoint corrector in the \name. The keypoint corrector adds a correction $\kpOptCorrection$ to the detected keypoints $\kpDetected$. 
This is intended to correct for any keypoint detection errors.
This section describes how we compute $\kpOptCorrection$, from the detected keypoints $\kpDetected$, and the implementation of the corrector as a differentiable optimization module.

First, we introduce some notation. Let $\predTransform(\kpCorrection)$
denote the optimal solution to the outlier-free registration problem in~\eqref{eq:pose_est}, but with the keypoints $\kpCorrected = \kpDetected + \kpCorrection$, instead of $\kpDetected$ in~\eqref{eq:pose_est}.  
For a given $\predTransform(\kpCorrection)$, let
\begin{equation}
\label{eq:outSpaceC3PO}
(\predModelKPCorrection, \predModelCorrection) \in \outputSpaceSampled,
\end{equation}
be the posed keypoints and CAD model, computed using~\eqref{eq:predModel}. 
Namely, $\predModelKPCorrection$ are the keypoints $\objectKeypoints$ transformed according to the pose
$\predTransform(\kpCorrection)$
and \predModelCorrection is the dense object point cloud \objectModelPC transformed by the same  transformation.
In~\eqref{eq:outSpaceC3PO}, we denoted with   \outputSpaceSampled the output space, \ie the set of all posed keypoints and 
sample CAD models.

\myParagraph{Corrector Optimization Problem} The correction $\kpCorrection$ is optimized so that the predicted object model $\predModelCorrection$ and model keypoints $\predModelKPCorrection$ are aligned to the input point cloud $\inputPC$ and the corrected keypoints $\kpCorrected$:

\begin{equation}
\begin{aligned}
& \underset{\kpCorrection \in \Real{3\times \nrKeypoints}}{\text{Minimize}} 
& & \text{ch}_{1/2}(\inputPC, \predModelCorrection) + \gamma || \kpCorrected - \predModelKPCorrection||^{2}_{2},  \label{eq:corrector}
\end{aligned}
\end{equation} 
where $\gamma$ is a positive constant and $\text{ch}_{1/2}(\inputPC, \predModel)$ is the half-Chamfer loss given by
\begin{equation}
\label{eq:half-Chamfer}
\text{ch}_{1/2}(\inputPC, \predModel) = \frac{1}{\nrPoints} \sum_{i \in [\nrPoints]} \min_{j \in [\nrModelPoints]} \lVert \inputPC[i] - \predModel[j] \rVert^{2}_{2}.
\end{equation}
The half-Chamfer loss  measures the average squared distance between each point in the input $\inputPC$ and the closest point in the estimated model $\predModel$.
The use of the half-Chamfer loss is motivated by the fact that the input point cloud $\inputPC$ is typically occluded, hence we are only interested in assessing how well the visible part of the object fits the posed model. %

\myParagraph{Forward Pass: Solving~\eqref{eq:corrector}} 
The corrector output \kpOptCorrection is the solution of the optimization problem~\eqref{eq:corrector}.
The optimization problem in~\eqref{eq:corrector} is non-linear and non-convex. A number of open-source solvers can be used to solve~\eqref{eq:corrector} in the forward pass. We tested different solvers and observed that two solvers work well, \ie they are able correct large errors in the keypoints. These are a simple constant-step-size gradient descent, which we implement in PyTorch, and a trust-region method~\cite{Conn00siam-trustRegMethodOpt}, implemented in the SciPy library~\cite{Virtanen20-scipy}.
In implementing these solvers, we need to obtain gradients of the objective function in~\eqref{eq:corrector}, which depends on the solution to the outlier-free registration problem~\eqref{eq:pose_est}. The solution to~\eqref{eq:pose_est} involves linear operations and SVD computation~\cite{Horn88, Arun87pami, wahba1965siam-wahbaProblem}, the derivative of which is computed using the autograd functionality in PyTorch. 
More specifically, we implement \emph{batch gradient descent} -- a simple, fixed-step-size gradient descent method implemented in PyTorch that solves~\eqref{eq:corrector} for a batch input on the GPU, and this leads to much faster compute times (\cref{pf:thm:corrector_bkprop}). 

\myParagraph{Gradient Computation for Back-Propagation}  
%
%
Implementing the corrector as a block in an end-to-end differentiable pipeline requires~\eqref{eq:corrector} to be differentiable, \ie there must be a way to compute $\partial \kpOptCorrection / \partial \kpDetected$.
We now show that the corrector optimization problem~\eqref{eq:corrector}, although being non-linear and non-convex, heeds to a very simple derivative (proof in Appendix~\ref{pf:thm:corrector_bkprop}). 
This allows us to implement the corrector as a differentiable optimization module in the \name architecture.
\begin{theorem}
	\label{thm:corrector_bkprop}
	The gradient of the correction $\kpOptCorrection$ with respect to the estimated keypoints $\kpDetected$ is the negative identity, i.e., 
	\begin{equation}
	\partial \kpOptCorrection / \partial \kpDetected = -\mathbf{I}.
	\end{equation}
\end{theorem}

\begin{remark}[Corrector vs. ICP] %
\label{rem:corr-or-icp}
The corrector is similar in spirit but computationally different from an 
ICP-based pose refinement. ICP-based methods tend to correct/refine the pose by optimizing over the object rotation or translation. The corrector, on the other hand, does so by optimizing over the keypoint space; see~\eqref{eq:corrector}.
In our experiments, 
we observe that correcting for pose over the space of keypoints is more powerful (\ie corrects larger errors) compared to doing so over the space of rotation and poses.
\end{remark}

\subsection{Certificates for Pose Estimation}
\label{sec:pose-certificates}
We now use the framework in \Sec~\ref{sec:cert-model} to derive two certificates for \name.
\name uses a dense point cloud representation \objectModelPC of the object \objectModel, to obtain the posed output \predModel; see~\eqref{eq:predModel}. While such a sampling is necessary for implementation (\eg to make it easy to compute distances between two posed models),  
 the definition of \zetaCorrectness (\cref{def:cert-correctness}) requires closeness between the CAD models $\predTransform \cdot \objectModel$ and $\gtTransform \cdot \objectModel$. %

To bridge this gap, 
we use two output spaces: (i) \outputSpaceSampled: the space of all outputs $(\predModelKP, \predModel)$ produced by \name; see~\eqref{eq:predModel}. This space is nothing but the space of all posed tuple $(\objectKeypoints, \objectModelPC)$ of model keypoints \objectKeypoints and sampled object CAD model \objectModelPC. (ii) \outputSpace: the space of all posed tuple $(\objectKeypoints, \objectModel)$:
\begin{equation}
\outputSpace = \left\{(\MT \cdot \objectKeypoints, \MT \cdot \objectModel)~\Big|~\MT \in \SEthree \right\},
\end{equation}
where now $\MB$ is the full CAD model.
%
%
%
\begin{remark}
	The choice of \outputSpace (or \outputSpaceSampled) to be the space of all posed keypoints and models, and not just the space of all poses (\ie \SEthree) circumvents the need to consider symmetric objects as a special case, and helps our analysis (Appendix~\ref{app:sec:pose-certificates}).
\end{remark}

\myParagraph{Certificate of Observable Correctness}
We conservatively approximate the observable correctness certificate by checking the geometric consistency between the input point cloud \inputPC and the output $\outputPred = \posedOutputPred$ in~\eqref{eq:predModel}. Since the input point cloud is partial, we only check if every point in the partial input \inputPC is close to a point in \predModel:
\begin{equation}
\label{eq:certificate-correctness}
\oc{\outputPred, \inputPC} = \indicator{\max_{i \in [\nrPoints]} \min_{j \in [\nrModelPoints]} \lVert \inputPC[i] - \predModel[j] \rVert_2 < \epsilon_\ocx},
\end{equation}
where $\epsilon_\ocx$ is a small positive constant. We show that this certificate can be derived as an outer approximation certificate when the noise \noisePC in~\eqref{eq:generative-model} is bounded, and the point cloud representation \objectModelPC of \objectModel is sufficiently dense (Appendix~\ref{app:sec:pose-oc-outer}).

We note that while the certificate~\eqref{eq:certificate-correctness} is written as a function of the output \outputPred and the input \inputPC, it can repurposed to the form $\oc{\Model, \inputPC}$ (in Section~\ref{sec:framework}) by setting $\outputPred = \Model(\inputPC)$.

\myParagraph{Certificate of Non-degeneracy}
Non-degeneracy depends on the size of the solution space \solSpace{\inputPC} (see~\eqref{eq:nondegeneracy}). Intuitively, the solution space is likely to be large if the input point cloud is highly occluded or is missing important features or parts of the object.
Consider the case of an otherwise symmetric mug, with a handle. If the input point cloud \inputPC misses the handle entirely, there will be multiple ways to register the CAD model to the input -- thereby, making \solSpace{\inputPC} large.

Motivated by this observation, we define the notion of a \emph{cover set} of points on a CAD model.

\begin{definition}[Cover Set]
A subset of points $\objectModelSubset$ on the surface of a CAD model $\objectModel$ is a \emph{cover set} if, given the 3D positions of points in \objectModelSubset, we can unambiguously estimate (up to noise) the pose of $\objectModel$: 
\begin{equation}
\label{eq:cover-set-condition}
\sup_{\vxx \in \aTransform' \cdot \objectModelSubset}\distSurf{\vxx}{\aTransform \cdot \objectModel} < \delta \implies \distHausdorff{\aTransform' \cdot \objectModel}{\aTransform \cdot \objectModel} < \delta,
\end{equation}
for any $\aTransform, \aTransform' \in \SEthree$ and %
where $\delta$ is a positive constant. 
\end{definition}
Eq.~\eqref{eq:cover-set-condition} states that if the partial point cloud \objectModelSubset of the object \objectModel is such that every point in $\aTransform' \cdot \objectModelSubset$ is close in distance to the posed model $\aTransform \cdot \objectModel$, then it will necessitate that that two posed models $\aTransform  \cdot \objectModel$ and $\aTransform' \cdot\objectModel$, are in fact very closely aligned.  
This implies that if our input point cloud \inputPC is large enough to encompass at least a cover set of the object, the size of the solution space must be small; implying non-degeneracy~\eqref{eq:nondegeneracy}. 

Our insight is that we can use semantic keypoints, intelligently annotated, to check for the size of the solution space, and if the observed \inputPC does indeed encompass a cover set. In our mug example, if we are able to detect a few keypoints on the mug handle, such that those points are also close to the input point cloud, then we can conclude that the input is non-degenerate. This leads to the notion of \emph{indicator set}.

\begin{definition}[Indicator Set]
	\label{def:indicator-set}
Let $\occFunction{\objectModel}$ be a sampled and visible portion of an object $\objectModel$ as in~\eqref{eq:generative-model}.
A subset of keypoints $\setG \subset [\nrKeypoints]$ is an \emph{indicator set} if when all keypoints $\{ \objectKeypoints[i] \}_{i \in \setG}$ are close to $\occFunction{\objectModel}$ by distance \deltaIndicator, \ie 
\begin{equation} 
\min_{j \in [\nrPoints]} \lVert \objectKeypoints[i] - \occFunction{\objectModel}[j] \rVert_{2} < \deltaIndicator~\forall~i \in \setG, 
\end{equation} 
then \occFunction{\objectModel} is a cover set for \objectModel. 
\end{definition}

Let $\left\{\setG_l \right\}_{l=1}^{g}$ be a collection of such indicator sets for the object.  
Given an input point cloud \inputPC and an output $\outputPred = \posedOutputPred \in \outputSpaceSampled$, we declare non-degeneracy if ---for at least one set $\setG_l$, $l \in [g]$---
every keypoint in $\predModelKP[i]$ in $\setG_l$
is close to a point in the input \inputPC.
Mathematically, this is given by: 
\begin{equation}
\label{eq:certificate-nondegen}
\nd{\outputPred, \inputPC}  = \indicator{
	\bigcup_{l=1}^{g} \bigcap_{i \in \setG_l} \left\{   \min_{j \in [\nrPoints]} \lVert \predModelKP[i] - \inputPC[j] \rVert_2  < \delta_\ndx \right\}},
\end{equation}
where $\delta_\ndx$ is a small positive constant.

\myParagraph{Certification Guarantees} We now prove that using the two certificates~\eqref{eq:certificate-correctness}-\eqref{eq:certificate-nondegen}, we can determine whether an estimate produced by the model \Model is \zetaCorrect (\cref{def:ep-cert-correct}), under some reasonable assumptions. The proof is given in Appendix~\ref{app:sec:pose-certificates}.  
\begin{assumption}[Bounded Noise]
	\label{as:bounded-noise}
	The noise~\eqref{eq:generative-model} is bounded, namely $\max_{i} \lVert \noisePC[i] \rVert_2 < \epsilon_w$.
\end{assumption}
\begin{assumption}[Dense CAD Model Sampling]
	\label{as:cad-model-sampling}
	The sampled point cloud \objectModelPC is such that every point on \objectModel is at most a distance $\epsilon_s$ away from a point in \objectModelPC, namely $\min_{j \in [\nrModelPoints]} \lVert \vxx - \objectModelPC[j] \rVert_2 < \epsilon_s$ for all $\vxx \in \objectModel$.
\end{assumption}
\begin{theorem}
\label{thm:pose-cert-guarantee}
Assume bounded noise, dense CAD model sampling (Assumptions~\ref{as:bounded-noise}-\ref{as:cad-model-sampling}), and that 
the sets $\{ \setG_l \}_{l=1}^{g}$ in~\eqref{eq:certificate-nondegen} are indicator sets.
Then, the output produced by a pose estimator \Model: 
$\outputPred = \posedOutputPred = \Model(\inputPC)$
is \zetaCorrect (\cref{def:ep-cert-correct}),
if 
$\oc{\outputPred, \inputPC}=\nd{\outputPred, \inputPC}=1$ 
in~\eqref{eq:certificate-correctness}, \eqref{eq:certificate-nondegen} for any $\zeta > \epsilon_\ocx + \epsilon_w$, provided $\epsilon_\ocx + \delta_\ndx + 2\epsilon_w < \deltaIndicator$. 
\end{theorem}

\begin{remark}[Choosing Indicator Sets]
The non-degeneracy certificate in~\eqref{eq:certificate-nondegen} provably works (Theorem~\ref{thm:pose-cert-guarantee}) when there exists indicator sets, among the annotated keypoints. %
Deriving indicator sets to provably meet its definition is hard and we do not choose that route. We, instead, hand-craft the indicator sets and show its usefulness towards correctly certifying \name's pose estimates. 
We show, in our experiments, that a simple choice of subsets (listed in Appendix~\ref{app:expt_shapenet_ycb}) suffices to determine non-degeneracy.
\end{remark}

\section{Self-Supervised Training}
\label{sec:training}

The certificates implemented in \name enable a simple yet effective self-supervised training procedure.
According to the setup described in~\cref{prob:ss}, we assume that 
the real-world, training dataset $\calD$ consists of a collection of input point clouds \inputPC, bearing no pose or keypoint annotation. The only trainable part in our \name architecture is the keypoint detector, which is initialized to a sim-trained model. In the self-supervised training, it is the keypoint detector that gets trained to better detect semantic keypoint on real-world data. The other components in \name (\ie corrector and certificates) enable this self-supervised training.

\myParagraph{Certificate-based Self-Supervised Training} Our self-supervised training is the same as a standard supervised training using stochastic gradient descent (SGD), except from two changes. 
First, at each iteration, when we see a batch of inputs $\{\inputPC^i\}$, we compute the output $\{(\predModelKP^i, \predModel^i)\}$ using the current model weights. We then compute the loss:
\begin{equation}
\label{eq:pair_training_loss}
\mathcal{L}_i = \text{ch}_{1/2}(\inputPC^i, \predModel^i) + \theta \sum_{j=1}^{\nrKeypoints} \lVert \kpCorrected^i[j] - \predModelKP^i[j] \rVert^{2}_{2},
\end{equation}
for each input-output pair $i$ in the batch;
where $\text{ch}_{1/2}$ is the half-Chamfer loss, $\{\kpCorrected^i\}_i$ are the corrected keypoints, and $\theta$ is a positive constant. 

Second, at each iteration, we also determine the observable correctness of each input-output pair, by computing certificates in~\eqref{eq:certificate-correctness}: $\ocx_i = \oc{(\predModelKP^i, \predModel^i), \inputPC^i}.$

We then compute the total training loss for the batch ---that we use for back-propagation--- as
\begin{equation}
\label{eq:tot_training_loss}
\mathcal{L}  = \sum_i \ocx_i \cdot \mathcal{L}_i,
\end{equation} 
which is nothing but the sum-total of loss~\eqref{eq:pair_training_loss} computed only for the observably correct input-output pairs, in the batch.

\begin{remark}[Role of Certification]
	Self-supervised training using the total loss $\calL = \sum_i \calL_i$, \ie using all and not just the observably correct instances, would not work well in practice. This is because, some of the predicted models $\predModel$ are not correctly registered to the input point clouds $\inputPC$. Including them in the loss function~\eqref{eq:tot_training_loss} induces incorrect supervision causing the self-supervised training to fail. 
	Certification, on the other hand, weeds out incorrectly registered object models during the self-supervised training and provides correct supervision to the keypoint detector during training. 
	In all our experiments we observe  that as our self-supervised training progresses, the fraction of observably correct instances increases, eventually converging to nearly 100\% of the model outputs being observably correct (see~\cref{fig:expt_shapenet_cert}, in~\cref{sec:expt_ycb}).
\end{remark}

%
%
%
%
%
%

%
%
%
%

%
%

\begin{figure*}[t]
	\vspace{-3mm}
	\centering
	\begin{subfigure}{.33\textwidth}
		\centering 
		\includegraphics[trim=10 0 0 0,clip,width=\linewidth]{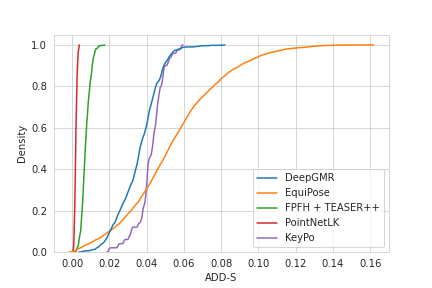}  
		\caption{Easy + Full PC\vspace{-1mm}}
	\end{subfigure}%
	\begin{subfigure}{.33\textwidth}
		\centering
		\includegraphics[trim=10 0 0 0,clip,width=\linewidth]{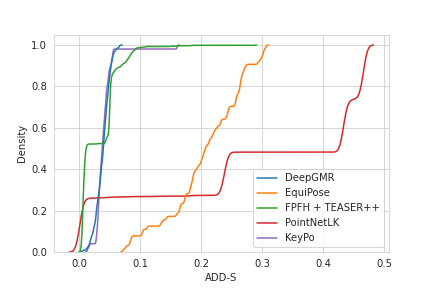} 
		\caption{Hard + Full PC\vspace{-1mm}}
	\end{subfigure}
	\begin{subfigure}{.33\textwidth}
		\centering
		\includegraphics[trim=10 0 0 0,clip,width=\linewidth]{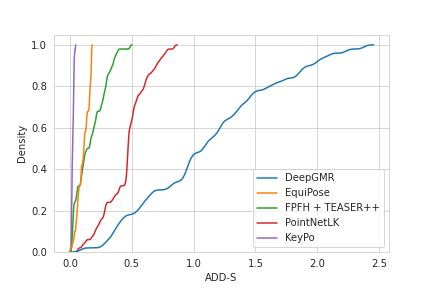} 
		\caption{Hard + Depth PC\vspace{-1mm}}
	\end{subfigure}
	\caption{Shows the distribution of ADD-S scores~\cite{Wang19cvpr-DenseFusion6D} for object pose estimation of a ShapeNet object car. Shows various baselines on three datasets: (a) Easy + Full PC, (b) Hard + Full PC, and (c) Hard + Full PC.}
	\label{fig:baseline_analysis}
	\vspace{-4mm}
\end{figure*}

\section{Experiments}
\label{sec:experiments}
We present five experiments. 
We first show that a standard semantic-keypoint-based architecture outperforms more recent alternatives in problems with partial point clouds and large object displacements (\cref{sec:keypo}).
We then analyze the ability of the corrector to correct errors in the keypoint detections (\cref{sec:expt_corrector_analysis}). We show the effectiveness of our self-supervised training and certification on a depth-point-cloud dataset generated using ShapeNet~\cite{Chang15arxiv-shapenet} objects (\cref{sec:expt_shapenet}) and on the YCB dataset~\cite{calli15-YCBobject} (\cref{sec:expt_ycb}); here we observe that \name significantly outperforms all the baselines and state-of-the-art approaches. 
Finally, we show that our self-supervised training method can work even when the training data has no object category labels on the input point clouds (\cref{sec:expt_fullss}). 

\begin{figure}[b]
	\centering
	\vspace{-5mm}
	\begin{subfigure}{.25\textwidth}
		\centering
		\includegraphics[trim=10 0 0 0,clip,width=\linewidth]{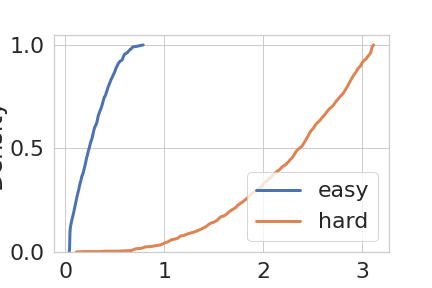} 
		\caption{Rotations\vspace{-1mm}}
	\end{subfigure}%
	\begin{subfigure}{.25\textwidth}
		\centering
		\includegraphics[trim=10 0 0 0,clip,width=\linewidth]{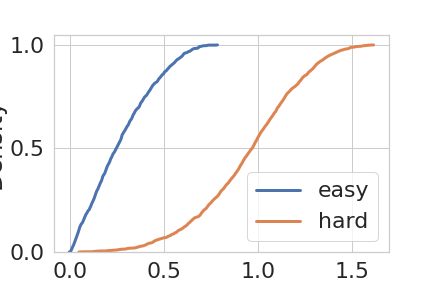} 
		\caption{Translations\vspace{-1mm}}
	\end{subfigure}
	\caption{Cumulative distribution function of object rotations and translations for easy/hard test dataset. The object rotations are in radians (rotation angle around a randomly chosen rotation axis), while the translations is measured as the norm of the translation vector normalized by the object diameter.}
	\label{fig:pose-err}
	\vspace{-1mm}
\end{figure}
\subsection{Why Do We Use Semantic Keypoints?}
\label{sec:keypo}

We start by providing an experimental analysis that justifies 
our choice of \KeyPo as the backbone for \name. This choice seems to go against the commonplace belief that 
neural models that mimic the ICP algorithm~\cite{Aoki19cvpr-pointnetlk, Li21cvpr-PointNetLKRevisited}, or  based on learned global representations~\cite{Yuan20eccv-DeepGMRLearning, Li21nips-LeveragingSE} outperform keypoint-based approaches. We do not consider dense correspondence-based methods like~\cite{Wang19ICCV-DeepClosestPoint, Wang19nips-prnet, Zhou16eccv-fastGlobalRegistration} as they are shown to be worse than PointNetLK~\cite{Aoki19cvpr-pointnetlk, Li21cvpr-PointNetLKRevisited} and DeepGMR~\cite{Yuan20eccv-DeepGMRLearning}.
This section empirically shows that this conclusion is only true in relatively easy problem instances, while keypoint-based approaches still constitute the go-to solution for hard problems with occlusions and arbitrary object poses. 

\myParagraph{Setup} We consider the following approaches: (i) FPFH + TEASER++~\cite{Rusu09icra-fast3Dkeypoints, Yang20tro-teaser}: a local-feature-based pose estimation method, that extracts local features and computes a pose estimate via robust registration; (ii) PointNetLK~\cite{Li21cvpr-PointNetLKRevisited}: a learning-based model that attempts to ``neuralize'' iterative closest point and is considered a state-of-the-art approach for point cloud alignment and pose estimation; (iii) DeepGMR~\cite{Yuan20eccv-DeepGMRLearning}: a learning-based model that attempts to extract deep features, by modeling them as latent Gaussian variables; the method is also known to be competitive to PointNetLK; (iv) EquiPose~\cite{Li21nips-LeveragingSE}: a learning-based model that attempts to solve shape completion, pose estimation, and in-category generalization; and (v) \KeyPo: a standard semantic keypoint-based pose estimation architecture (\cref{fig:basic-architecture}) {with a point transformer~\cite{zhao20arxiv-PointTransformer} as a keypoint detector.} %

 We analyze these models under two experimental settings, named \emph{easy} and \emph{hard}. In the \emph{easy} case, we consider relatively small rotations and translations of the objects, while the \emph{hard} case we induce larger rotations and translations. 
The cumulative distributions of the object rotations and translations are visualized in~\cref{fig:pose-err}. Since several methods tend to perform well when the input \inputPC is a full point cloud, as opposed to a depth or a partial point cloud, we  consider the following scenarios: (i) Easy + Full PC, where the object displacement is small and the input point cloud is the full point cloud, (ii) Hard + Full PC, where the object displacement is large and the input point cloud is the full point cloud, and (iii) Hard + Depth PC, where the object displacement is large, and the input point cloud is partial and computed from the 
rendered depth image of the object.

\myParagraph{Results}
\Cref{fig:baseline_analysis} shows the distribution of the {ADD-S scores}~\cite{Wang19cvpr-DenseFusion6D} for object pose estimation of a ShapeNet car object and for all the techniques  above.
From the figure, we note that the performance of all the approaches, except \KeyPo, degrades as we go from easy to hard test setups and from full point clouds to depth point clouds. PointNetLK, while performing extremely well on the Easy + Full PC dataset degrades significantly when the object rotations and translations are large. DeepGMR, while still performing well with large pose magnitudes, shows significant performance degradation when the input point cloud is a depth point cloud. EquiPose, while generalizing better, shows consistent sub-optimal performance.
 In particular, EquiPose seems to trade off its pose estimation accuracy for the objective of in-category generalization, and we consistently see the object shape/size not exactly matching the input. Like EquiPose, the local feature-based method FPFH + TEASER++ also generalizes well, but 
 are still outperformed by \KeyPo on the Hard + DepthPC dataset.

\myParagraph{Insights} We gather the following insights from the above analysis. Methods attempting to ``neuralize'' ICP, for solving object pose estimation, can yield optimal performance under small rotations and translation, while yielding sub-optimal performance when the object displacement is large; this is not dissimilar from ICP, which works well when initialized close to the ground-truth pose, but is prone to converge to local minima otherwise.
Methods like DeepGMR, that extract learned representations, while succeeding when the input is a full point cloud, fail when the input is occluded.
The performance gap between \KeyPo and DeepGMR (on Hard + Depth PC) indicates that {we are better off using semantic keypoints as a learned representation for the task of pose estimation.}

\begin{figure*}[h]
		\begin{tabular}{ccc}
			\includegraphics[trim=15 1 50 1,clip,width=0.31\linewidth]{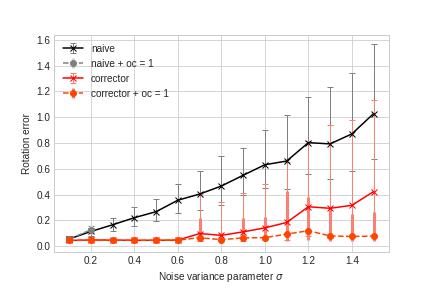}
			&   %
			\includegraphics[trim=15 1 50 1,clip,width=0.31\linewidth]{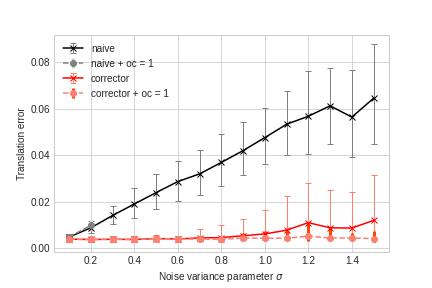}
			&   %
			\includegraphics[trim=15 1 50 1,clip,width=0.31\linewidth]{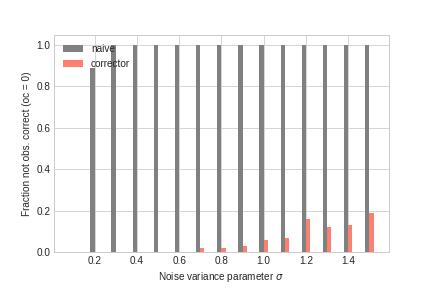} 
		\end{tabular}\vspace{-3mm}
		\caption{Rotation error (left), translation error (mid), and  fraction of \emph{not} observably correct instances ($\ocx = 0$) (right), as a function of the noise  parameter $\sigma$ for the ``chair''object in ShapeNet. The rotation error is in radians and the translation error is normalized by the object diameter.}
		\label{fig:corrector_analysis_chair}
\end{figure*}

\subsection{Keypoint Corrector Analysis}
\label{sec:expt_corrector_analysis}
This section shows that the keypoint corrector module is 
able to correct large errors in the keypoint detections. %

\myParagraph{Setup} We consider ShapeNet objects~\cite{Chang15arxiv-shapenet} and use the semantic keypoints labeled by the KeypointNet dataset~\cite{You20cvpr-KeypointNetLargescale}. Give an object model and semantic keypoints $(\objectKeypoints, \objectModel)$, 
we extract depth point cloud of $\objectModel$ from a certain viewing angle, transform $(\objectKeypoints, \objectModel)$ by a random pose, and add perturbations to the keypoints. 
The induced pose error is the same as in the \emph{hard} case described in~\cref{sec:keypo}.
For each keypoint $\objectKeypoints[i]$, with probability $f$, we add uniform noise distributed in $[-\sigma d / 2, \sigma d / 2]$, and with probability $1-f$, keep $\objectKeypoints[i]$ unperturbed. Here, $d$ is the diameter of the object. We set $f=0.8$. 
We study the performance of the corrector as a function of the noise variance parameter $\sigma$.

\begin{table*}[t]
	\vspace{-1mm}

\centering
\caption{Evaluation of \name and baselines for the ShapeNet experiment.}
\label{tab:expt_shapenet}

\begin{tabular}{|l|rr|rr|rr|rr|rr|rr|}
	\toprule
	ADD-S~ADD-S (AUC) &\multicolumn{2}{c}{car}  &\multicolumn{2}{c}{chair}  &\multicolumn{2}{c}{helmet}  &\multicolumn{2}{c}{laptop}  &\multicolumn{2}{c}{skateboard}  &\multicolumn{2}{c|}{table}  \\
	\midrule
	KeyPo (sim) & $0.00$ & $0.00$  & $0.00$ & $12.01$  & $0.00$ & $2.00$  & $0.00$ & $0.00$  & $4.00$ & $11.30$  & $0.00$ & $3.13$    \\
	KeyPo (sim) + ICP & $0.00$ & $0.00$  & $0.00$ & $11.56$  & $0.00$ & $2.00$  & $2.00$ & $2.00$  & $2.00$ & $9.80$  & $0.00$ & $0.00$    \\
	KeyPo (sim) + Corr. & $0.00$ & $13.69$  & $68.00$ & $57.16$  & $10.00$ & $20.14$  & $26.00$ & $20.57$  & $70.00$ & $59.17$  & $80.00$ & $53.54$    \\
	C-3PO & $80.00$ & $64.00$  & $100.00$ & $79.06$  & $100.00$ & $70.49$  & $\mathbf{100.00}$ & $\underline{64.70}$  & $\mathbf{100.00}$ & $\underline{82.90}$  & $\underline{98.00}$ & $\underline{67.69}$    \\
	C-3PO (oc=1, nd=1) & $\mathbf{100.00}$ & $\mathbf{84.59}$  & $\mathbf{100.00}$ & $\mathbf{90.13}$  & $\mathbf{100.00}$ & $\underline{71.53}$  & $\mathbf{100.00}$ & $\underline{64.70}$  & $\mathbf{100.00}$ & $\underline{82.90}$  & $\mathbf{100.00}$ & $\mathbf{69.07}$    \\
	DeepGMR & $0.00$ & $2.00$  & $0.00$ & $0.00$  & $0.00$ & $0.00$  & $0.00$ & $0.00$  & $0.00$ & $0.00$  & $0.00$ & $0.00$    \\
	PointNetLK & $0.00$ & $2.76$  & $6.00$ & $7.06$  & $0.00$ & $0.00$  & $0.00$ & $0.00$  & $2.00$ & $5.14$  & $4.00$ & $5.27$    \\
	FPFH + TEASER++ & $26.00$ & $25.03$  & $26.00$ & $25.65$  & $50.00$ & $46.29$  & $42.00$ & $32.32$  & $38.00$ & $40.33$  & $48.00$ & $39.23$    \\
	EquiPose & $16.00$ & $19.69$  & $14.00$ & $20.33$  & -- &  --  & -- &  --  & -- &  --  & -- &  --    \\
	KeyPo (real) & $\mathbf{100.00}$ & $\underline{79.08}$  & $\mathbf{100.00}$ & $\underline{79.92}$  & $\mathbf{100.00}$ & $\mathbf{80.28}$  & $\mathbf{100.00}$ & $\mathbf{83.11}$  & $\mathbf{100.00}$ & $\mathbf{84.02}$  & $54.00$ & $44.67$    \\
	\bottomrule
\end{tabular} 	\vspace{-3mm}
\end{table*}

\myParagraph{Results} \Cref{fig:corrector_analysis_chair}(left, mid) show the rotation and translation error as a function of the noise parameter $\sigma$; for the ShapeNet object chair. We compare the output of the corrector + registration with a naive method that applies only the registration block (eq.~\eqref{eq:pose_est}) to the detected/perturbed keypoints~$\kpDetected$. Not all outputs produced by these two models ---naive and corrector--- are observably correct per eq.~\eqref{eq:certificate-correctness}. 
The figures also plot the rotation and translation errors for the observably correct instances. 
\Cref{fig:corrector_analysis_chair}(right) shows the fraction of \emph{non}-observably correct outputs produced by the 
two methods.
We conduct similar analysis for several other ShapeNet objects in~\arxivadd{\cref{app:expt_corrector_analysis}}\arxivomit{\cite{Talak23arxiv-c3po}}.

\myParagraph{Insights} 
We observe that the corrector significantly improves the rotation and translation error, viz-a-viz the keypoint registration used in standard keypoint-based methods. 
A further performance boost is seen when we inspect the observably correct, \ie $\ocx=1$, instances produced by the corrector.
This boost is not only in the absolute reduction of the two error metrics, but also in the reduction of the error variance.
We observe a near-constant error, \wrt increasing keypoint noise $\sigma$, for the observably correct instances produced by the corrector.
Finally, we note that the fraction observably correct remains significantly high, when using the corrector.
For instance, more than $90\%$ of the outputs produced by the corrector remain observably correct, for $\sigma < 1$, whereas the naive method fails to produce any observably correct output for $\sigma \geq 0.4$.

Note that the purpose of the corrector module in \name is to ensure that there is a non-negligible fraction of input-output instances that are observably correct for self-supervised training, thus surmounting the sim-to-real gap. This analysis indicates that the corrector can indeed help bridge the sim-to-real gap. 

\subsection{The ShapeNet Experiment}
\label{sec:expt_shapenet}

This section demonstrates our self-supervised training on depth-point-cloud data generated using ShapeNet~\cite{Chang15arxiv-shapenet} objects.

\myParagraph{Setup} 
ShapeNet~\cite{Chang15arxiv-shapenet} classifies objects into 16 categories. We select one object in each category and use the semantic keypoints labeled by the KeypointNet dataset~\cite{You20cvpr-KeypointNetLargescale}. We use uniformly sampled point clouds of these 16 objects as the simulation dataset, and a collection of depth point clouds, rendered using Open3D~\cite{Zhou18arxiv-open3D}, as the real-world dataset. This choice ensures a large sim-to-real gap, and enables us to showcase the utility of our self-supervised training.
 We consider object displacements corresponding to the \emph{hard} setup described in~\cref{sec:keypo}.
We initially train the detector using the simulation dataset, and self-train it on the depth point cloud dataset, %
as discussed in~\cref{sec:training}. 
Hyper-parameter tuning and the implementation of the non-degeneracy certificates is discussed in~\cref{app:expt_shapenet_ycb}.

We report the following baselines in~\cref{tab:expt_shapenet}: (i) \KeyPo (sim):  a simulation-trained keypoint detector \KeyPo; (ii) \KeyPo (sim) + ICP: the simulation-trained keypoint detector, with iterative closest point refinement using the input \inputPC and the predicted point cloud $\predModel$; (iii)  \KeyPo (sim) + corrector, and (iv) \KeyPo (real): the keypoint detector trained on the real-wold dataset of depth point clouds, with full supervision. \KeyPo (real), therefore, marks an upper-bound on the performance of any self-supervised method. 
\name denotes the proposed method, obtained after the self-supervised training on real-data.
We use the point transformer regression model~\cite{zhao20arxiv-PointTransformer} for the keypoint detector in \KeyPo and \name.
We also compare the performance of \name against: (i) DeepGMR, (ii) PointNetLK, (iii) EquiPose: only in cases where the trained model is available, and (iv) FPFH + TEASER++. 
\Cref{tab:expt_shapenet} shows the performance on 6 of the 16 objects. 
The remaining objects are 
reported in~\arxivadd{\cref{app:expt_shapenet_ycb}}\arxivomit{\cite{Talak23arxiv-c3po}}.
We evaluate the performance of each model using threshold ADD-S and ADD-S (AUC) score~\cite{Wang19cvpr-DenseFusion6D}. 
We choose thresholds for the ADD-S and ADD-S (AUC) to be $5\%$ and $10\%$ of the object diameter $d$.

\begin{figure}
	\centering 
	\includegraphics[trim=25 0 43 35,clip,width=0.65\linewidth]{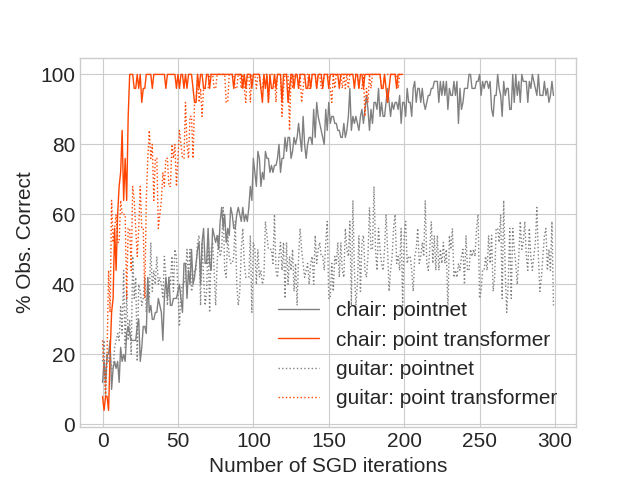}
	\caption{Percentage of observably correct instances~\eqref{eq:certificate-correctness} vs. number of SGD iterations during the proposed self-supervised training (\cref{sec:training}).
		\label{fig:expt_shapenet_cert} \vspace{-5mm}
	}
\end{figure}

\myParagraph{Results}
\Cref{fig:expt_shapenet_cert} sheds light on the performance of the proposed self-supervised approach and
shows how the number of observably correct instances in \name increases with the number of SGD iterations during self-supervised training, reaching close to 100\% in most cases.
In~\cref{fig:expt_shapenet_cert}, there is a noticeable difference in convergence time when the keypoint detector is modeled as a point transformer regression model vs. PointNet++ regression model. The point transformer regression model is consistently better than the PointNet++ across all objects (see~\cite{Talak23arxiv-c3po}). 

 \cref{tab:expt_shapenet} shows that \KeyPo (sim) performs very poorly and is not helped much by ICP; thereby, confirming the large sim-to-real gap in the experiment.
\KeyPo (sim) + Corr., on the other hand, shows a significant performance improvement in~\cref{tab:expt_shapenet}; confirming that the corrector indeed helps bridge the sim-to-real gap, to an extent. 
We, therefore, deduce that correcting for pose over the space of keypoints is more powerful than doing so over the space of rotation and poses, which is what the ICP attempts, thus validating~\cref{rem:corr-or-icp}.

\begin{figure}
	\centering
	\includegraphics[trim=10 0 35 35,clip,width=0.8\linewidth]{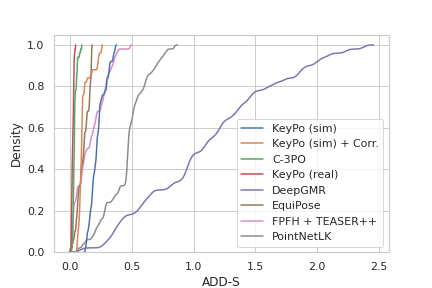}
	\caption{Distribution of ADD-S score for \name and a few baselines in the ShapeNet Experiment.}
	\label{fig:shapenet.real.hard.png}
	\vspace{-3mm}
\end{figure}

We observe that \name is able to significantly outperform all the baselines, and match the performance of the fully supervised \KeyPo (real). A further performance boost is attained by evaluating only the observably correct and/or non-degenerate outputs produced. We see this in~\cref{tab:expt_shapenet}, as well as in \Cref{fig:shapenet.real.hard.png}, which shows the distribution of ADD-S scores, over the test dataset.
The reasons for very low scores of the baselines such as DeepGMR, PointNetLK, FPFH + TEASER++, and EquiPose remain the same as discussed in~\cref{sec:keypo}.

\myParagraph{Degeneracy} Degeneracy arises when the depth point cloud of an object is severely occluded, and there are multiple ways to fit the object model to it. We observe this specifically in the case of two objects in the ShapeNet dataset: mug and cap. For the mug, degeneracy occurs when the handle of the mug is not visible, whereas for the cap, when the visor of the cap is not visible (\cref{fig:expt_shapenet_degeneracy}).
\Cref{tab:expt_shapenet_deg} shows the ADD-S and ADD-S (AUC) scores attained by \name on these two objects (cap and mug). We observe that, while the self-supervised training works correctly to reach $100\%$ and $82\%$ observably correct instances for cap and mug, respectively, the ADD-S and ADD-S (AUC) scores remain low. This is because the %
observably correct instances include degenerate cases.  
This causes the predicted output to deviate from the ground truth. 
\Cref{tab:expt_shapenet_deg} also shows the ADD-S and ADD-S AUC scores for $\ocx=1$ and $\ocx=\ndx=1$ instances produced by \name; see~\Cref{app:expt_shapenet_ycb} for implementation details of the non-degeneracy certificate. We observe that the non-degeneracy check (\ie $\ndx=1$) on observably correct instances (\ie $\ocx=1$) significantly improves the performance, empirically validating~\cref{thm:pose-cert-guarantee}.  

\subsection{The YCB Experiment}
\label{sec:expt_ycb}

\begin{table}[b]
	\vspace{-2mm}

\centering
\caption{\name w/wo observable correctness~\eqref{eq:certificate-correctness} and non-degeneracy certificates~\eqref{eq:certificate-nondegen}.}
\label{tab:expt_shapenet_deg}
\begin{tabular}{|l|l|rr|r|}
	\toprule
	object & C-3PO & ADD-S & ADD-S AUC & percent \\
	\midrule
	\multirow{3}{*}{cap} & all & 78.00 & 63.87 & \textbf{100.00} \\
	& \ocx=1 & 78.00 & 63.87 & \textbf{100.00} \\
	& \ocx, \ndx=1 & \bf{100.00} & \textbf{80.11} & 32.00 \\
	\midrule
	\multirow{3}{*}{mug} & all & 68.00 & 53.55 & \textbf{100.00} \\
	& \ocx=1 & 82.93 & 60.63 & 82.00 \\
	& \ocx, \ndx=1 & \bf{100.00} & \textbf{70.06} & 22.00 \\
	\midrule
	\multirow{3}{*}{cracker box} & all & 68.00 & 54.24 & \textbf{100.00}\\
	& \ocx=1 & 69.39 & 55.35 & 98.00 \\
	& \ocx, \ndx=1 & \bf{100.00} & \textbf{81.08} & 42.00 \\
	\bottomrule
\end{tabular} %
\end{table} 

\begin{table*}

\centering
\caption{Evaluation of \name and baselines for the YCB experiment.}
\label{tab:expt_ycb}

\begin{tabular}{|l|rr|rr|rr|rr|rr|rr|}
\toprule
ADD-S~~~~ADD-S (AUC) &\multicolumn{2}{c}{chips can}  &\multicolumn{2}{c}{mustard bottle}  &\multicolumn{2}{c}{banana}  &\multicolumn{2}{c}{scissors}  &\multicolumn{2}{c}{extra large clamp}  &\multicolumn{2}{c|}{cracker box}  \\
\midrule
KeyPo (sim) & $0.00$ & $3.84$  & $0.00$ & $15.15$  & $2.00$ & $27.29$  & $34.00$ & $53.22$  & $0.00$ & $11.73$  & $0.00$ & $2.00$    \\
KeyPo (sim) + ICP & $0.00$ & $4.47$  & $0.00$ & $13.97$  & $4.00$ & $26.09$  & $36.00$ & $52.55$  & $2.00$ & $10.79$  & $0.00$ & $2.00$    \\
KeyPo (sim) + Corr. & $50.00$ & $41.68$  & $40.00$ & $48.63$  & $72.00$ & $68.06$  & $92.00$ & $80.59$  & $54.00$ & $53.14$  & $38.00$ & $36.18$    \\
C-3PO & $98.00$ & $78.48$  & $\textbf{100.00}$ & $\underline{84.04}$  & $\textbf{100.00}$ & $85.97$  & $\textbf{100.00}$ & $84.42$  & $\textbf{100.00}$ & $\textbf{84.22}$  & $68.00$ & $54.24$    \\
C-3PO (oc=1, nd=1) & $\textbf{100.00}$ & $\underline{78.78}$  & $\textbf{100.00}$ & $\textbf{85.05}$  & $\textbf{100.00}$ & $\underline{87.34}$  & $\textbf{100.00}$ & $\underline{86.77}$  & $\textbf{100.00}$ & $\textbf{84.80}$  & $\textbf{100.00}$ & $\textbf{81.08}$    \\
DeepGMR & $0.00$ & $7.62$  & $0.00$ & $16.51$  & $2.50$ & $30.49$  & $14.14$ & $35.64$  & $0.76$ & $7.06$  & $0.00$ & $0.00$    \\
PointNetLK & $0.00$ & $0.00$  & $0.00$ & $3.72$  & $22.17$ & $28.37$  & $7.07$ & $8.08$  & $0.00$ & $0.00$  & $0.00$ & $0.00$    \\
FPFH + TEASER++ & $0.17$ & $5.90$  & $0.33$ & $13.23$  & $4.83$ & $24.43$  & $47.81$ & $50.07$  & $5.29$ & $13.20$  & $0.17$ & $4.91$    \\
KeyPo (real) & $\textbf{100.00}$ & $\textbf{83.67}$  & $\textbf{100.00}$ & $74.12$  & $\textbf{100.00}$ & $\textbf{89.92}$  & $\textbf{100.00}$ & $\textbf{88.51}$  & $98.00$ & $82.06$  & $\underline{98.00}$ & $\underline{79.34}$    \\
\bottomrule
\end{tabular}

 	\vspace{-2mm}
\end{table*} 
This section shows that the proposed self-training method allows bridging the sim-to-real gap in the YCB dataset~\cite{calli15-YCBobject}. 

\myParagraph{Setup}
The YCB dataset~\cite{calli15-YCBobject} includes RGB-D images of household objects on a tabletop. 
The dataset provides Poisson-reconstructed meshes, which we use as object models $\objectModel$. We manually annotate semantic keypoints on each model. 
For the simulation data, we use uniformly sampled point clouds on $\objectModel$, and for the real-world data we use the segmented depth point clouds extracted from the RGB-D images in the dataset.

We use the same baselines as in the ShapeNet experiment, except EquiPose. This is because a trained EquiPose models are not available for the category-less YCB objects.
We evaluate the performance of each model using threshold ADD-S and ADD-S (AUC) score~\cite{Wang19cvpr-DenseFusion6D}.  
We choose thresholds for the ADD-S and ADD-S (AUC) to be $2$cm and $5$cm.  

\myParagraph{Results and Insights}
Table~\ref{tab:expt_ycb} shows the results for 6 of the 21 YCB objects.
The remaining objects are 
reported in~\arxivadd{\cref{app:expt_shapenet_ycb}}\arxivomit{\cite{Talak23arxiv-c3po}}.
Even though the YCB dataset is generated using real objects and sensors (unlike the dataset in the ShapeNet experiment), the key results and insights in~\cref{sec:expt_shapenet} hold.
We also see a large sim-to-real gap, and that the \KeyPo (sim) and \KeyPo (sim) + ICP do not yield good performance. The corrector helps to an extent to bridge the sim-to-real gap, and provide a respectable, initial fraction of observably correct instances for self-supervised training.
We observe that the self-supervised training works, as the observable correct instances increase with training iterations. 
\name significantly outperforms all the baseline approaches, and attains performance close to the supervised baseline \KeyPo (real). %
We observe several degenerate cases (see cracker box in~\cref{tab:expt_ycb}), but our implemented non-degeneracy certificate (see~\cref{app:expt_shapenet_ycb}) is able to filter them and boost performance: see~\cref{tab:expt_shapenet_deg}. This, again, validates~\cref{thm:pose-cert-guarantee}.

\subsection{Learning without Object Category Labels}
\label{sec:fully-ss}
\label{sec:expt_fullss}
As a final result, 
we show that \name's self-training works even when the data does not contain object category labels.

\myParagraph{Setup}
We use the depth point clouds from the ShapeNet and YCB experiments (see~\cref{sec:expt_shapenet,sec:expt_ycb}). We omit the object category labels and create two mixed datasets. The first one contains five ShapeNet objects (table, chair, bottle, laptop, skateboard) and the second contains five YCB objects (master chef can, mustard bottle, banana, scissors, extra large clamp). We train \name for each object in the dataset. We set $\epsilon_\ocx = 0.0316 d$ where $d$ is the object diameter.
\begin{table}[t]
\centering
\caption{Percentage of observably correct instances (\ie $\% \ocx=1$) for the sim-trained \name, for object A (row),  evaluated on the depth point cloud dataset for object B (column)}
\label{tab:expt_fullss_sim_eval}
\begin{tabular}{|l|r|r|r|r|r|}
		\toprule
		$\%$ obs. correct	& table 	& chair 	& bottle 	& laptop 	& skateboard \\  
		\midrule
		\name (table)				& $\textbf{74.00}$  	& $00.00$ 	& $00.00$	& $00.00$ 	& $00.00$ \\ %
		\name(chair)				& $00.00$  	& $\textbf{58.00}$ 	& $00.00$	& $00.00$ 	& $00.00$ \\ %
		\name(bottle)				& $00.00$  	& $00.00$ 	& $\textbf{98.00}$	& $00.00$ 	& $00.00$ \\ %
		\name(laptop)				& $00.00$  	& $00.00$ 	& $00.00$	& $\textbf{06.00}$ 	& $00.00$ \\ %
		\name(skateboard)			& $00.00$  	& $00.00$ 	& $00.00$	& $00.00$ 	& $\textbf{52.00}$ \\ %
		\bottomrule
\end{tabular} 	\vspace{-2mm}
\end{table}
\myParagraph{Results and Insights} 
A typical pose estimation method produces object poses without any certificates. 
The use of certificates in \name helps it distinguish not only if the estimated pose is correct, but also if the input is a partial point cloud of the same object. 
\Cref{tab:expt_fullss_sim_eval} shows \name(A) (for instance, ``\name (table)''), trained to estimates pose of object A in sim, produces $\ocx=0$ when it sees other objects (\ie chair, bottle, laptop, skateboard). 
Since we use observably correct instances to self-supervise (\cref{sec:training}), this indicates that the training will ignore pose estimates produced on other objects.

\Cref{tab:expt_fullss_chair} indeed shows that also in this case the 
self-supervised training in~\cref{sec:training} succeeds. 
In particular,~\cref{tab:expt_fullss_chair} evaluates \name(chair), trained to estimate the pose of a chair, on a mixed dataset containing depth point clouds of tables, chairs, bottles, laptop, and skateboard. We use the same ADD-S and ADD-S (AUC) thresholds as in~\cref{sec:expt_shapenet}. We see that \name(chair) is able to attain performance close to \name and the supervised baseline \KeyPo (real) in~\cref{tab:expt_shapenet}.
We get similar success when training and evaluating \name for other ShapeNet and YCB objects. 

\begin{table}[t]
\centering
\caption{
{Cross evaluation of \name(chair) trained on the mixed ShapeNet dataset to estimate pose of the chair object.} }
\label{tab:expt_fullss_chair}
\begin{tabular}{|l|r|r|r|r|r|}
		\toprule
							& table 	& chair 	& bottle 	& laptop 	& skateboard \\  
		\midrule
		ADD-S				& $00.00$  	& $\textbf{100.00}$ 	& $00.00$	& $00.00$ 	& $00.00$ \\ %
		ADD-S (AUC)			& $00.00$  	& $\textbf{83.73}$ 	& $00.00$	& $00.00$ 	& $00.00$ \\ %
		\bottomrule
\end{tabular} 	\vspace{-3mm}
\end{table}

\section{Conclusion}
\label{sec:conclusion}
We introduced the problem of certifiable pose estimation from partial point clouds, and defined the notion of \zetaCorrectness. We developed a theory to certify end-to-end perception models, and showed that implementing two certificates ---observable correctness and non-degeneracy--- provides a way to ascertain \zetaCorrectness of the resulting pose estimates. 
We proposed \name, a certifiable model, that implements the two certificates derived using our theory. 
\name relies on a semantic keypoint-based architecture, and 
also implements a keypoint corrector module, which we saw to be more effective at correcting pose errors than traditional methods like ICP.
Finally, we introduced a self-supervised training procedure, that leverages the certificate of observable correctness to provide a supervisory signal and enable self-training on unlabeled real data. 
Our experiments show that (i) 
standard semantic-keypoint-based methods (which constitute the backbone of \name) outperform more recent alternatives in challenging problem instances, (ii) \name further improves performance and significantly outperforms all the baselines, (iii) \name's certificates are able to discern correct pose estimates.

This work opens many avenues for future investigation. The success of the semantic-keypoint-based architecture for pose estimation indicates the need to explore in-category generalization power of a semantic keypoint detector.  We note here that self-supervised discovery of semantic keypoints on CAD models has been shown in~\cite{Suwajanakorn18nips-DiscoveryLatent}, thus obviating the need for hand-annotation of semantic keypoints. Self-supervised, category-level keypoint discovery is a problem, which, once solved, can extend \name to solve category-level object pose estimation.
Secondly, the theory of certifiable models developed in this work remains to be applied to other problems. We believe this can pave the way towards certifying end-to-end perception pipelines.
Finally, we believe that two other ideas presented in this paper, namely the corrector and the self-supervised training approach, can be extended and applied to other perception problems and deserve further investigation. 
%

\appendix

\subsection{Appendix: Hausdorff Metric for Certification} 
\label{app:sec:problem-statement}

The \zetaCorrectness of a pose estimate is defined in terms of the Hausdorff distance between the estimated $\predTransform \cdot \objectModel$ and the ground-truth $\gtTransform \cdot \objectModel$ posed CAD models.
Here we address the question of why we chose the Hausdorff distance, as opposed to other metrics that are more common in the pose estimation literature, like the pose error or the ADD-S score~\cite{Wang19cvpr-DenseFusion6D}. 

\newcommand{\gtRotation}{\ensuremath{\MR^\ast}\xspace}
\myParagraph{Pose Error} 
A common choice of metrics is to use rotation and translation error between the estimated pose $\predTransform$ and the ground truth $\gtTransform$. The translation error is typically computed as the Euclidean distance between the estimated and ground-truth translation. Several alternatives exist for measuring the rotation error~\cite{Hartley13ijcv,supplemental3Dinit}. Two popular options are (i) the angular distance, (ii) the chordal distance, which corresponds to the Frobenius norm of $\MI - \predRotation\tran\gtRotation$. 
While these metrics are appealing, they cause problems for symmetric objects. For a symmetric object, there exist at least two different poses from which the object is identical, \ie $\aTransform \cdot \objectModel = \objectModel$ for some $\aTransform \in \SEthree$.
Using rotation and translation error, therefore, necessitates us to first address the question of such equivalent poses, for each object, thus adding complexity and getting in the way of our analysis.  

\myParagraph{ADD-S Metric} A popular metric for evaluating pose estimation is the ADD-S metric~\cite{Wang19cvpr-DenseFusion6D}. It is defined as the Chamfer loss between the estimated $\predTransform \cdot \objectModelPC$ and the ground-truth $\gtTransform \cdot \objectModelPC$ posed CAD models. Unlike the metric in~\eqref{eq:hd}, ADD-S is computed on sampled point clouds. 
The difference between such a metric and the Hausdorff distance metric in~\eqref{eq:hd} is that the latter computes the worst-case distance from a point on $\predTransform \cdot \objectModel$ to the surface $\gtTransform \cdot \objectModel$, and vice-versa. The ADD-S metric, on the other hand, operates on averages, which can be problematic. 
An average metric, like ADD-S, tends to average out and potentially miss high errors in a few parts of the object, which gets in the way of certification. The Hausdorff distance metric, using ``max'' instead of ``average'', is able to single out such instances.

\subsection{Appendix: Certifiable Models} 
\label{app:sec:certifiable-model}

\myParagraph{Proof of~\cref{thm:epsilon-cert-general}} 
The result follows from the definitions. 
Recall that the ground truth $\outputGT$ generates the input data $\inputPC$, hence ---according to~\eqref{eq:solspace-from-genFun}--- it belongs to the solution space: $\outputGT \in \solSpace{\inputPC}$. 
Now assume that, for an estimate $\outputA = \Model(\inputPC)$, both the certificate of \observableCorrectness~\eqref{eq:correctness} and non-degeneracy~\eqref{eq:nondegeneracy} are equal to $1$. This implies  (i) $\outputA \in \solSpace{\inputPC}$ and (ii) $\diam{\solSpace{\inputPC}} < \delta$. Now, (ii) implies that $\distOutSpace{\outputB}{\outputC} < \delta$ for any $\outputB, \outputC \in \solSpace{\inputPC}$. Setting $\outputB = \outputA$ and $\outputC = \outputGT$, and using the assumption $\delta \leq \zeta$, we obtain the desired result.

\myParagraph{Proof of Theorem~\ref{thm:epsilon-cert-general-outer}}
Since $\solSpaceOuter{\inputPC}$ is an outer approximation of the solution set (\ie $\solSpace{\inputPC} \subseteq \solSpaceOuter{\inputPC}$) and $\outputGT 
\in \solSpace{\inputPC}$, it follows $\outputGT 
\in \solSpaceOuter{\inputPC}$. Moreover, since $\oc{\Model, \inputPC}=1$, we also know that $\outputA \in \solSpaceOuter{\inputPC}$. %
Now, $\nd{\Model, \inputPC}=1$ implies that $\distOutSpace{\outputB}{\outputC} < \delta$ for all $\outputB, \outputC \in \solSpaceOuter{\inputPC}$. Putting these three conclusion together, and using the assumption $\delta \leq \zeta$, we obtain the desired result.

\subsection{Appendix: Observable Correctness Certificate~\eqref{eq:certificate-correctness} as an Outer Approximation}
\label{app:sec:pose-oc-outer}
%
%
%
%
The observable correctness certificate~\eqref{eq:certificate-correctness} is given by
\begin{equation}
\oc{\outputPred, \inputPC} = \indicator{\max_{i \in [\nrPoints]} \min_{j \in [\nrModelPoints]} \lVert \inputPC[i] - \predTransform \cdot \objectModelPC[j] \rVert_2 < \epsilon_\ocx},
\end{equation}
where we have used the fact that $\predModel = \predTransform \cdot \objectModelPC$ in~\eqref{eq:certificate-correctness}. Note that we use the densely sampled CAD model \objectModelPC (instead of \objectModel) to implement the certificate, while the definition of \zetaCorrectness is with respect to the posed CAD model \objectModel.
For mathematical analysis, we also consider the following certificate
\begin{equation}
\label{eq:certificate-correctness-ideal}
\ocB{\outputPred, \inputPC} = \indicator{\max_{i \in [\nrPoints]}\distSurf{\inputPC[i]}{\predTransform \cdot \objectModel}  < \epsilon_{\ocx'}}.
\end{equation}
Note that~\eqref{eq:certificate-correctness-ideal} is an ``idealized'' version of the certificate in~\eqref{eq:certificate-correctness}, as it computes the exact distance $\distSurf{\inputPC[i]}{\MT \cdot \objectModel}$ from any point in the point cloud $\inputPC$ to the posed CAD model $\MT \cdot \objectModel$.
We prove the following.

\begin{theorem}[Outer Approximation]
\label{thm:outer}
Let $\solSpaceOuter{\inputPC}$ be given by
\begin{equation}
\nonumber
\solSpaceOuter{\inputPC} = \left\{ (\aTransform \cdot \objectKeypoints, \aTransform \cdot \objectModel)~\Big|~\max_{i \in [\nrPoints]}\distSurf{\inputPC[i]}{\MT \cdot \objectModel}  < \epsilon_{\ocx'} \right\}.
\end{equation}
Assuming bounded noise (Assumption~\ref{as:bounded-noise}) and dense CAD model sampling (Assumption~\ref{as:cad-model-sampling}), we have 
\begin{enumerate}[(i)]
\item $\solSpaceOuter{\inputPC}$ is an outer approximation of the solution space $\solSpace{\inputPC}$, provided $\epsilon_w < \epsilon_{\ocx'}$.

\item $\ocB{\outputA, \inputPC} \leq \oc{\outputA, \inputPC}$, provided $\epsilon_w + \epsilon_s < \epsilon_\ocx$.
\end{enumerate}  
\end{theorem}
\begin{IEEEproof} \textbf{(i)~}We first show that \solSpaceOuter{\inputPC} is an outer approximation to the solution space $\solSpace{\inputPC}$.
Recall that, given a generative model $\phi: \outputSpace \rightarrow \inputSpace$, the solution space \solSpace{\inputPC} is given by
\begin{equation}
\solSpace{\inputPC} = \left\{ \outputA \in \outputSpace~|~\phi(\outputA) = \inputPC \right\}.
\end{equation}
For the certifiable pose estimation problem, we have the generative model~\eqref{eq:predModel}:
\begin{equation}
\inputPC = \occFunction{\gtTransform \cdot \objectModel} + \noisePC.
\end{equation}
Let a tuple $(\aTransform \cdot \objectKeypoints, \aTransform \cdot \objectModel) \in \outputSpace$ be in the solution space \solSpace{\inputPC}. 
This implies that there exists %
a realization of the  
noise $\noisePCb$ such that $\inputPC - \noisePCb = \occFunction{\aTransform \cdot \objectModel}$, which is equivalent to saying that $\inputPC - \noisePCb \subseteq \aTransform \cdot \objectModel$. We can write this as 
\begin{equation}
\label{eq:01-oc-superset}
\distSurf{\inputPC[i] - \noisePCb[i]}{\aTransform \cdot\objectModel} = 0,~\forall~i \in [\nrPoints]
\end{equation}
The equation simply states that the output, up to noise, is a rigid transformation of points in the CAD model.
Since we assume bounded noise (\cref{as:bounded-noise}), we can write~\eqref{eq:01-oc-superset} as
\begin{equation}
\label{eq:02-oc-superset}
\max_{i \in [\nrPoints]} \distSurf{\inputPC[i]}{\aTransform \cdot \objectModel} < \epsilon_w,
\end{equation}
after making use of the triangle inequality.
From this, we know that whenever $(\aTransform \cdot \objectKeypoints, \aTransform \cdot \objectModel) \in \solSpace{\inputPC}$ we must have~\eqref{eq:02-oc-superset}, which is equivalent to saying $(\aTransform \cdot \objectKeypoints, \aTransform \cdot \objectModel) \in \solSpaceOuter{\inputPC}$ with $\epsilon_w < \epsilon_{\ocx'}$.

\textbf{(ii)~}We now show $\ocB{\outputA, \inputPC} \leq \oc{\outputA, \inputPC}$, provided $\epsilon_w + \epsilon_s < \epsilon_{\ocx}$. It suffices to prove that whenever $\ocB{\outputA, \inputPC} = 1$ we have $\oc{\outputA, \inputPC} = 1$. We do this by showing
\begin{multline}
\label{eq:condition}
\max_{i \in [\nrPoints]}\distSurf{\inputPC[i]}{\MT \cdot  \objectModel}  < \epsilon_w \\
\implies 
\max_{i \in [\nrPoints]} \min_{j \in [\nrModelPoints]} \lVert \inputPC[i] - \predModel[j] \rVert_2 < \epsilon_\ocx,
\end{multline}
provided $\epsilon_w + \epsilon_s < \epsilon_\ocx$.

Let the left-hand side (LHS) of~\eqref{eq:condition} hold for a \aTransform. We note that 
\begin{equation}
\label{eq:03-oc-superset}
\lVert \inputPC[i] - \predModel[j] \rVert_2 \leq \lVert \inputPC[i] - \vxx \rVert_2 + \lVert \vxx - \predModel[j] \rVert_2,
\end{equation}
for a $\vxx \in \aTransform \cdot \objectModel$ such that $\lVert \inputPC[i] - \vxx \rVert_2 < \epsilon_w$; which is possible due the LHS of~\eqref{eq:condition}. Taking minimum over $j \in [\nrModelPoints]$, we obtain
\begin{equation}
\label{eq:04-oc-superset}
\min_{j \in [\nrModelPoints]} \lVert \inputPC[i] - \predModel[j] \rVert_2 < \epsilon_w + \epsilon_s,
\end{equation}
after applying the CAD model sampling~\cref{as:cad-model-sampling}. This is nothing but the right-hand side of~\eqref{eq:condition}, provided $\epsilon_w + \epsilon_s < \epsilon_\ocx$.
\end{IEEEproof}

\begin{remark}
We remark here that Theorems~\ref{thm:outer} and~\ref{thm:epsilon-cert-general-outer} imply that the implemented observable correctness certificate is an upper-bound, \ie
\begin{equation}
\label{eq:temp02}
\idealObsCorrect \leq \oc{\outputA = \Model(\inputPC), \inputPC},
\end{equation}
under Assumptions~\ref{as:bounded-noise}-\ref{as:cad-model-sampling} and $\epsilon_\ocx > \epsilon_w + \epsilon_s$.  
\end{remark}

\subsection{Appendix: Proof of Theorem~\ref{thm:pose-cert-guarantee}}
\label{app:sec:pose-certificates}

We prove the result in three parts:
 (i) we first prove that if the certificate of non-degeneracy holds, then the posed keypoints 
cannot be far from the posed  ground-truth CAD model;
(ii) if the certificate of \observableCorrectness holds, then the posed CAD model cannot be far away from the posed and occluded ground-truth CAD model; 
and finally (iii) we use these two to show that the estimated pose \predTransform is $\zeta$-correct.

\smallheading{(i)}
Let $\outputPred = \posedOutputPred = \Model(\inputPC)$ be the model output as in~\eqref{eq:predModel}. Let $\nd{\outputPred, \inputPC}=1$. Then, there exists a $\setG_l$ such that
\begin{equation}
\label{eq:01-pose-cert-pf}
\min_{j \in [\nrPoints]} \lVert \predModelKP[i] - \inputPC[j] \rVert_2 < \delta_\ndx,~\forall~i \in \setG_l.
\end{equation} 
Now, $\inputPC = \occFunction{\gtTransform \cdot \objectModel} + \noisePC$. This implies  
\begin{align}
\lVert \predModelKP[i] - \occFunction{\gtTransform \cdot\objectModel}[j] \rVert_2 
&\leq \lVert \predModelKP[i] - \inputPC[j] \rVert_2 + \lVert \noisePC[j] \rVert_2, \\
&\leq \lVert \predModelKP[i] - \inputPC[j] \rVert_2 + \epsilon_w, \label{eq:02-pose-cert-pf}
\end{align}
where we used the bounded noise~\cref{as:bounded-noise}. 
Now, eqs.~\eqref{eq:02-pose-cert-pf} and~\eqref{eq:01-pose-cert-pf} imply 
\begin{equation}
\label{eq:03-pose-cert-pf}
\min_{j \in [\nrPoints]} \lVert \predModelKP[i] - \occFunction{\gtTransform \cdot\objectModel}[j] \rVert_2 < \delta_\ndx + \epsilon_w,~\forall~i \in \setG_l.
\end{equation}
This proves that the posed keypoints  $\predModelKP = \predTransform \cdot \objectKeypoints$, \cf~\eqref{eq:predModel}, are not far away from the posed ground-truth CAD model. We use 
$\predModelKP = \predTransform \cdot \objectKeypoints$, \cf~\eqref{eq:predModel}, to write~\eqref{eq:03-pose-cert-pf} as  
\begin{equation}
\label{eq:03b-pose-cert-pf}
\min_{j \in [\nrPoints]} \lVert \objectKeypoints[i] - \MC[j] \rVert_2 < \delta_\ndx + \epsilon_w,~\forall~i \in \setG_l,
\end{equation}
where $\MC = \predTransform\inv \cdot \occFunction{\gtTransform \cdot \objectModel}$. 
\smallheading{(ii)}
We also have $\oc{\outputPred, \inputPC}=1$. This implies 
\begin{equation}
\label{eq:04-pose-cert-pf}
\min_{i \in [\nrModelPoints]} \lVert \inputPC[j] - \predModel[i] \rVert_2 < \epsilon_\ocx,~\forall~j \in [\nrPoints].
\end{equation}
Again, using the fact that $\inputPC = \occFunction{\gtTransform \cdot \objectModel} + \noisePC$, and that the noise $\noisePC$ is bounded (\cref{as:bounded-noise}), {we can show}
\begin{equation}
\label{eq:05-pose-cert-pf}
\min_{i \in [\nrModelPoints]} \lVert \occFunction{\gtTransform\cdot \objectModel}[j] - \predModel[i] \rVert_2 < \epsilon_\ocx + \epsilon_w,~\forall~j \in [\nrPoints].
\end{equation}
Recall that $\predModel = \predTransform \cdot \objectModelPC$ and
consider the posed CAD model $\predTransform \cdot\objectModel$. Using~\eqref{eq:05-pose-cert-pf}, and the fact that every point in \predModel is in the set $\predTransform \cdot\objectModel$, we get
\begin{equation}
\label{eq:06-pose-cert-pf}
\distSurf{\occFunction{\gtTransform \cdot \objectModel}[j]}{\predTransform \cdot \objectModel} < \epsilon_\ocx + \epsilon_w,~\forall~j \in [\nrPoints].
\end{equation}
Substituting $\MC = \predTransform\inv \cdot \occFunction{\gtTransform \cdot \objectModel}$ in~\eqref{eq:06-pose-cert-pf} we get
\begin{equation}
\label{eq:06b-pose-cert-pf}
\distSurf{\MC}{\objectModel} < \epsilon_\ocx + \epsilon_w,~\forall~j \in [\nrPoints].
\end{equation}
This implies that there exists an occlusion and sampling function \occFunctionTwoX such that 
\begin{equation}
\label{eq:06c-pose-cert-pf}
\distHausdorff{\MC}{\occFunctionTwo{\objectModel}} < \epsilon_\ocx + \epsilon_w.
\end{equation}
Such a point cloud \occFunctionTwo{\objectModel} can be obtained by only sampling points on \objectModel that are closest to every point in $\MC$.

\smallheading{(iii)} From~\eqref{eq:03b-pose-cert-pf} and~\eqref{eq:06c-pose-cert-pf}, we obtain 
\begin{equation}
\label{eq:07-pose-cert-pf}
\min_{j} \lVert \objectKeypoints[i] - \occFunctionTwo{\objectModel}[j] \rVert_2 < \epsilon_\ocx + \delta_\ndx + 2\epsilon_w,
\end{equation}
for all $i \in \setG_l$, using the triangle inequality. This means that all the keypoints in the indicator set $\setG_l$ are close to the non-occluded part of \occFunctionTwo{\objectModel}. Since $\setG_l$ is an indicator set (\cref{def:indicator-set}), we have that \occFunctionTwo{\objectModel} is a cover set of \objectModel, provided 
\begin{equation}
\epsilon_\ocx + \delta_\ndx + 2\epsilon_w < \deltaIndicator.
\end{equation}
Now,~\eqref{eq:06c-pose-cert-pf} also implies that $\distSurf{\occFunctionTwo{\objectModel}[j]}{\MC} < \epsilon_\ocx + \epsilon_w$, from which we can deduce
\begin{equation}
\distSurf{\occFunctionTwo{\objectModel}[j]}{\predTransform^{-1} \cdot \gtTransform \cdot \objectModel} < \epsilon_\ocx + \epsilon_w,
\end{equation}
for all $j$. Knowing that \occFunctionTwo{\objectModel} is a cover set, this gives us $\distHausdorff{\objectModel}{\predTransform^{-1} \cdot \gtTransform \cdot \objectModel} < \epsilon_\ocx + \epsilon_w$, and therefore,
\begin{equation}
\distHausdorff{\predTransform \cdot \objectModel}{\gtTransform \cdot \objectModel} < \epsilon_\ocx + \epsilon_w,
\end{equation}
which proves the result. This proves that the estimate produced $\predTransform$ is $\zeta$-correct for any $\zeta > \epsilon_\ocx + \epsilon_w$. 

\subsection{Appendix: Corrector} 
\label{app:sec:corrector}

\myParagraph{Batch Gradient Descent} 
\begin{figure}
	\centering 
	\includegraphics[trim=0 0 0 0,clip,width=0.70\linewidth]{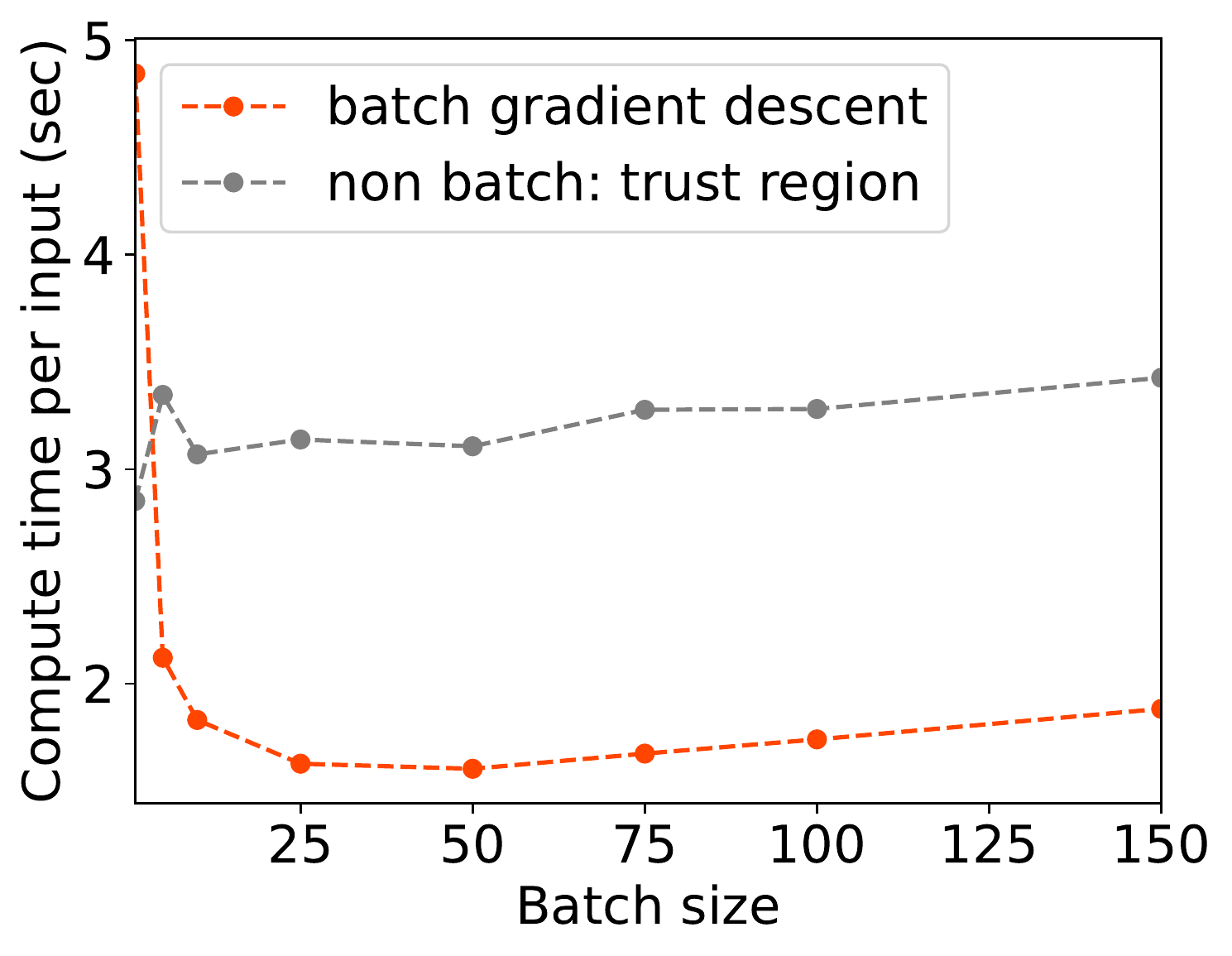}
	\caption{Compute time (sec) for solving~\eqref{eq:corrector} per data point in a batched input, as a function of the batch size.} 
	\label{fig:corrector_time}
	\vspace{-4mm}
\end{figure}
Figure~\ref{fig:corrector_time} shows a comparison of the compute time advantage of batch gradient descent, as opposed to using the SciPy optimization solver. The figure plots the time taken by a solver to output the solution to~\eqref{eq:corrector}, per data point in the batch, and a function of the total batch size. We see that with increasing batch size, the implemented batch gradient descent yields a compute time advantage over the SciPy solver and over using batch size of one. 

\begin{remark}[Batch Solvers for Differentiable Optimization Layers] We remark here that any differentiable optimization layer require batch solvers to enable faster training. We observe that implementations using existing solvers, and iterating over data points in the batch, results in slower training times; whereas PyTorch solvers (\eg SDG, ADAM) are not suited to solve non-linear optimization problems by design.%
\end{remark}

\label{pf:thm:corrector_bkprop}
\myParagraph{Proof of Theorem~\ref{thm:corrector_bkprop}}
Note that the only places where $\kpCorrection$ appears in the corrector is in conjunction with $\kpDetected$, and that too, in the form $\kpDetected + \kpCorrection$ (see~\eqref{eq:pose_est}, \eqref{eq:corrector}). The corrector optimization problem can therefore be written as:
\begin{equation}
\begin{aligned}
& \min_{\kpCorrection} %
& & f(\kpDetected + \kpCorrection), \label{eq:corrector-proof}
\end{aligned}
\end{equation}
for some function $f$, obtained by subsuming the constraints in~\eqref{eq:corrector} into the objective function and realizing that $\MR(\kpCorrection)$ and $\vt(\kpCorrection)$, can in fact be written as $\MR(\kpDetected + \kpCorrection)$ and $\vt(\kpDetected + \kpCorrection)$.

Let $\vy^{\ast}$ be a solution of the optimization problem $\min_{\vy} f(\vy)$, for $f$ in~\eqref{eq:corrector-proof}. Then, the optimal correction will be $\kpOptCorrection = \vy^{\ast} - \kpDetected$.
Taking derivative with respect to $\kpDetected$ gives the result. 
\label{sec:batch-gd}

\arxivadd{
\begin{table}
\centering
\caption{Comparing the keypoint detector architectures used in \name: PointNet++ vs Point Transformer.}
\label{tab:expt_shapenet_extra02}
\begin{tabular}{|l|rrr|l|}
\toprule
Object & ADD-S & ADD-S AUC & \% oc=1 & keypoint detector \\
\midrule
\multirow{2}{*}{airplane}
& \textbf{100.00} &\textbf{ 87.76} & \textbf{100.00} & Point Transformer \\
& 58.00 & 63.35 & 54.00 & PointNet++ \\
\midrule
\multirow{2}{*}{bathtub}
& \textbf{72.00} & \textbf{57.71} & \textbf{60.00} & Point Transformer \\
& 54.00 & 50.40 & 50.00 & PointNet++ \\
\midrule
\multirow{2}{*}{bed}
& \textbf{56.00} & \textbf{52.68} & 0.00 & Point Transformer \\
& 34.00 & 38.14 & 0.00 & PointNet++ \\
\midrule
\multirow{2}{*}{bottle}
& \textbf{100.00} & \textbf{70.35} & \textbf{100.00} & Point Transformer \\
& \textbf{100.00} & 68.41 & \textbf{100.00} & PointNet++ \\
\midrule
\multirow{2}{*}{cap}
& \textbf{78.00} & \textbf{63.87} & \textbf{100.00} & Point Transformer \\
& 60.00 & 47.55 & 82.00 & PointNet++ \\
\midrule
\multirow{2}{*}{car}
& \textbf{80.00} & \textbf{64.00} & \textbf{6.00} & Point Transformer \\
& 46.00 & 44.53 & 0.00 & PointNet++ \\
\midrule
\multirow{2}{*}{chair}
& \textbf{100.00} & \textbf{79.06} & \textbf{12.00} & Point Transformer \\
& \textbf{100.00} & 73.44 & 2.00 & PointNet++ \\
\midrule
\multirow{2}{*}{guitar}
& \textbf{100.00} & \textbf{83.77} & \textbf{28.00} & Point Transformer \\
& \textbf{100.00} & 80.03 & 0.00 & PointNet++ \\
\midrule
\multirow{2}{*}{helmet}
& \textbf{100.00} & \textbf{70.49} & \textbf{92.00} & Point Transformer \\
& 60.00 & 47.78 & 32.00 & PointNet++ \\
\midrule
\multirow{2}{*}{knife}
& \textbf{100.00} & \textbf{86.94} & \textbf{86.00} & Point Transformer \\
& \textbf{100.00} & 80.28 & 20.00 & PointNet++ \\
\midrule
\multirow{2}{*}{laptop}
& \textbf{100.00} & \textbf{64.70} & \textbf{100.00} & Point Transformer \\
& 82.00 & 56.67 & 86.00 & PointNet++ \\
\midrule
\multirow{2}{*}{motorcycle}
& \textbf{100.00} & \textbf{75.02} & \textbf{18.00} & Point Transformer \\
& 98.00 & 71.33 & 0.00 & PointNet++ \\
\midrule
\multirow{2}{*}{mug}
& \textbf{68.00} & \textbf{53.55} & \textbf{82.00} & Point Transformer \\
& 32.00 & 37.38 & 52.00 & PointNet++ \\
\midrule
\multirow{2}{*}{skateboard}
& \textbf{100.00} & \textbf{82.90} & \textbf{100.00} & Point Transformer \\
& 96.00 & 79.66 & 96.00 & PointNet++ \\
\midrule
\multirow{2}{*}{table}
& 98.00 & \textbf{67.69} & 98.00 & Point Transformer \\
& \textbf{100.00} & 64.61 & \textbf{100.00} & PointNet++ \\
\midrule
\multirow{2}{*}{vessel}
& \textbf{100.00} & \textbf{78.21} & \textbf{80.00} & Point Transformer \\
& 80.00 & 63.23 & 20.00 & PointNet++ \\
\bottomrule
\end{tabular}
 \end{table}
}

\begin{table}
\centering
\caption{\name w/wo observable correctness and non-degeneracy certificates.}
\label{tab:expt_ycb_extra02}
\begin{tabular}{|l|l|rr|r|}
	\toprule
	object & C-3PO & ADD-S & ADD-S AUC & percent \\
	\midrule
	\multirow{3}{*}{sugar box} & all & 58.00 & 52.03 & $\mathbf{100.00}$ \\
	& \ocx=1 & 58.33 & 52.57 & 96.00 \\
	& \ocx, \ndx=1 & \textbf{100.00} & \textbf{87.43} & 50.00 \\
	\midrule
	\multirow{3}{*}{pudding box} & all & 50.00 & 53.82 & \textbf{100.00} \\
	& \ocx=1 & 26.09 & 36.52 & 46.00 \\
	& \ocx, \ndx=1 & \textbf{100.00} & \textbf{89.73} & 8.00 \\
	\midrule
	\multirow{3}{*}{gelatin box} & all & 82.00 & 78.07 & \textbf{100.00} \\
	& \ocx=1 & 73.53 & 74.41 & 68.00 \\
	& \ocx, \ndx=1 & \textbf{100.00} & \textbf{89.26} & 42.00 \\
	\midrule
	\multirow{3}{*}{potted meat can} & all & 54.00 & 72.17 & \textbf{100.00} \\
	& \ocx=1 & 61.90 & 75.86 & 84.00 \\
	& \ocx, \ndx=1 & \textbf{100.00} & \textbf{88.81} & 36.00 \\
	\midrule
	\multirow{3}{*}{power drill} & all & 86.00 & 81.42 & \textbf{100.00} \\
	& \ocx=1 & 97.22 & 85.53 & 72.00 \\
	& \ocx, \ndx=1 & \textbf{100.00} & \textbf{87.13} & 56.00 \\
	\midrule
	\multirow{3}{*}{wood block} & all & 62.00 & 50.17 & \textbf{100.00} \\
	& \ocx=1 & 48.28 & 39.85 & 58.00 \\
	& \ocx, \ndx=1 & $\mathbf{100.00}$ & \textbf{83.11} & 22.00 \\
	\midrule
	\multirow{3}{*}{foam brick} & all & 98.00 & 85.33 & \textbf{100.00} \\
	& \ocx=1 & 96.67 & 85.74 & 60.00 \\
	& \ocx, \ndx=1 & \textbf{100.00} & \textbf{89.68} & 42.00 \\
	\bottomrule
\end{tabular}
\vspace{-0.3cm} \end{table} %
\arxivadd{ 

\subsection{Appendix: Keypoint Corrector Analysis Experiment}
\label{app:expt_corrector_analysis}

We study the performance of the corrector across 16 object categories in the ShapeNet dataset~\cite{Chang15arxiv-shapenet}. We randomly pick an object in each category and perform the ablation study described in Section~\ref{sec:expt_corrector_analysis}. 
Figures~\ref{fig:corrector-analysis-extra-01},~\ref{fig:corrector-analysis-extra-02},~\ref{fig:corrector-analysis-extra-03}, and~\ref{fig:corrector-analysis-extra-04} plot ADD-S, normalized to the object diameter, and percent \emph{not} observably correct, as a function of the noise variance parameter $\sigma$. 
The sub-figures (c) visualizes an instance of keypoint perturbation at $\sigma = 0.8$. 
In all cases we observe that the corrector is able to correct added noise to the keypoints, even when $80\%$ of the keypoints are randomly perturbed by a measure proportional to the objects' diameter. 
We observe a reduction in the variance of the ADD-S score and improved fraction of observaly correct instances, across all object categories. The insights noted in Section~\ref{sec:expt_corrector_analysis} hold across all the objects we analyzed. }

\subsection{Appendix: The ShapeNet and YCB Experiment}
\label{app:expt_shapenet_ycb}  

\arxivadd{
	
\myParagraph{ShapeNet Data Preparation} 
ShapeNet~\cite{Chang15arxiv-shapenet} classifies objects into 16 categories. We select one object in each category and use the semantic keypoints labeled by the KeypointNet dataset~\cite{You20cvpr-KeypointNetLargescale}. We use uniformly sampled point clouds of these 16 objects as the simulation dataset and a collection of depth point clouds, rendered using Open3D~\cite{Zhou18arxiv-open3D}, as the real-world dataset. This choice ensures a large sim-to-real gap, and enables us to showcase the utility of our self-supervised training.
We initially train the detector using the simulation dataset, and self-train it on the depth point cloud dataset, %
as proposed in Section~\ref{sec:training}. 
For the real-world, depth point cloud data, we generate $5000$ point clouds for training, $50$ point clouds each for validation and testing, respectively.  
We note that the pose error induced is equal to the \emph{hard} case described in Section~\ref{sec:keypo}.

\myParagraph{YCB Data Preparation} 
The YCB dataset~\cite{calli15-YCBobject} includes RGB-D images of household objects on a tabletop. 
The dataset provides Poisson-reconstructed meshes, which we use as object models $\objectModel$. We manually annotate semantic keypoints on each model. 
We choose a subset of 19 objects that span across varied shapes and sizes within the dataset, and use the Poisson reconstructed meshes as the object model $\objectModel$. We manually annotated semantic keypoints on each of the 19 objects, using Open3D's visualizer interface~\cite{Zhou18arxiv-open3D}.
For the simulation data, we use uniformly sampled point clouds on $\objectModel$, and for the real-world data we use the segmented depth point clouds extracted from the RGB-D images in the dataset.
The number of depth point clouds extracted varies across the YCB objects. We generate the train/test/val data using a $80/20/20$ split.

To increase the train dataset size and to ensure that the learned model is robust to small changes in the point cloud, we perform data augmentation on the train and validation data. For each depth point cloud, in the train and validation set, we remove a randomly selected $10\%$ of the points, and perturb the rest with Gaussian noise, with a standard deviation of 0.001m. We also discarded point clouds with less than 500 points. 

}

\begin{figure*}
	\centering
	\begin{subfigure}{.17\textwidth}
		\centering
		\includegraphics[trim=24 1 0 1,clip,width=0.8\linewidth]{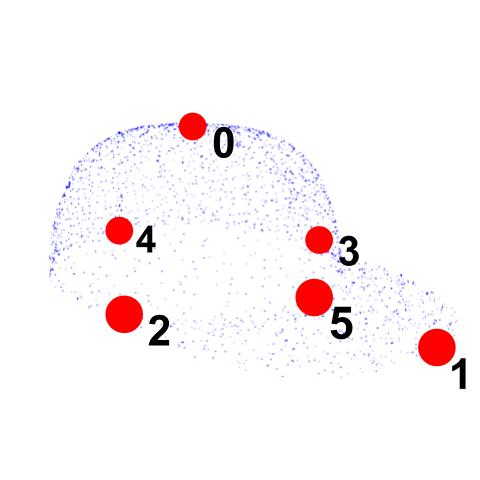} 
		\caption{cap}
	\end{subfigure}%
	\begin{subfigure}{.18\textwidth}
		\centering
		\includegraphics[trim=0 0 10 1,clip,width=0.8\linewidth]{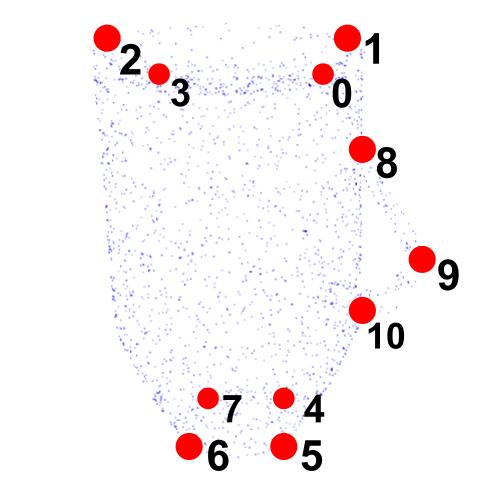} 
		\caption{mug}
	\end{subfigure}
	\begin{subfigure}{.19\textwidth}
		\centering
		\includegraphics[trim=2 1 0 1,clip,width=0.8\linewidth]{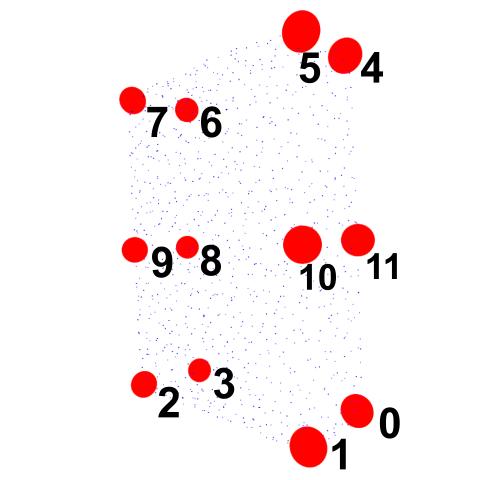}  
		\caption{004 sugar box}
	\end{subfigure}%
	\begin{subfigure}{.19\textwidth}
		\centering
		\includegraphics[trim=2 1 0 1,clip,width=0.8\linewidth]{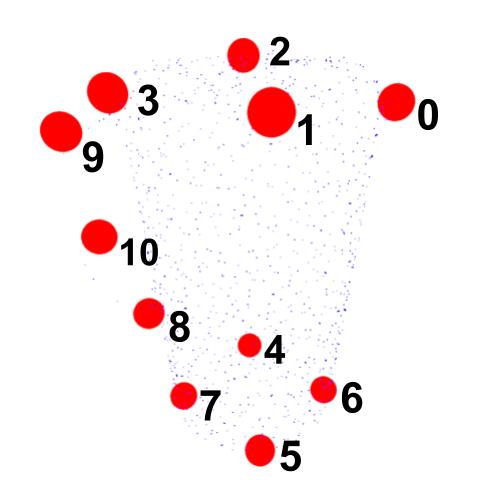} 
		\caption{019 pitcher base}
	\end{subfigure}
	\begin{subfigure}{.19\textwidth}
		\centering
		\includegraphics[trim=2 1 0 1,clip,width=0.8\linewidth]{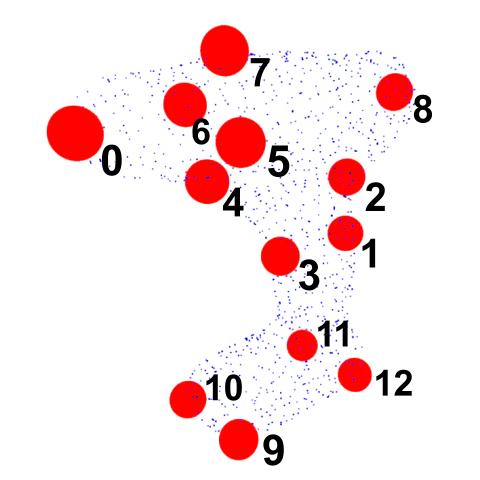}
		\caption{035 power drill}  
	\end{subfigure}%
	\caption{Annotated keypoints and keypoint indices for ShapeNet objects (a) cap, (b) mug and for YCB objects (c) 004 sugar box, (d) 019 pitcher base, and (e) 035 power drill.}
	\label{fig:nd-cert}
	\vspace{-4mm}
\end{figure*}

\myParagraph{Non-degeneracy Certificate} 
In this appendix, we discuss the implementation of the non-degeneracy certificates and the choice of indicator sets, used in the experiments. We note that deriving indicator sets to provably meet the specifications in~\cref{def:indicator-set} is hard and we do not choose that route. We, instead, hand-craft the indicator sets and show its usefulness towards correctly certifying \name's pose estimates. 

For most of the ShapeNet and YCB objects, we choose the indicator set to be empty, \ie $\setG_l = \emptyset$, implying that $\nd{\outputPred, \inputPC} = 1$ always. This is because any depth point cloud of the object turns out to be a valid cover set, and therefore, can uniquely identify the object pose (see~\eqref{eq:cover-set-condition}). 
However, for some objects, this turns out not to be the case (\eg cap in Figure~\ref{fig:expt_shapenet_degeneracy}). We implement non-trivial non-degeneracy certificates for ShapeNet object's cap and mug, and for YCB object's boxes (of any kind), pitcher base, and power drill.

Let $\calG = \left\{ \setG_l \right\}_{l=1}^{g}$ denote the collection of all the indicator sets. We describe our implementation of $\calG$ for the few cases in which we implement a non-trivial non-degeneracy certificate. Figure~\ref{fig:nd-cert} shows a list of five ShapeNet and YCB objets, along with annotated semantic keypoints and their indices. For these objects we implement a non-trivial, non-degeneracy certificate. 
For the cap and mug, we choose $\calG$ = \{set([1])\} and $\calG$ = \{set([9])\}, respectively. Note that [1] is the keypoint on the flap of the cap, while [9] is the keypoint on the handle of the mug. This choice requires that the input point cloud \inputPC have points near the flap of the cap and the handle of the mug, to be deemed as a cover set, and thereby, declare the input \inputPC as non-degenerate.
For 019 pitcher base we choose  $\calG$ = \{ set([8]), set([10])\} as non-visibility of the handle causes degeneracy, and for 035 power drill we choose  $\calG$ = \{set([9]), set([10]), set([11]), set([12])\} as non-visibility of the base of the power drill leads to degeneracy.

We implement non-trivial non-degeneracy certificate for all the boxes in the YCB dataset. This is because it is impossible to determine the correct pose of the box by seeing the depth point cloud of only one side. We therefore implement $\calG$ = \{set([0,1,3,4]), set([0,1,2,5]), set([1,2,3,7]), set([0,2,3,6])set([4,5,6,0]), set([4,5,7,1]), set([5,6,7,2]), set([4,6,7,3])\}, which requires visibility of at least three sides of the box to be declared non-degenerate in~\eqref{eq:nondegeneracy}.
We choose $\delta_\ndx$ in~\eqref{eq:certificate-nondegen} to be $1.5\%$ of the object diameter for ShapeNet objects and $1.5$cm for the YCB objects.

\begin{remark}[Learning Indicator Sets]
While the implementation of the non-degeneracy certificate is hand-crafted, we only use it at test time and on a few objects. For most objects, evaluations show that observable correctness imply certifiable correctness. We leave it to future work to implement a more exact and systematic non-degeneracy certificate --probably a learning-based model, that does not require hand-crafting.
\end{remark}

\myParagraph{Hyper-parameters} We describe our choice of learning rate, number of epochs, the $\epsilon_\ocx$ used in certification (\cref{def:cert-correctness}), and other parameters used during the self-supervised training. We set the maximum number of epochs to be $20$ for ShapeNet and $50$ for YCB objects. We made two exceptions in the case of ShapeNet, where we observed that the models for ``car'' and ``helmet'' were still training (\ie observed continuing decrease in the train and validation loss, and improvement in the percent certifiable). For these two models we set the maximum number of epochs to $40$. For the learning rate, we used $0.02$, momentum to be $0.9$, while the training batch size was $50$. We did not optimize much over these parameters, but only found ones that worked. We believe some improvement can be expected by exhaustively optimizing these training hyper-parameters.

We observe that correctly setting up the certification parameter $\epsilon_\ocx$ in~\eqref{eq:certificate-correctness} plays a crucial role in efficiently training the model in a self-supervised manner. Setting $\epsilon_\ocx$ too high allows for the model to make and accept errors in training, while setting it to too tight leads to increased training time, as a very small number of input-output pairs remain certifiable, for the simulation-trained model with the corrector, at the beginning of the self-supervised training.  
A good $\epsilon_\ocx$ can be obtained either by grid search or by visually inspecting the simulation-trained \KeyPo, applied on the real-world data, with the corrector. 
For instance, if all the declared $\ocx=1$ instances are also visually correct (\ie posed CAD model \predModel is correctly aligned with the input point cloud \inputPC), and the $\ocx=0$ instances are not, then the chosen $\epsilon_\ocx$ is correct. \Cref{tab:epsilon} lists the $\epsilon_\ocx$ used for various objects in the ShapeNet and YCB dataset.

\begin{remark}[Automatic Tuning of $\epsilon_\ocx$]
The parameter $\epsilon_\ocx$ in~\eqref{eq:certificate-correctness} (i) determines the observable correctness of the input-output pairs and (ii) influences the proposed self-supervised training (see~\cref{sec:training}). This dual purpose makes $\epsilon_\ocx$ amenable for automatic tuning during self-supervised training. We think that it is possible to use the number of observably correct instances, during training, as an indicator to automatically adjust $\epsilon_\ocx$. However, we leave this thread of inquiry for future work. 
\end{remark}

\begin{table*}
\centering
\caption{$\epsilon_\ocx$ for ShapeNet (in \% of object diameter $d$) and YCB objects (in cm).}
\label{tab:epsilon}
\begin{tabular}{|l|r|l|r|l|r|l|r|l|r|l|r|}
	\toprule
	object & $\epsilon_\ocx$   &   object & $\epsilon_\ocx$   &   object & $\epsilon_\ocx$   &   object & $\epsilon_\ocx$   &   object & $\epsilon_\ocx$   &   object & $\epsilon_\ocx$   \\
	\midrule
	airplane &  3.16 & 
	bathtub &  4.47 & 
	bed &  4.47 & 
	bottle &  10.03 & 
	cap &  10.03 & 
	car &  3.16 \\
	chair &  3.16 & 
	guitar &  2.24 & 
	helmet &  4.47 & 
	knife &  2.24 & 
	laptop &  7.08 & 
	motorcycle &  3.16 \\
	mug &  7.08 & 
	skateboard &  3.16 & 
	table &  10.03 & 
	vessel &  3.16 & 
	&  & 
	&  \\
	\bottomrule
\end{tabular}

\vspace{0.3cm}

\begin{tabular}{|l|r|l|r|l|r|l|r|l|r|}
	\toprule
	object & $\epsilon_\ocx$   &   object & $\epsilon_\ocx$   &   object & $\epsilon_\ocx$   &   object & $\epsilon_\ocx$   &   object & $\epsilon_\ocx$   \\
	\midrule
	chips can &  1.00 & 
	master chef can &  1.00 & 
	cracker box &  1.41 & 
	sugar box &  1.00 & 
	tomato soup can &  0.71 \\
	mustard bottle &  1.00 & 
	tuna fish can &  0.77 & 
	pudding box &  0.71 & 
	gelatin box &  0.71 & 
	potted meat can &  1.00 \\
	banana &  0.89 & 
	pitcher base &  1.58 & 
	bleach cleanser &  1.14 & 
	power drill &  0.89 & 
	wood block &  1.00 \\
	scissors &  0.95 & 
	large marker &  1.00 & 
	large clamp &  0.71 & 
	extra large clamp &  0.89 & 
	foam brick &  0.71 \\ 
	\bottomrule
\end{tabular}

 	\vspace{-2mm}
\end{table*} 

\arxivadd{
\myParagraph{PointNet++ vs Point Transformer} 
In Section~\ref{sec:expt_shapenet}, we noted that the \name works better when we use point transformer as a  keypoint detector, as opposed to PointNet++, \ie the former converges to 100\% certifiability much sooner during training (see~\cref{fig:expt_shapenet_cert}). \Cref{tab:expt_shapenet_extra02} reports the ADD-S, ADD-S (AUC), and percentage certifiable for models trained using the point transformer and the PointNet++ architecture. We observe that the point transformer performs better in all the cases.

\myParagraph{Extended Results} 
For the ShapeNet dataset, we provide the results for 16 randomly selected objects, one in each object category, in Table~\ref{tab:expt_shapenet_extra01}. We observe similar trends as noted in Section~\ref{sec:expt_shapenet} across all the object categories. We observe that additions such as ICP or outlier rejection of perturbed keypoints using RANSAC does not help much. The performance of \name reaches close to that of the supervised baseline \KeyPo (real) in most cases. We note that mug and cap remain the only two degenerate cases, for which we report the results using the certificate of non-degeneracy in Table~\ref{tab:expt_shapenet_deg} in Section~\ref{sec:expt_shapenet}.

For the YCB experiment, we present the extended set of results, for all baselines and 18 YCB object models, in Table~\ref{tab:expt_ycb_extra01}. We observe a similar trend as in Table~\ref{tab:expt_ycb}. We, however, observe quite a few degenerate cases, as opposed to the ShapeNet dataset. These cases arise for different types of boxes, blocks, brick, can and a power drill. For the boxes, degeneracy arises when only one face of the box is visible. Similar degnerate cases are observed for other box-like objects such as brick and blocks. Degeneracy arises in the power drill when only the drill portion is visible, but not its handle. We report our evaluation for these cases in Table~\ref{tab:expt_ycb_extra02}, and show the performance boost in implementing the certificate of non-degeneracy. We observe that while the ADD-S and ADD-S (AUC) numbers are low over all, non-certified instances, we see high ADD-S and ADD-S (AUC) scores for the observably correct and non-degenerate instances produced by \name. This empirically validates our implemented non-degeneracy certificates and Theorem~\ref{thm:pose-cert-guarantee}. %
}

\bibliographystyle{ieeetr}

\arxivadd{

\newpage
\begin{table*}
\centering
\caption{Evaluation of \name and baselines for the ShapeNet experiment}
\label{tab:expt_shapenet_extra01}

\begin{tabular}{|l|rr|rr|rr|rr|rr|rr|}

	\toprule
	ADD-S~~~~ADD-S (AUC) &\multicolumn{2}{c}{airplane}  &\multicolumn{2}{c}{bathtub}  &\multicolumn{2}{c}{bed}  &\multicolumn{2}{c}{bottle}  &\multicolumn{2}{c}{cap}  &\multicolumn{2}{c|}{car}  \\
	\midrule
	KeyPo (sim) & $0.00$ & $2.75$  & $0.00$ & $0.00$  & $0.00$ & $2.00$  & $0.00$ & $0.00$  & $0.00$ & $0.00$  & $0.00$ & $0.00$    \\
	KeyPo (sim) + ICP & $0.00$ & $2.43$  & $0.00$ & $0.00$  & $0.00$ & $0.00$  & $0.00$ & $0.00$  & $0.00$ & $0.00$  & $0.00$ & $0.00$    \\
	KeyPo (sim) + RANSAC + ICP & $0.00$ & $3.99$  & $0.00$ & $0.00$  & $0.00$ & $0.00$  & $0.00$ & $0.00$  & $0.00$ & $0.00$  & $0.00$ & $0.00$    \\
	KeyPo (sim) + Corr. & $36.00$ & $42.55$  & $18.00$ & $26.86$  & $12.00$ & $24.37$  & $96.00$ & $67.23$  & $30.00$ & $25.59$  & $0.00$ & $13.69$    \\
	KeyPo (sim) + Corr. + ICP & $32.00$ & $41.55$  & $14.00$ & $23.78$  & $6.00$ & $20.16$  & $96.00$ & $67.51$  & $24.00$ & $20.44$  & $2.00$ & $13.44$    \\
	C-3PO & $\textbf{100.00}$ & $87.76$  & $72.00$ & $57.71$  & $\underline{56.00}$ & $\underline{52.68}$  & $\textbf{100.00}$ & $70.35$  & $78.00$ & $63.87$  & $80.00$ & $64.00$    \\
	C-3PO (oc=1, nd=1) & $\textbf{100.00}$ & $\underline{87.76}$  & $\underline{90.00}$ & $\underline{70.92}$  & $0.00$ & $0.00$  & $\textbf{100.00}$ & $\underline{70.35}$  & $\textbf{100.00}$ & $\underline{80.11}$  & $\textbf{100.00}$ & $\textbf{84.59}$    \\
	DeepGMR & $0.00$ & $0.00$  & $0.00$ & $0.00$  & $0.00$ & $0.00$  & $0.00$ & $0.00$  & $0.00$ & $0.00$  & $0.00$ & $2.00$    \\
	PointNetLK & $0.00$ & $0.00$  & $0.00$ & $2.00$  & $0.00$ & $0.00$  & $0.00$ & $2.00$  & $0.00$ & $0.00$  & $0.00$ & $2.76$    \\
	FPFH + TEASER++ & $40.00$ & $45.24$  & $26.00$ & $21.63$  & $2.00$ & $7.74$  & $80.00$ & $60.08$  & $46.00$ & $44.96$  & $26.00$ & $25.03$    \\
	EquiPose & $8.00$ & $22.81$  & -- &  --  & -- &  --  & $\textbf{100.00}$ & $75.11$  & -- &  --  & $16.00$ & $19.69$    \\
	KeyPo (real) & $\textbf{100.00}$ & $\textbf{87.95}$  & $\textbf{98.00}$ & $\textbf{74.97}$  & $\textbf{100.00}$ & $\textbf{77.88}$  & $\textbf{100.00}$ & $\textbf{74.72}$  & $\textbf{100.00}$ & $\textbf{82.59}$  & $\textbf{100.00}$ & $\underline{79.08}$    \\
	\bottomrule
	
	\addlinespace[2ex]
	
	\toprule
	ADD-S~~~~ADD-S (AUC) &\multicolumn{2}{c}{chair}  &\multicolumn{2}{c}{guitar}  &\multicolumn{2}{c}{helmet}  &\multicolumn{2}{c}{knife}  &\multicolumn{2}{c}{laptop}  &\multicolumn{2}{c|}{motorcycle}  \\
	\midrule
	KeyPo (sim) & $0.00$ & $12.01$  & $0.00$ & $2.51$  & $0.00$ & $2.00$  & $0.00$ & $2.89$  & $0.00$ & $0.00$  & $4.00$ & $15.01$    \\
	KeyPo (sim) + ICP & $0.00$ & $11.56$  & $2.00$ & $3.80$  & $0.00$ & $2.00$  & $0.00$ & $2.43$  & $2.00$ & $2.00$  & $2.00$ & $14.74$    \\
	KeyPo (sim) + RANSAC + ICP & $0.00$ & $2.00$  & $0.00$ & $3.06$  & $0.00$ & $2.00$  & $14.00$ & $16.24$  & $0.00$ & $0.00$  & $0.00$ & $2.00$    \\
	KeyPo (sim) + Corr. & $68.00$ & $57.16$  & $96.00$ & $72.63$  & $10.00$ & $20.14$  & $62.00$ & $68.51$  & $26.00$ & $20.57$  & $74.00$ & $59.25$    \\
	KeyPo (sim) + Corr. + ICP & $60.00$ & $50.51$  & $90.00$ & $68.10$  & $12.00$ & $21.82$  & $62.00$ & $67.03$  & $18.00$ & $14.59$  & $74.00$ & $58.01$    \\
	C-3PO & $\textbf{100.00}$ & $79.06$  & $\textbf{100.00}$ & $83.77$  & $\textbf{100.00}$ & $70.49$  & $\textbf{100.00}$ & $86.94$  & $\textbf{100.00}$ & $\underline{64.70}$  & $\textbf{100.00}$ & $75.02$    \\
	C-3PO (oc=1, nd=1) & $\textbf{100.00}$ & $\textbf{90.13}$  & $\textbf{100.00}$ & $\underline{86.63}$  & $\textbf{100.00}$ & $\underline{71.53}$  & $\textbf{100.00}$ & $\underline{87.70}$  & $\textbf{100.00}$ & $\underline{64.70}$  & $\textbf{100.00}$ & $\underline{81.03}$    \\
	DeepGMR & $0.00$ & $0.00$  & $0.00$ & $0.00$  & $0.00$ & $0.00$  & $0.00$ & $0.00$  & $0.00$ & $0.00$  & $0.00$ & $0.00$    \\
	PointNetLK & $6.00$ & $7.06$  & $4.00$ & $4.93$  & $0.00$ & $0.00$  & $4.00$ & $5.23$  & $0.00$ & $0.00$  & $2.00$ & $3.62$    \\
	FPFH + TEASER++ & $26.00$ & $25.65$  & $74.00$ & $55.81$  & $50.00$ & $46.29$  & $92.00$ & $71.62$  & $42.00$ & $32.32$  & $34.00$ & $30.98$    \\
	EquiPose & $14.00$ & $20.33$  & -- &  --  & -- &  --  & -- &  --  & -- &  --  & -- &  --    \\
	KeyPo (real) & $\textbf{100.00}$ & $\underline{79.92}$  & $\textbf{100.00}$ & $\textbf{87.02}$  & $\textbf{100.00}$ & $\textbf{80.28}$  & $\textbf{100.00}$ & $\textbf{90.66}$  & $\textbf{100.00}$ & $\textbf{83.11}$  & $\textbf{100.00}$ & $\textbf{83.25}$    \\
	\bottomrule
	
	\addlinespace[2ex]
	
	\toprule
	ADD-S~~~~ADD-S (AUC) &\multicolumn{2}{c}{mug}  &\multicolumn{2}{c}{skateboard}  &\multicolumn{2}{c}{table}  &\multicolumn{2}{c}{vessel}  
	&\multicolumn{2}{c}{ }  &\multicolumn{2}{c|}{ } \\
	\midrule
	KeyPo (sim) & $0.00$ & $2.43$  & $4.00$ & $11.30$  & $0.00$ & $3.13$  & $0.00$ & $22.21$   & & & & \\
	KeyPo (sim) + ICP & $0.00$ & $2.40$  & $2.00$ & $9.80$  & $0.00$ & $0.00$  & $2.00$ & $23.20$   & & & &  \\
	KeyPo (sim) + RANSAC + ICP & $0.00$ & $0.00$  & $0.00$ & $0.00$  & $0.00$ & $0.00$  & $0.00$ & $2.56$ & & & &   \\
	KeyPo (sim) + Corr. & $44.00$ & $38.44$  & $70.00$ & $59.17$  & $80.00$ & $53.54$  & $12.00$ & $21.90$  & & & &  \\
	KeyPo (sim) + Corr. + ICP & $44.00$ & $40.70$  & $70.00$ & $57.51$  & $80.00$ & $54.14$  & $8.00$ & $20.82$  & & & &  \\
	C-3PO & $68.00$ & $53.55$  & $\textbf{100.00}$ & $\underline{82.90}$  & $\underline{98.00}$ & $\underline{67.69}$  & $\textbf{100.00}$ & $78.21$  & & & &  \\
	C-3PO (oc=1, nd=1) & $\textbf{100.00}$ & $\underline{70.06}$  & $\textbf{100.00}$ & $\underline{82.90}$  & $\textbf{100.00}$ & $\textbf{69.07}$  & $\textbf{100.00}$ & $\underline{80.36}$ & & & &   \\
	DeepGMR & $0.00$ & $2.00$  & $0.00$ & $0.00$  & $0.00$ & $0.00$  & $0.00$ & $0.00$   & & & & \\
	PointNetLK & $0.00$ & $2.00$  & $2.00$ & $5.14$  & $4.00$ & $5.27$  & $2.00$ & $3.57$  & & & &  \\
	FPFH + TEASER++ & $78.00$ & $48.67$  & $38.00$ & $40.33$  & $48.00$ & $39.23$  & $34.00$ & $30.52$ & & & &   \\
	EquiPose & -- &  --  & -- &  --  & -- &  --  & -- &  --  & & & &  \\
	KeyPo (real) & $\textbf{100.00}$ & $\textbf{72.80}$  & $\textbf{100.00}$ & $\textbf{84.02}$  & $54.00$ & $44.67$  & $\textbf{100.00}$ & $\textbf{85.00}$  & & & &  \\
	\bottomrule
	
\end{tabular}
 \end{table*}

\begin{table*}
\centering
\caption{Evaluation of \name and baselines for the YCB experiment.}
\label{tab:expt_ycb_extra01}

\begin{tabular}{|l|rr|rr|rr|rr|rr|rr|}
 
	\toprule
	ADD-S~~~~ADD-S (AUC) &\multicolumn{2}{c}{chips can}  &\multicolumn{2}{c}{master chef can}  &\multicolumn{2}{c}{cracker box}  &\multicolumn{2}{c}{sugar box}  &\multicolumn{2}{c}{tomato soup can}  &\multicolumn{2}{c|}{mustard bottle}  \\
	\midrule
	KeyPo (sim) & $0.00$ & $3.84$  & $0.00$ & $2.14$  & $0.00$ & $2.00$  & $0.00$ & $2.92$  & $0.00$ & $3.60$  & $0.00$ & $15.15$    \\
	KeyPo (sim) + ICP & $0.00$ & $4.47$  & $0.00$ & $2.19$  & $0.00$ & $2.00$  & $0.00$ & $2.64$  & $0.00$ & $3.14$  & $0.00$ & $13.97$    \\
	KeyPo (sim) + RANSAC + ICP & $0.00$ & $3.54$  & $0.00$ & $2.37$  & $0.00$ & $2.41$  & $0.00$ & $2.93$  & $2.00$ & $4.68$  & $0.00$ & $3.18$    \\
	KeyPo (sim) + Corr. & $50.00$ & $41.68$  & $24.00$ & $40.30$  & $38.00$ & $36.18$  & $44.00$ & $43.09$  & $48.00$ & $60.51$  & $40.00$ & $48.63$    \\
	KeyPo (sim) + Corr. + ICP & $58.00$ & $47.90$  & $20.00$ & $35.79$  & $36.00$ & $33.56$  & $40.00$ & $41.27$  & $48.00$ & $59.88$  & $46.00$ & $52.73$    \\
	C-3PO & $98.00$ & $78.48$  & $\textbf{100.00}$ & $\underline{82.36}$  & $68.00$ & $54.24$  & $\underline{58.00}$ & $52.03$  & $\textbf{100.00}$ & $\underline{87.49}$  & $\textbf{100.00}$ & $\underline{84.04}$    \\
	C-3PO (oc=1, nd=1) & $\textbf{100.00}$ & $\underline{78.78}$  & $\textbf{100.00}$ & $\textbf{83.01}$  & $\textbf{100.00}$ & $\textbf{81.08}$  & $\textbf{100.00}$ & $\textbf{87.43}$  & $\textbf{100.00}$ & $\textbf{88.48}$  & $\textbf{100.00}$ & $\textbf{85.05}$    \\
	DeepGMR & $0.00$ & $7.62$  & $0.00$ & $7.09$  & $0.00$ & $0.00$  & $0.00$ & $15.08$  & $0.00$ & $23.86$  & $0.00$ & $16.51$    \\
	PointNetLK & $0.00$ & $0.00$  & $0.00$ & $0.00$  & $0.00$ & $0.00$  & $0.00$ & $0.22$  & $0.00$ & $1.23$  & $0.00$ & $3.72$    \\
	FPFH + TEASER++ & $0.17$ & $5.90$  & $1.33$ & $11.19$  & $0.17$ & $4.91$  & $1.00$ & $15.92$  & $0.83$ & $17.93$  & $0.33$ & $13.23$    \\
	KeyPo (real) & $\textbf{100.00}$ & $\textbf{83.67}$  & $50.00$ & $61.99$  & $\underline{98.00}$ & $\underline{79.34}$  & $52.00$ & $\underline{61.58}$  & $100.00$ & $85.91$  & $\textbf{100.00}$ & $74.12$    \\
	\bottomrule
	
	\addlinespace[2ex]
	
 	\toprule
 	ADD-S~~~~ADD-S (AUC) &\multicolumn{2}{c}{tuna fish can}  &\multicolumn{2}{c}{pudding box}  &\multicolumn{2}{c}{gelatin box}  &\multicolumn{2}{c}{potted meat can}  &\multicolumn{2}{c}{banana}  &\multicolumn{2}{c|}{bleach cleanser}  \\
 	\midrule
 	KeyPo (sim) & $0.00$ & $33.57$  & $0.00$ & $9.84$  & $14.00$ & $44.27$  & $0.00$ & $2.34$  & $2.00$ & $27.29$  & $0.00$ & $0.00$    \\
 	KeyPo (sim) + ICP & $0.00$ & $33.65$  & $0.00$ & $10.60$  & $14.00$ & $44.58$  & $0.00$ & $2.25$  & $4.00$ & $26.09$  & $0.00$ & $0.00$    \\
 	KeyPo (sim) + RANSAC + ICP & $0.00$ & $11.52$  & $0.00$ & $4.19$  & $0.00$ & $9.61$  & $0.00$ & $4.15$  & $2.00$ & $11.54$  & $0.00$ & $0.00$    \\
 	KeyPo (sim) + Corr. & $46.00$ & $58.06$  & $38.00$ & $36.52$  & $72.00$ & $68.30$  & $34.00$ & $38.01$  & $72.00$ & $68.06$  & $34.00$ & $43.53$    \\
 	KeyPo (sim) + Corr. + ICP & $50.00$ & $59.79$  & $42.00$ & $38.64$  & $68.00$ & $66.45$  & $36.00$ & $37.46$  & $80.00$ & $73.20$  & $34.00$ & $41.50$    \\
 	C-3PO & $\textbf{100.00}$ & $\textbf{87.12}$  & $50.00$ & $53.82$  & $\underline{82.00}$ & $\underline{78.07}$  & $54.00$ & $\underline{72.17}$  & $\textbf{100.00}$ & $85.97$  & $82.00$ & $74.56$    \\
 	C-3PO (oc=1, nd=1) & $\textbf{100.00}$ & $\underline{86.92}$  & $\textbf{100.00}$ & $\textbf{89.73}$  & $\textbf{100.00}$ & $\textbf{89.26}$  & $\textbf{100.00}$ & $\textbf{88.81}$  & $\textbf{100.00}$ & $\underline{87.34}$  & $\underline{84.44}$ & $\underline{75.89}$    \\
 	DeepGMR & $0.00$ & $34.26$  & $0.00$ & $19.37$  & $0.00$ & $22.02$  & $0.00$ & $31.92$  & $2.50$ & $30.49$  & $0.00$ & $9.56$    \\
 	PointNetLK & $0.00$ & $0.17$  & $0.00$ & $0.86$  & $0.00$ & $0.61$  & $0.00$ & $3.51$  & $22.17$ & $28.37$  & $0.00$ & $2.81$    \\
 	FPFH + TEASER++ & $2.00$ & $19.16$  & $1.17$ & $22.01$  & $9.70$ & $23.56$  & $0.83$ & $16.22$  & $4.83$ & $24.43$  & $0.00$ & $10.81$    \\
 	KeyPo (real) & $\textbf{100.00}$ & $78.11$  & $\underline{74.00}$ & $\underline{71.57}$  & $80.00$ & $72.49$  & $\underline{62.00}$ & $64.85$  & $\textbf{100.00}$ & $\textbf{89.92}$  & $\textbf{100.00}$ & $\textbf{85.00}$    \\
 	\bottomrule
	
	\addlinespace[2ex]
	
 	\toprule
 	ADD-S~~~~ADD-S (AUC) &\multicolumn{2}{c}{power drill}  &\multicolumn{2}{c}{wood block}  &\multicolumn{2}{c}{scissors}  &\multicolumn{2}{c}{large clamp}  &\multicolumn{2}{c}{extra large clamp}  &\multicolumn{2}{c|}{foam brick}  \\
 	\midrule
 	KeyPo (sim) & $2.00$ & $22.45$  & $0.00$ & $7.16$  & $34.00$ & $53.22$  & $0.00$ & $29.48$  & $0.00$ & $11.73$  & $6.00$ & $36.44$    \\
 	KeyPo (sim) + ICP & $2.00$ & $21.16$  & $0.00$ & $7.12$  & $36.00$ & $52.55$  & $0.00$ & $28.23$  & $2.00$ & $10.79$  & $2.00$ & $38.94$    \\
 	KeyPo (sim) + RANSAC + ICP & $0.00$ & $2.00$  & $0.00$ & $2.94$  & $0.00$ & $8.90$  & $0.00$ & $13.58$  & $0.00$ & $3.62$  & $0.00$ & $0.00$    \\
 	KeyPo (sim) + Corr. & $22.00$ & $44.07$  & $50.00$ & $40.77$  & $92.00$ & $80.59$  & $40.00$ & $50.91$  & $54.00$ & $53.14$  & $86.00$ & $69.81$    \\
 	KeyPo (sim) + Corr. + ICP & $18.00$ & $42.32$  & $48.00$ & $41.11$  & $86.00$ & $79.04$  & $38.00$ & $50.11$  & $52.00$ & $52.49$  & $90.00$ & $72.37$    \\
 	C-3PO & $86.00$ & $81.42$  & $62.00$ & $50.17$  & $\textbf{100.00}$ & $84.42$  & $\underline{98.00}$ & $\underline{85.61}$  & $\textbf{100.00}$ & $\underline{84.22}$  & $\underline{98.00}$ & $\underline{85.33}$    \\
 	C-3PO (oc=1, nd=1) & $\textbf{100.00}$ & $\textbf{87.13}$  & $\textbf{100.00}$ & $\textbf{83.11}$  & $\textbf{100.00}$ & $\underline{86.77}$  & $\textbf{100.00}$ & $\textbf{88.63}$  & $\textbf{100.00}$ & $\textbf{84.80}$  & $\textbf{100.00}$ & $\textbf{89.68}$    \\
 	DeepGMR & $0.00$ & $7.55$  & $0.00$ & $1.71$  & $14.14$ & $35.64$  & $0.00$ & $1.16$  & $0.76$ & $7.06$  & $0.00$ & $48.73$    \\
 	PointNetLK & $0.00$ & $0.17$  & $0.00$ & $0.98$  & $7.07$ & $8.08$  & $0.00$ & $0.42$  & $0.00$ & $0.00$  & $0.00$ & $0.99$    \\
 	FPFH + TEASER++ & $2.00$ & $12.87$  & $0.50$ & $7.64$  & $47.81$ & $50.07$  & $13.43$ & $30.76$  & $5.29$ & $13.20$  & $10.33$ & $35.05$    \\
 	KeyPo (real) & $\textbf{100.00}$ & $\underline{86.87}$  & $\textbf{100.00}$ & $\underline{75.47}$  & $\textbf{100.00}$ & $\textbf{88.51}$  & $96.00$ & $79.51$  & $98.00$ & $82.06$  & $82.00$ & $68.65$    \\
 	\bottomrule
 	
\end{tabular}
 \end{table*}

\newpage
\begin{figure*}
	\begin{subfigure}[b]{0.7\textwidth}
		\begin{tabular}{ll}
			\includegraphics[trim=10 1 50 1,clip,width=0.45\linewidth]{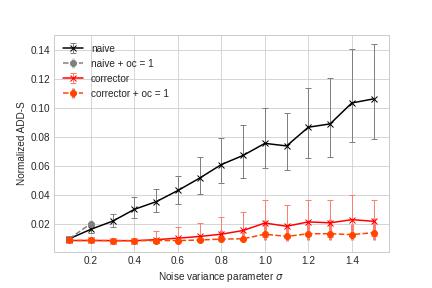} 
			&   \includegraphics[trim=10 1 50 1,clip,width=0.45\linewidth]{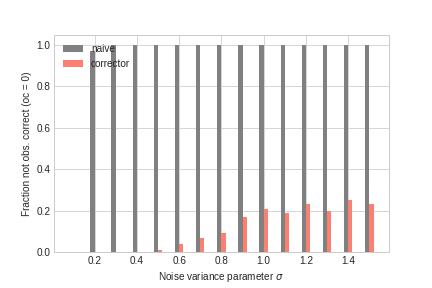} 
		\end{tabular}
		\caption{ADD-S normalized to object diameter (left) and fraction certifiable (right) as a function of the noise variance parameter $\sigma$.}
		\label{fig:airplane}
	\end{subfigure}
	\hfill
	\begin{subfigure}[b]{0.27\textwidth}
		\includegraphics[trim=24 35 50 30,clip,width=0.9\linewidth]{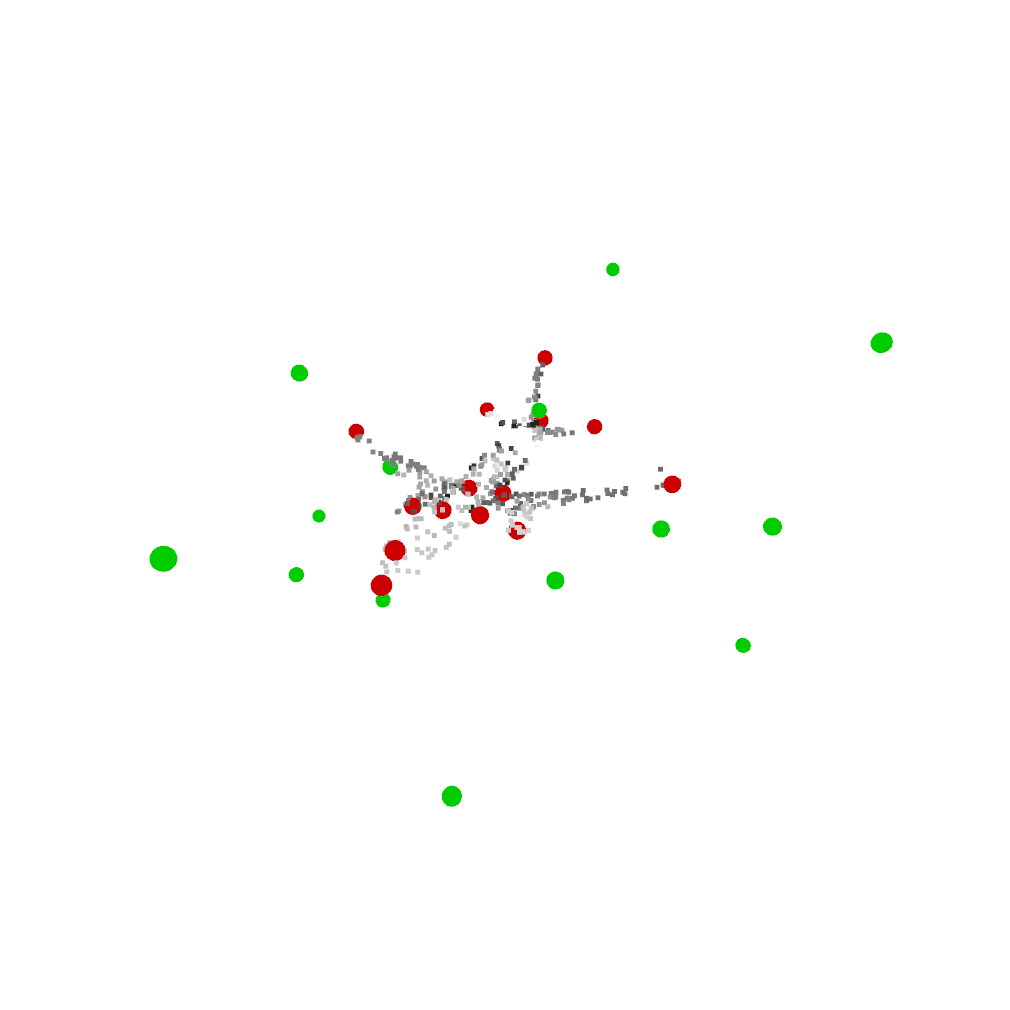} 
		\caption{Perturbed keypoints (green) and true keypoints (red) at $\sigma=0.8$.}
		\label{fig:airplane-fig}
	\end{subfigure}
	
	\begin{subfigure}[b]{0.7\textwidth}
		\begin{tabular}{ll}
			\includegraphics[trim=10 1 50 1,clip,width=0.45\linewidth]{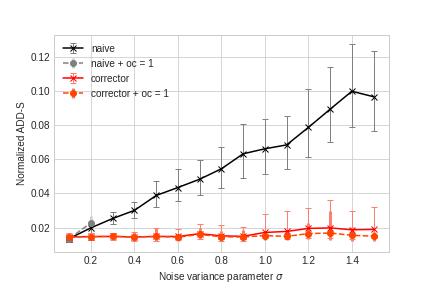} 
			&   \includegraphics[trim=10 1 50 1,clip,width=0.45\linewidth]{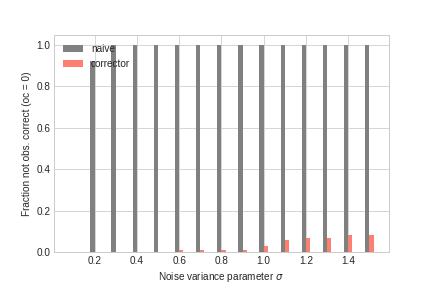} 
		\end{tabular}
		\caption{ADD-S normalized to object diameter (left) and fraction certifiable (right) as a function of the noise variance parameter $\sigma$.}
		\label{fig:bathtub}
	\end{subfigure}
	\hfill
	\begin{subfigure}[b]{0.27\textwidth}
		\includegraphics[trim=24 35 50 30,clip,width=0.9\linewidth]{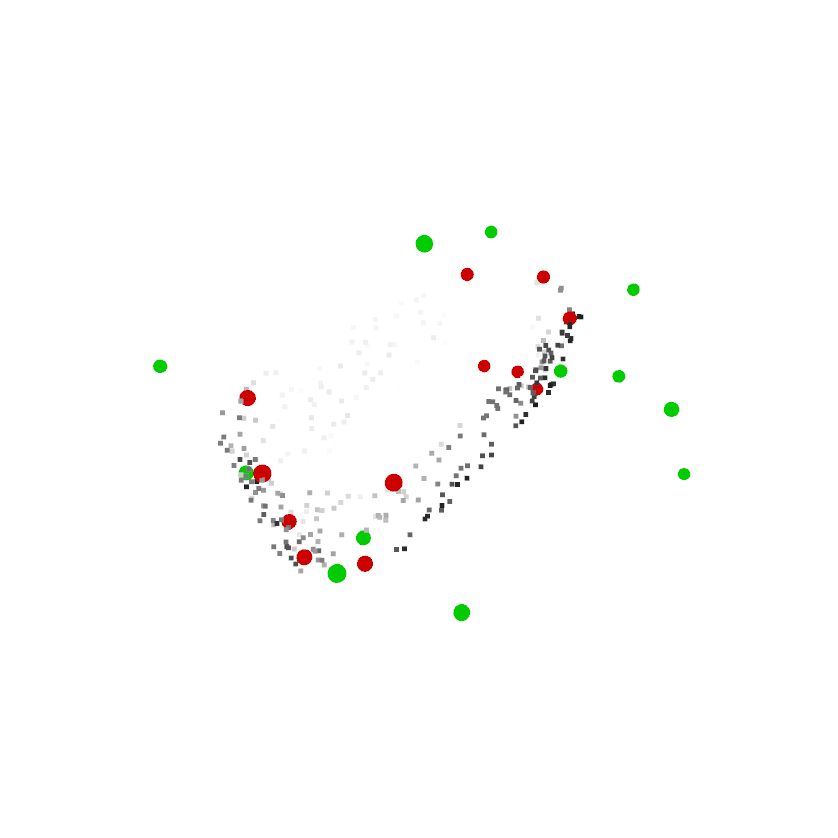} 
		\caption{Perturbed keypoints (green) and true keypoints (red) at $\sigma=0.8$.}
		\label{fig:bathtub-fig}
	\end{subfigure}

	\begin{subfigure}[b]{0.7\textwidth}
		\begin{tabular}{ll}
			\includegraphics[trim=10 1 50 1,clip,width=0.45\linewidth]{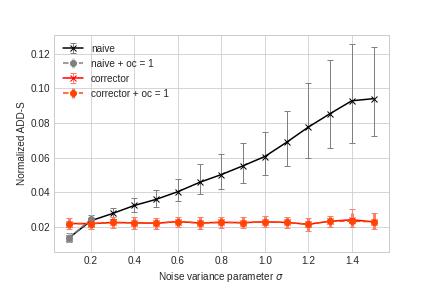} 
			&   \includegraphics[trim=10 1 50 1,clip,width=0.45\linewidth]{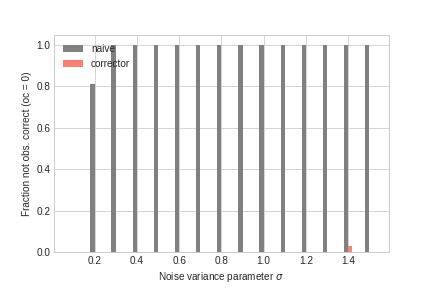} 
		\end{tabular}
		\caption{ADD-S normalized to object diameter (left) and fraction certifiable (right) as a function of the noise variance parameter $\sigma$.}
		\label{fig:bed}
	\end{subfigure}
	\hfill
	\begin{subfigure}[b]{0.27\textwidth}
		\includegraphics[trim=24 35 50 30,clip,width=0.9\linewidth]{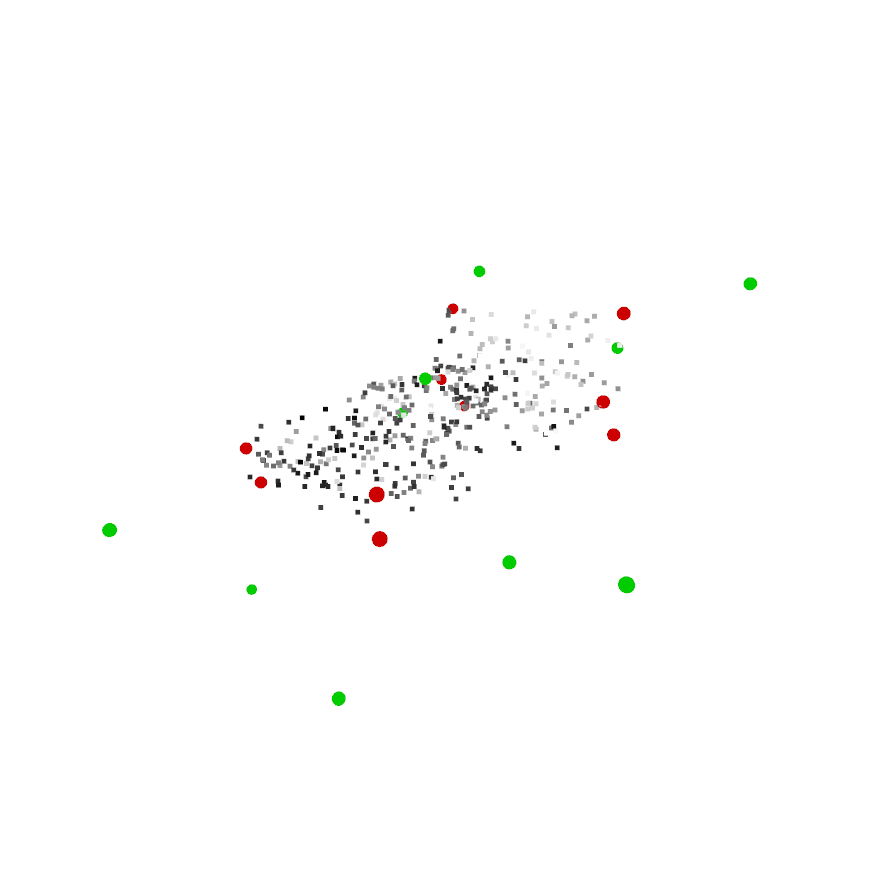} 
		\caption{Perturbed keypoints (green) and true keypoints (red) at $\sigma=0.8$.}
		\label{fig:bed-fig}
	\end{subfigure}

	\begin{subfigure}[b]{0.7\textwidth}
		\begin{tabular}{ll}
			\includegraphics[trim=10 1 50 1,clip,width=0.45\linewidth]{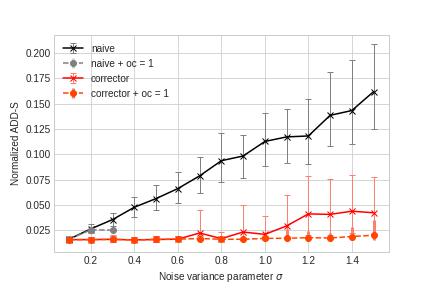} 
			&   \includegraphics[trim=10 1 50 1,clip,width=0.45\linewidth]{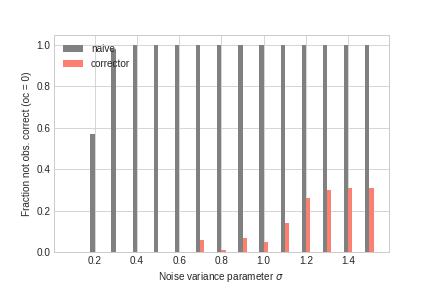} 
		\end{tabular}
		\caption{ADD-S normalized to object diameter (left) and fraction certifiable (right) as a function of the noise variance parameter $\sigma$.}
		\label{fig:cap}
	\end{subfigure}
	\hfill
	\begin{subfigure}[b]{0.27\textwidth}
		\includegraphics[trim=24 35 50 30,clip,width=0.9\linewidth]{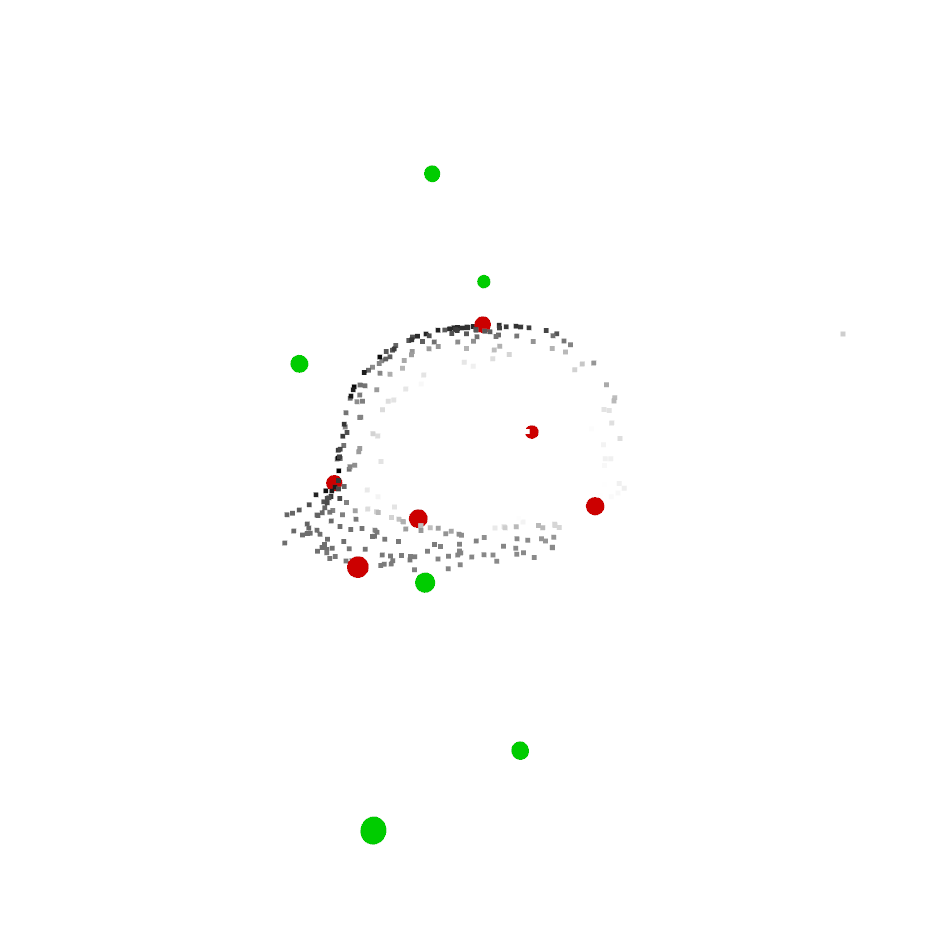} 
		\caption{Perturbed keypoints (green) and true keypoints (red) at $\sigma=0.8$.}
		\label{fig:cap-fig}
	\end{subfigure} 

\caption{ShapeNet objects: airplane (\ref{fig:airplane}-\ref{fig:airplane-fig}), bathtub (\ref{fig:bathtub}-\ref{fig:bathtub-fig}), bed (\ref{fig:bed}-\ref{fig:bed-fig}), cap (\ref{fig:cap}-\ref{fig:cap-fig}).}
\label{fig:corrector-analysis-extra-01}
\end{figure*}

\begin{figure*}
	\begin{subfigure}[b]{0.7\textwidth}
		\begin{tabular}{ll}
			\includegraphics[trim=10 1 50 1,clip,width=0.45\linewidth]{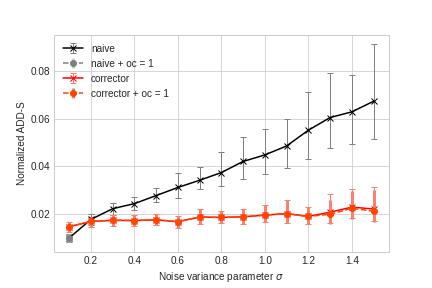} 
			&   \includegraphics[trim=10 1 50 1,clip,width=0.45\linewidth]{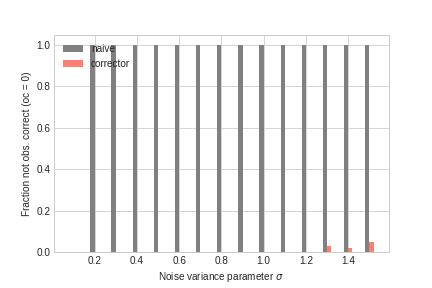} 
		\end{tabular}
		\caption{ADD-S normalized to object diameter (left) and fraction certifiable (right) as a function of the noise variance parameter $\sigma$.}
		\label{fig:car}
	\end{subfigure}
	\hfill
	\begin{subfigure}[b]{0.27\textwidth}
		\includegraphics[trim=24 35 50 30,clip,width=0.9\linewidth]{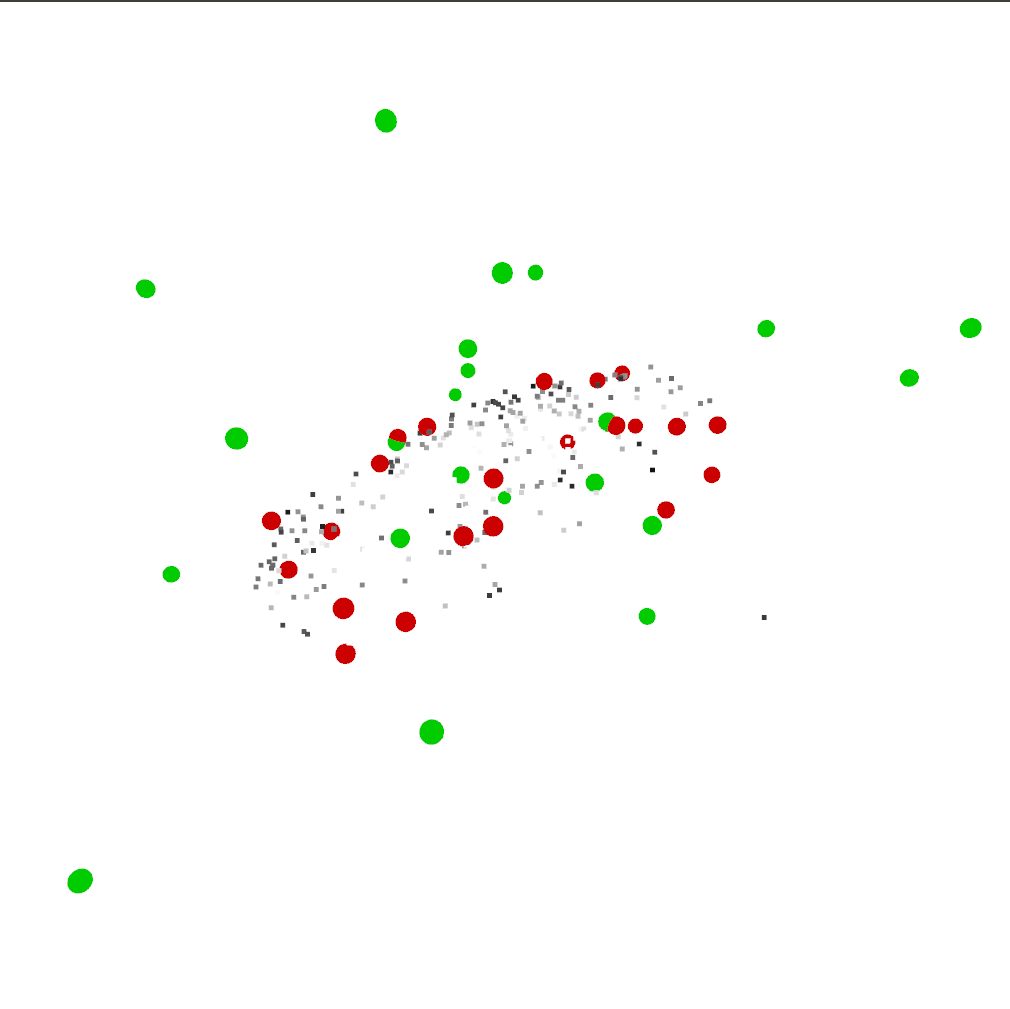} 
		\caption{Perturbed keypoints (green) and true keypoints (red) at $\sigma=0.8$.}
		\label{fig:car-fig}
	\end{subfigure}

	\begin{subfigure}[b]{0.7\textwidth}
		\begin{tabular}{ll}
			\includegraphics[trim=10 1 50 1,clip,width=0.45\linewidth]{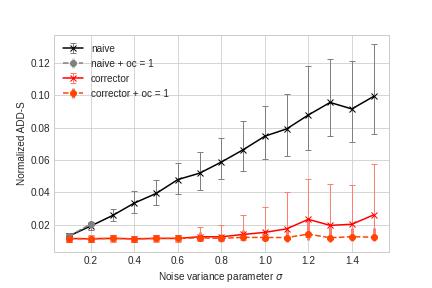} 
			&   \includegraphics[trim=10 1 50 1,clip,width=0.45\linewidth]{chair-cert.jpg} 
		\end{tabular}
		\caption{ADD-S normalized to object diameter (left) and fraction certifiable (right) as a function of the noise variance parameter $\sigma$.}
		\label{fig:chair}
	\end{subfigure}
	\hfill
	\begin{subfigure}[b]{0.27\textwidth}
		\includegraphics[trim=24 35 50 30,clip,width=0.9\linewidth]{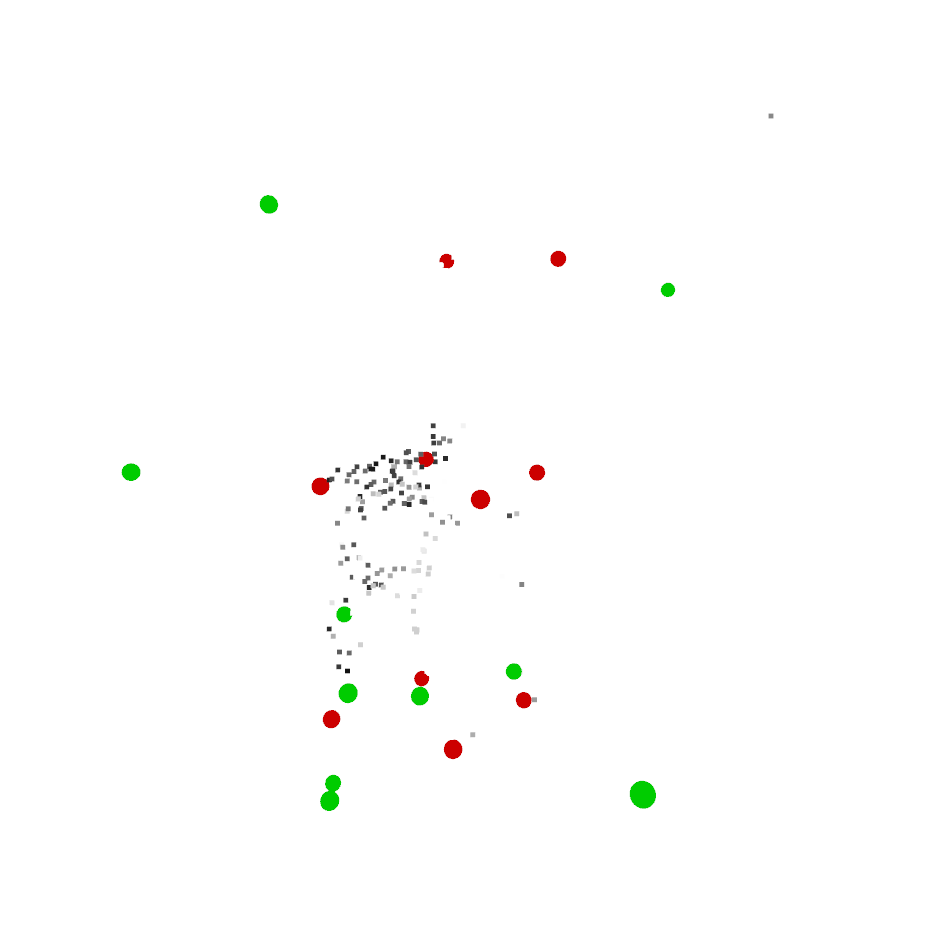} 
		\caption{Perturbed keypoints (green) and true keypoints (red) at $\sigma=0.8$.}
		\label{fig:chair-fig}
	\end{subfigure}

	\begin{subfigure}[b]{0.7\textwidth}
		\begin{tabular}{ll}
			\includegraphics[trim=10 1 50 1,clip,width=0.45\linewidth]{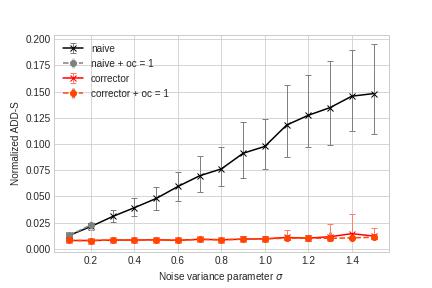} 
			&   \includegraphics[trim=10 1 50 1,clip,width=0.45\linewidth]{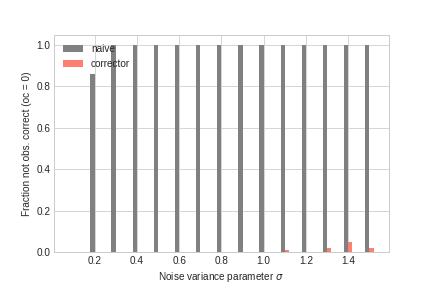} 
		\end{tabular}
		\caption{ADD-S normalized to object diameter (left) and fraction certifiable (right) as a function of the noise variance parameter $\sigma$.}
		\label{fig:guitar}
	\end{subfigure}
	\hfill
	\begin{subfigure}[b]{0.27\textwidth}
		\includegraphics[trim=24 35 50 30,clip,width=0.9\linewidth]{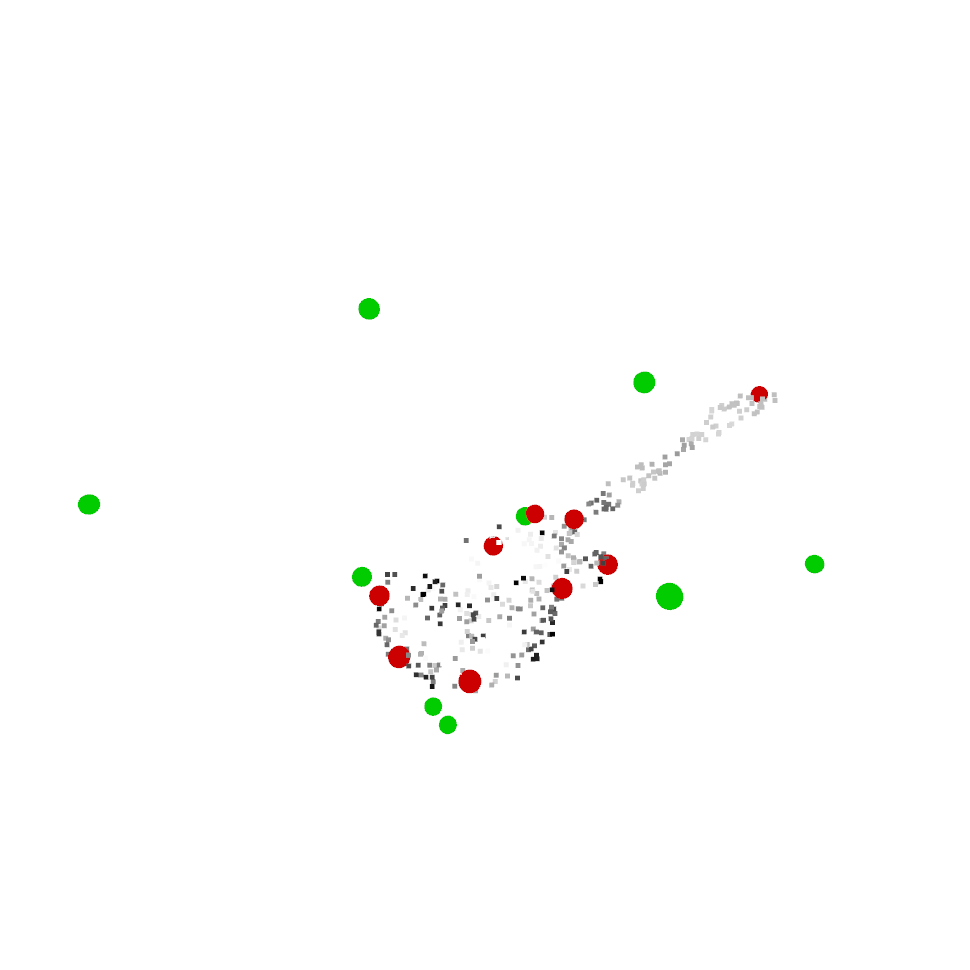} 
		\caption{Perturbed keypoints (green) and true keypoints (red) at $\sigma=0.8$.}
		\label{fig:guitar-fig}
	\end{subfigure}

	\begin{subfigure}[b]{0.7\textwidth}
		\begin{tabular}{ll}
			\includegraphics[trim=10 1 50 1,clip,width=0.45\linewidth]{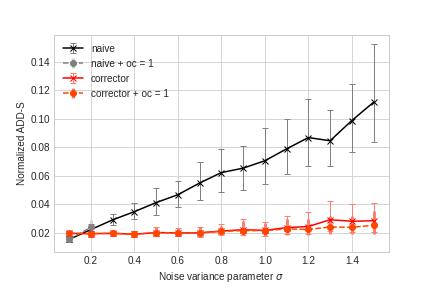} 
			&   \includegraphics[trim=10 1 50 1,clip,width=0.45\linewidth]{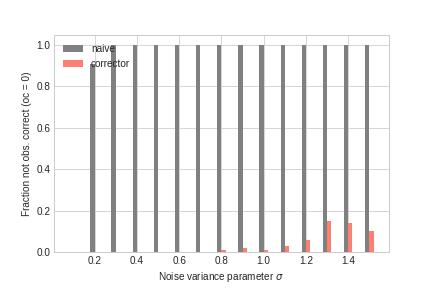} 
		\end{tabular}
		\caption{ADD-S normalized to object diameter (left) and fraction certifiable (right) as a function of the noise variance parameter $\sigma$.}
		\label{fig:helmet}
	\end{subfigure}
	\hfill
	\begin{subfigure}[b]{0.27\textwidth}
		\includegraphics[trim=24 35 50 30,clip,width=0.9\linewidth]{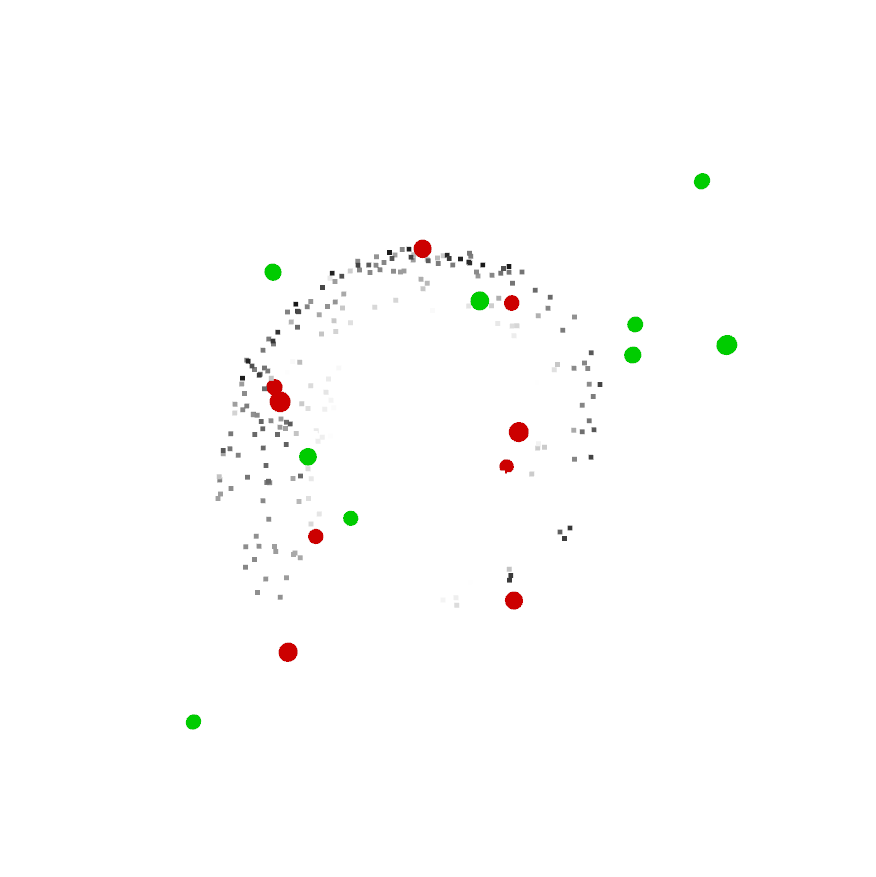} 
		\caption{Perturbed keypoints (green) and true keypoints (red) at $\sigma=0.8$.}
		\label{fig:helmet-fig}
	\end{subfigure}

\caption{ShapeNet objects: car (\ref{fig:car}-\ref{fig:car-fig}), chair (\ref{fig:chair}-\ref{fig:chair-fig}), guitar (\ref{fig:guitar}-\ref{fig:guitar-fig}), helmet (\ref{fig:helmet}-\ref{fig:helmet-fig}).}
\label{fig:corrector-analysis-extra-02}
\end{figure*}

\begin{figure*}
	\begin{subfigure}[b]{0.7\textwidth}
		\begin{tabular}{ll}
			\includegraphics[trim=10 1 50 1,clip,width=0.45\linewidth]{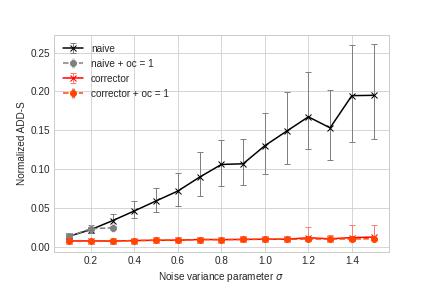} 
			&   \includegraphics[trim=10 1 50 1,clip,width=0.45\linewidth]{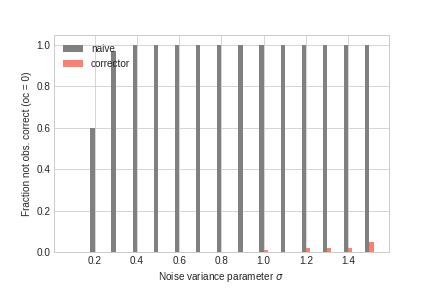} 
		\end{tabular}
		\caption{ADD-S normalized to object diameter (left) and fraction certifiable (right) as a function of the noise variance parameter $\sigma$.}
		\label{fig:knife}
	\end{subfigure}
	\hfill
	\begin{subfigure}[b]{0.27\textwidth}
		\includegraphics[trim=24 35 50 30,clip,width=0.9\linewidth]{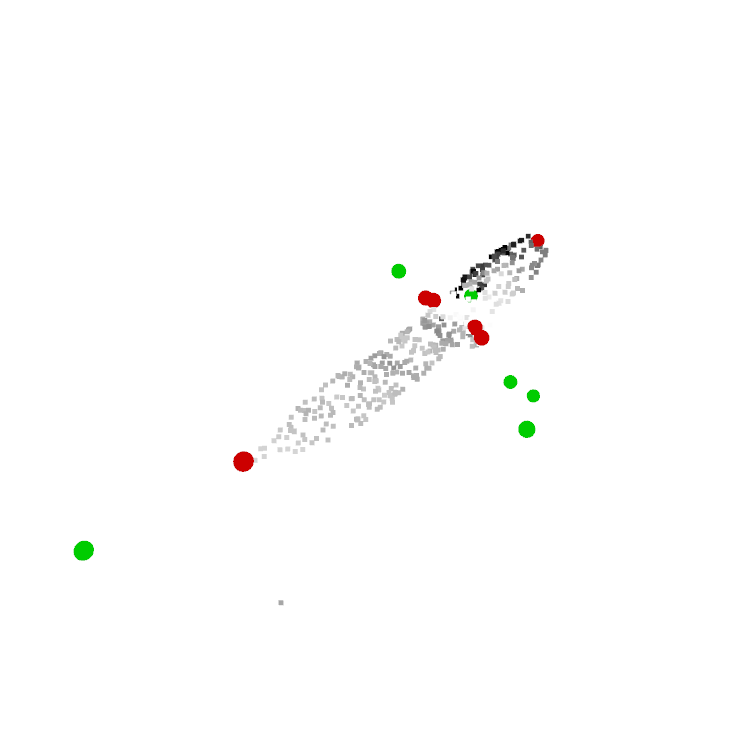} 
		\caption{Perturbed keypoints (green) and true keypoints (red) at $\sigma=0.8$.}
		\label{fig:knife-fig}
	\end{subfigure}

	\begin{subfigure}[b]{0.7\textwidth}
		\begin{tabular}{ll}
			\includegraphics[trim=10 1 50 1,clip,width=0.45\linewidth]{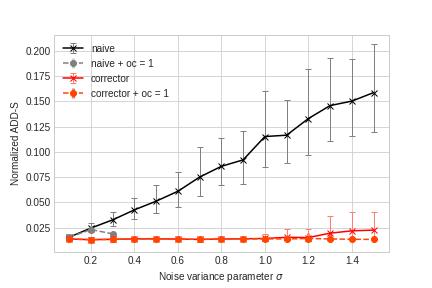} 
			&   \includegraphics[trim=10 1 50 1,clip,width=0.45\linewidth]{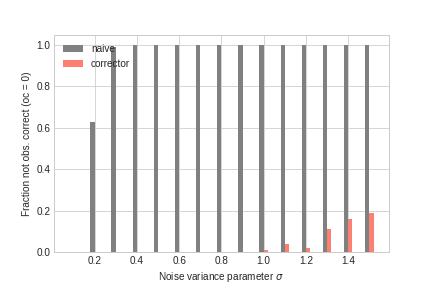} 
		\end{tabular}
		\caption{ADD-S normalized to object diameter (left) and fraction certifiable (right) as a function of the noise variance parameter $\sigma$.}
		\label{fig:laptop}
	\end{subfigure}
	\hfill
	\begin{subfigure}[b]{0.27\textwidth}
		\includegraphics[trim=24 35 50 30,clip,width=0.9\linewidth]{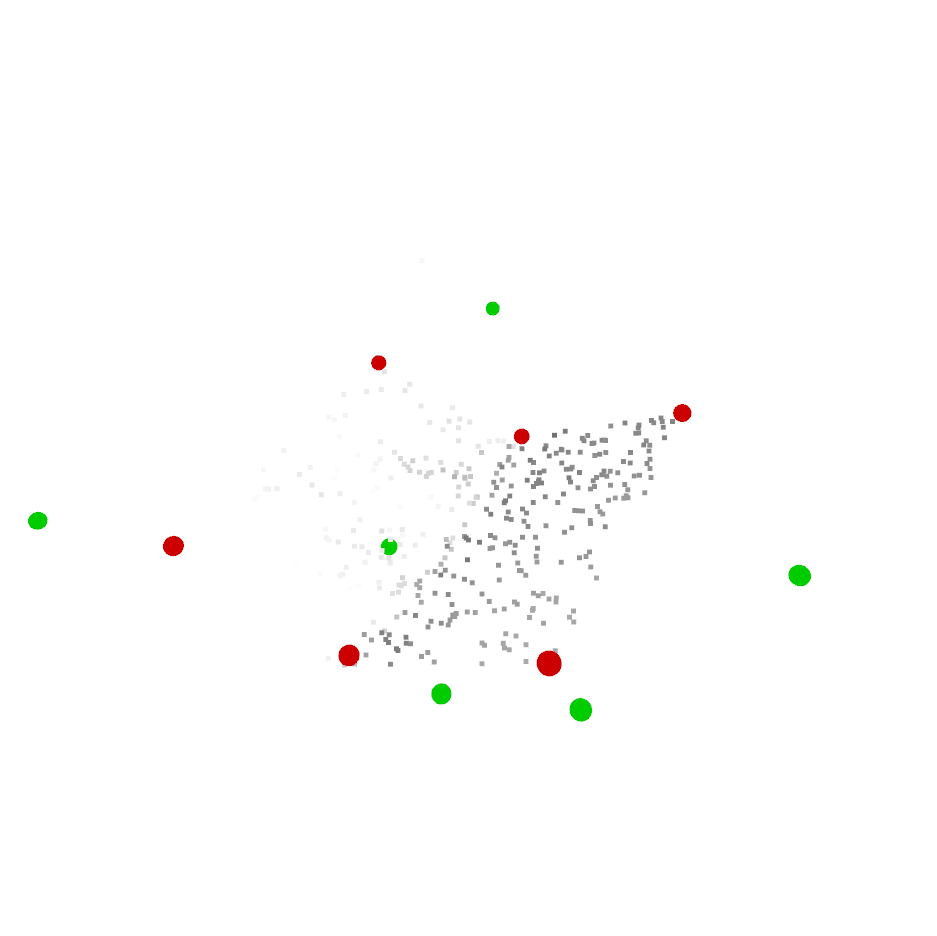} 
		\caption{Perturbed keypoints (green) and true keypoints (red) at $\sigma=0.8$.}
		\label{fig:laptop-fig}
	\end{subfigure}

	\begin{subfigure}[b]{0.7\textwidth}
		\begin{tabular}{ll}
			\includegraphics[trim=10 1 50 1,clip,width=0.45\linewidth]{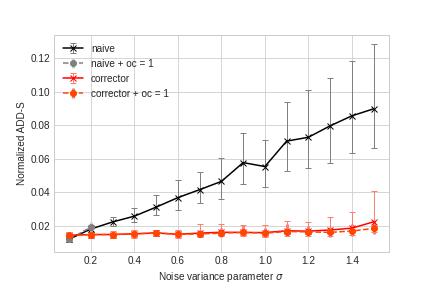} 
			&   \includegraphics[trim=10 1 50 1,clip,width=0.45\linewidth]{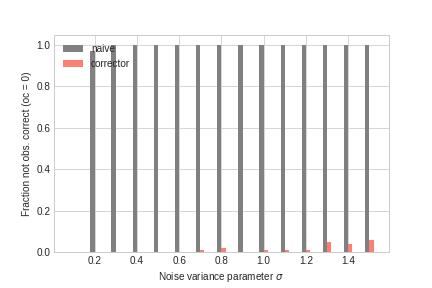} 
		\end{tabular}
		\caption{ADD-S normalized to object diameter (left) and fraction certifiable (right) as a function of the noise variance parameter $\sigma$.}
		\label{fig:motorcycle}
	\end{subfigure}
	\hfill
	\begin{subfigure}[b]{0.27\textwidth}
		\includegraphics[trim=24 35 50 30,clip,width=0.9\linewidth]{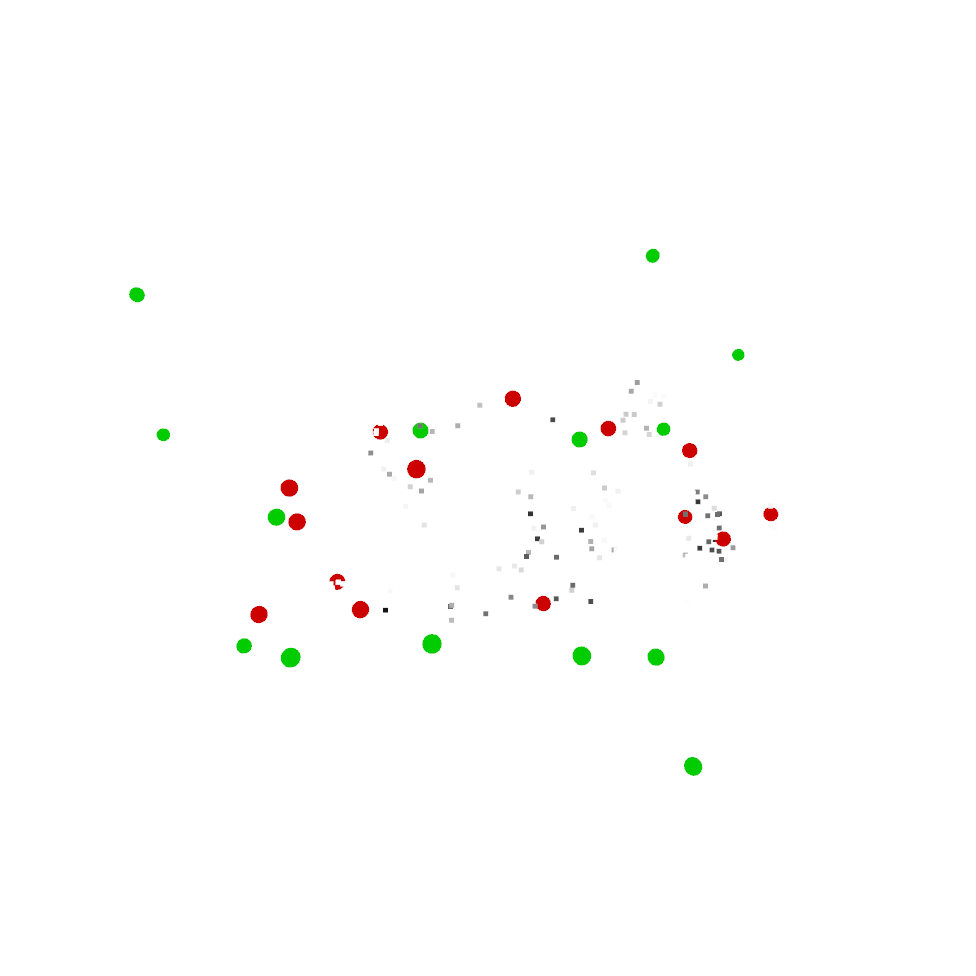} 
		\caption{Perturbed keypoints (green) and true keypoints (red) at $\sigma=0.8$.}
		\label{fig:motorcycle-fig}
	\end{subfigure}

	\begin{subfigure}[b]{0.7\textwidth}
		\begin{tabular}{ll}
			\includegraphics[trim=10 1 50 1,clip,width=0.45\linewidth]{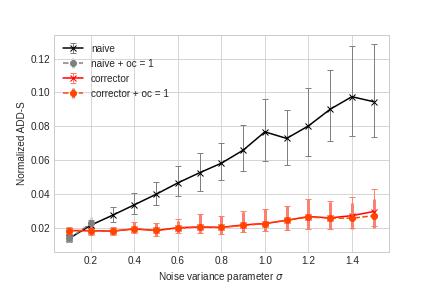} 
			&   \includegraphics[trim=10 1 50 1,clip,width=0.45\linewidth]{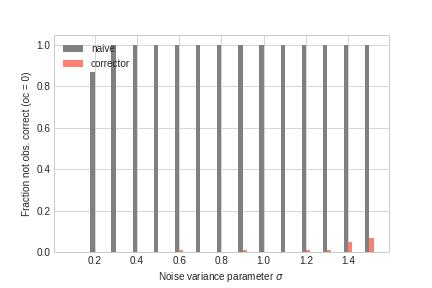} 
		\end{tabular}
		\caption{ADD-S normalized to object diameter (left) and fraction certifiable (right) as a function of the noise variance parameter $\sigma$.}
		\label{fig:mug}
	\end{subfigure}
	\hfill
	\begin{subfigure}[b]{0.27\textwidth}
		\includegraphics[trim=24 35 50 30,clip,width=0.9\linewidth]{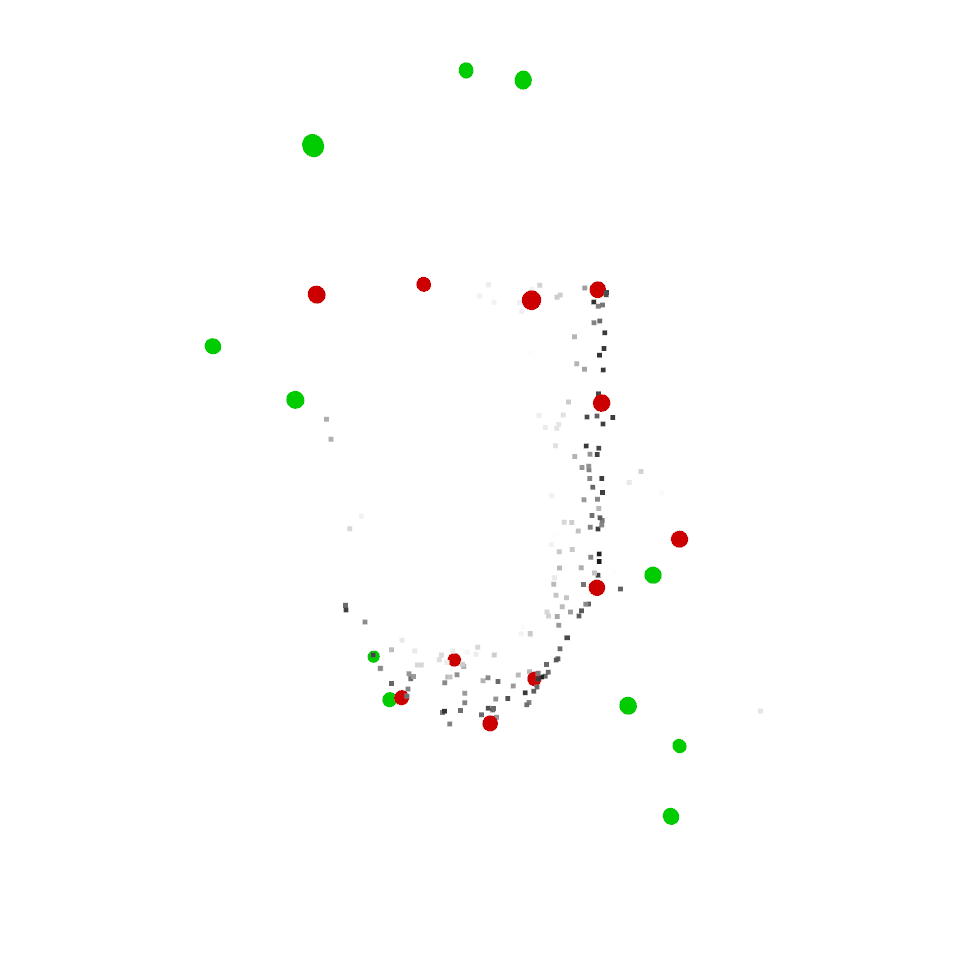} 
		\caption{Perturbed keypoints (green) and true keypoints (red) at $\sigma=0.8$.}
		\label{fig:mug-fig}
	\end{subfigure}

	\caption{ShapeNet objects: knife (\ref{fig:knife}-\ref{fig:knife-fig}), laptop (\ref{fig:laptop}-\ref{fig:laptop-fig}), motorcycle (\ref{fig:motorcycle}-\ref{fig:motorcycle-fig}), mug (\ref{fig:mug}-\ref{fig:mug-fig}).}
	\label{fig:corrector-analysis-extra-03}
\end{figure*}

\begin{figure*}
	
	\begin{subfigure}[b]{0.7\textwidth}
		\begin{tabular}{ll}
			\includegraphics[trim=10 1 50 1,clip,width=0.45\linewidth]{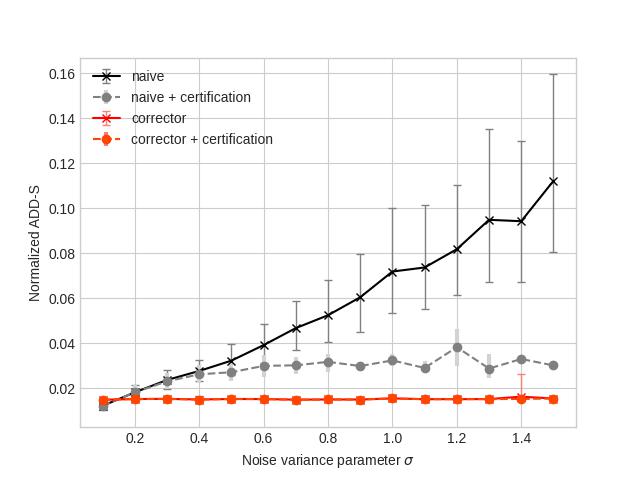} 
			&   \includegraphics[trim=10 1 50 1,clip,width=0.45\linewidth]{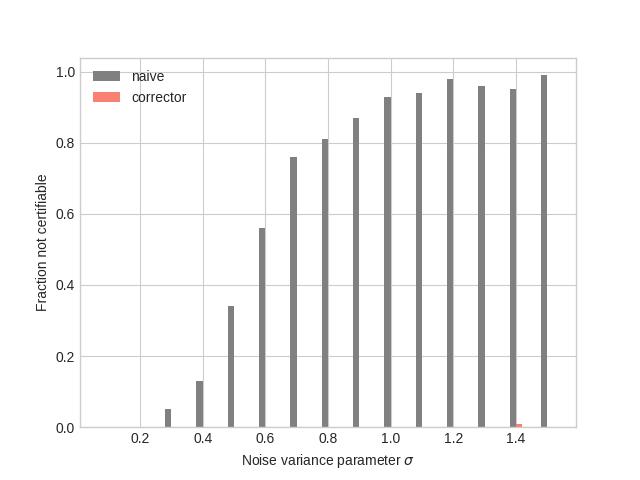} 
		\end{tabular}
		\caption{ADD-S normalized to object diameter (left) and fraction certifiable (right) as a function of the noise variance parameter $\sigma$.}
		\label{fig:bottle}
	\end{subfigure}
	\hfill
	\begin{subfigure}[b]{0.27\textwidth}
		\includegraphics[trim=24 35 50 30,clip,width=0.9\linewidth]{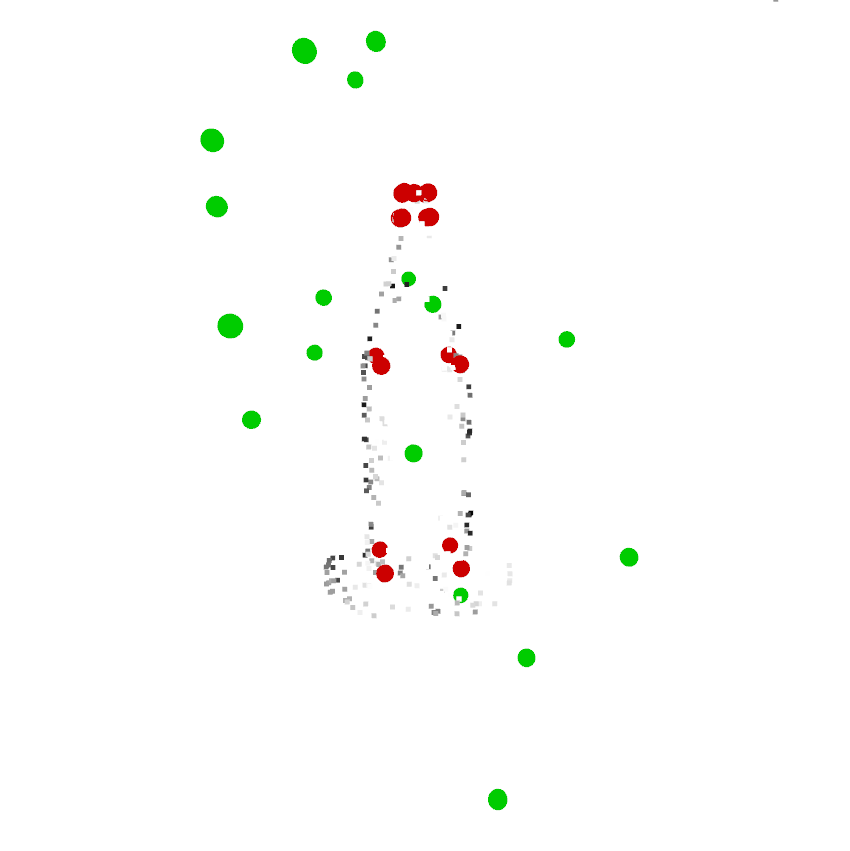} 
		\caption{Perturbed keypoints (green) and true keypoints (red) at $\sigma=0.8$.}
		\label{fig:bottle-fig}
	\end{subfigure}

	\begin{subfigure}[b]{0.7\textwidth}
		\begin{tabular}{ll}
			\includegraphics[trim=10 1 50 1,clip,width=0.45\linewidth]{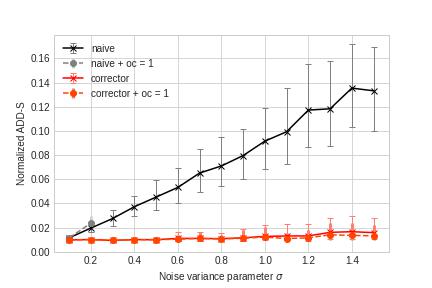} 
			&   \includegraphics[trim=10 1 50 1,clip,width=0.45\linewidth]{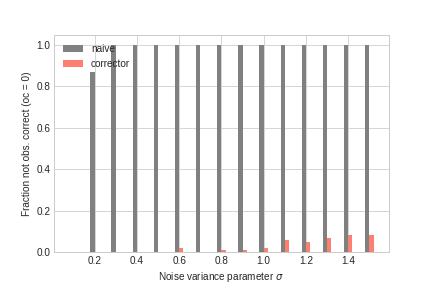} 
		\end{tabular}
		\caption{ADD-S normalized to object diameter (left) and fraction certifiable (right) as a function of the noise variance parameter $\sigma$.}
		\label{fig:skateboard}
	\end{subfigure}
	\hfill
	\begin{subfigure}[b]{0.27\textwidth}
		\includegraphics[trim=24 35 50 30,clip,width=0.9\linewidth]{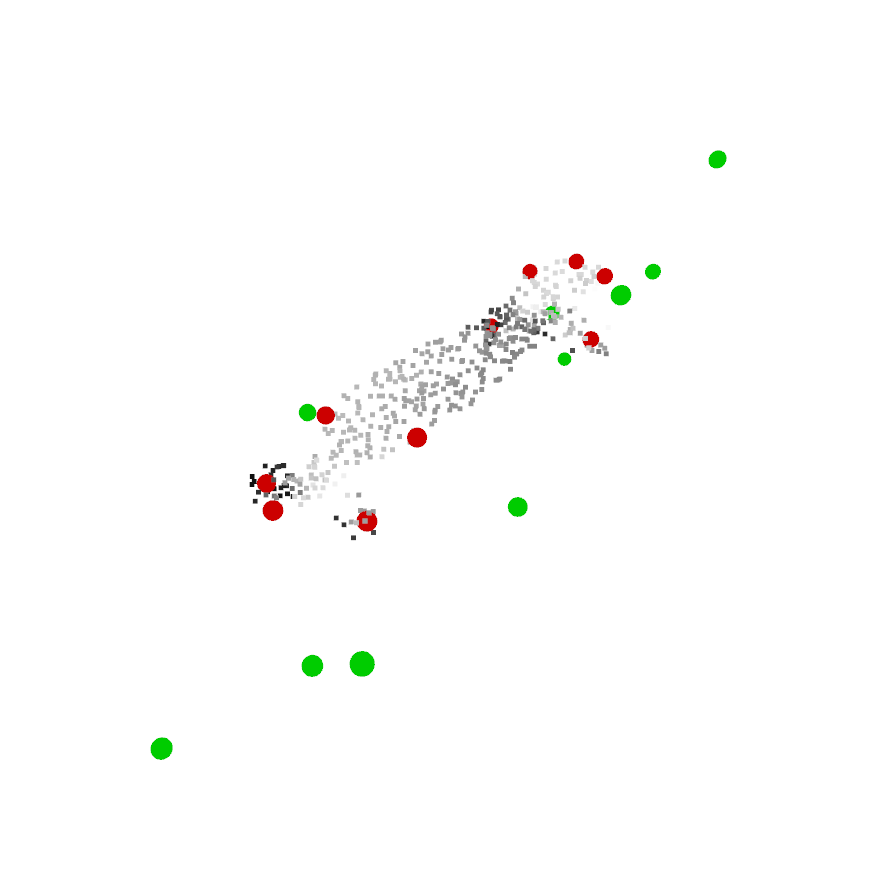} 
		\caption{Perturbed keypoints (green) and true keypoints (red) at $\sigma=0.8$.}
		\label{fig:skateboard-fig}
	\end{subfigure}

	\begin{subfigure}[b]{0.7\textwidth}
		\begin{tabular}{ll}
			\includegraphics[trim=10 1 50 1,clip,width=0.45\linewidth]{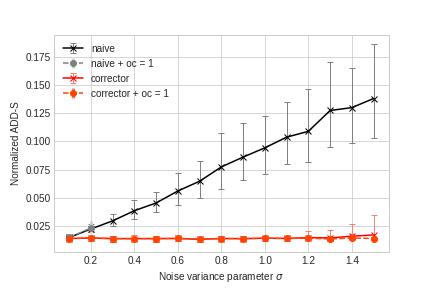} 
			&   \includegraphics[trim=10 1 50 1,clip,width=0.45\linewidth]{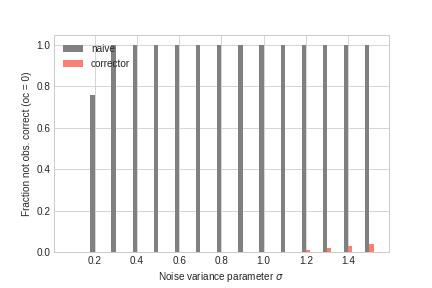} 
		\end{tabular}
		\caption{ADD-S normalized to object diameter (left) and fraction certifiable (right) as a function of the noise variance parameter $\sigma$.}
		\label{fig:table}
	\end{subfigure}
	\hfill
	\begin{subfigure}[b]{0.27\textwidth}
		\includegraphics[trim=24 35 50 30,clip,width=0.9\linewidth]{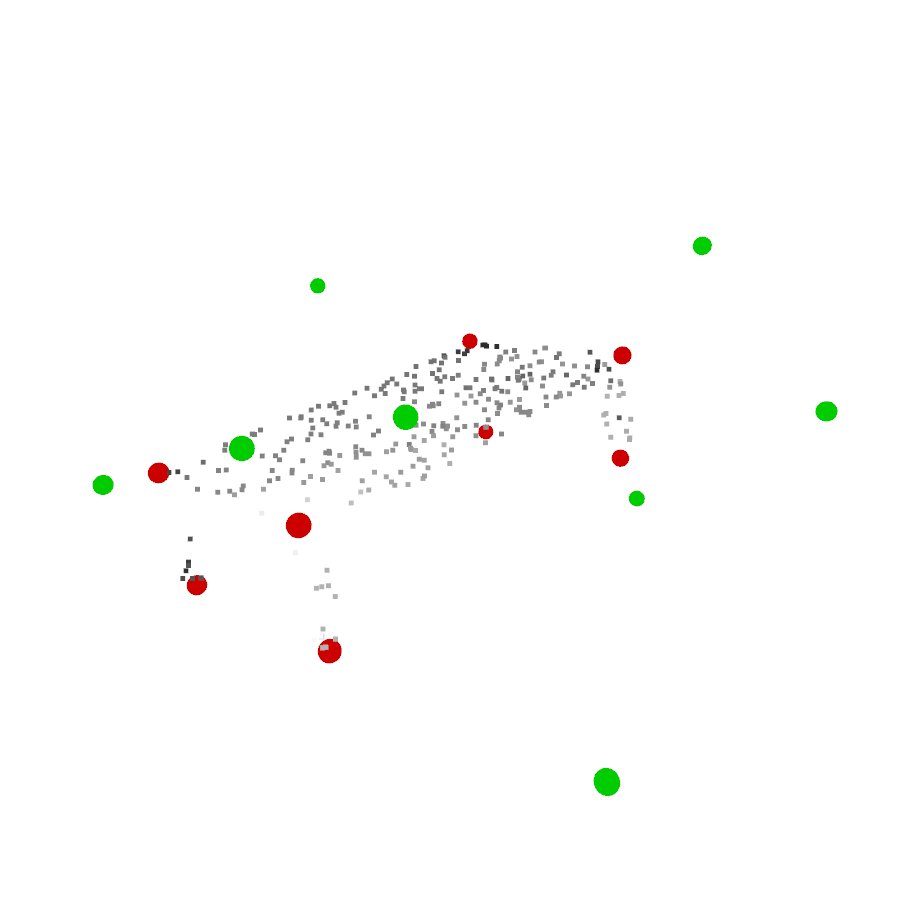} 
		\caption{Perturbed keypoints (green) and true keypoints (red) at $\sigma=0.8$.}
		\label{fig:table-fig}
	\end{subfigure}

	\begin{subfigure}[b]{0.7\textwidth}
		\begin{tabular}{ll}
			\includegraphics[trim=10 1 50 1,clip,width=0.45\linewidth]{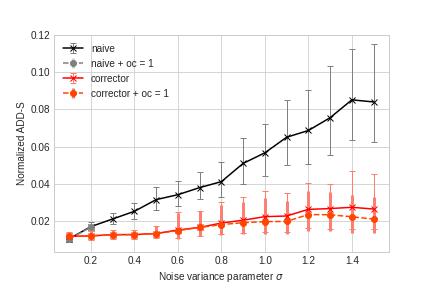} 
			&   \includegraphics[trim=10 1 50 1,clip,width=0.45\linewidth]{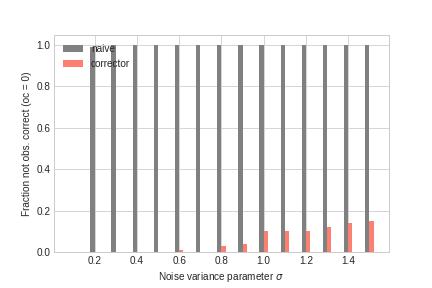} 
		\end{tabular}
		\caption{ADD-S normalized to object diameter (left) and fraction certifiable (right) as a function of the noise variance parameter $\sigma$.}
		\label{fig:vessel}
	\end{subfigure}
	\hfill
	\begin{subfigure}[b]{0.27\textwidth}
		\includegraphics[trim=24 35 50 30,clip,width=0.9\linewidth]{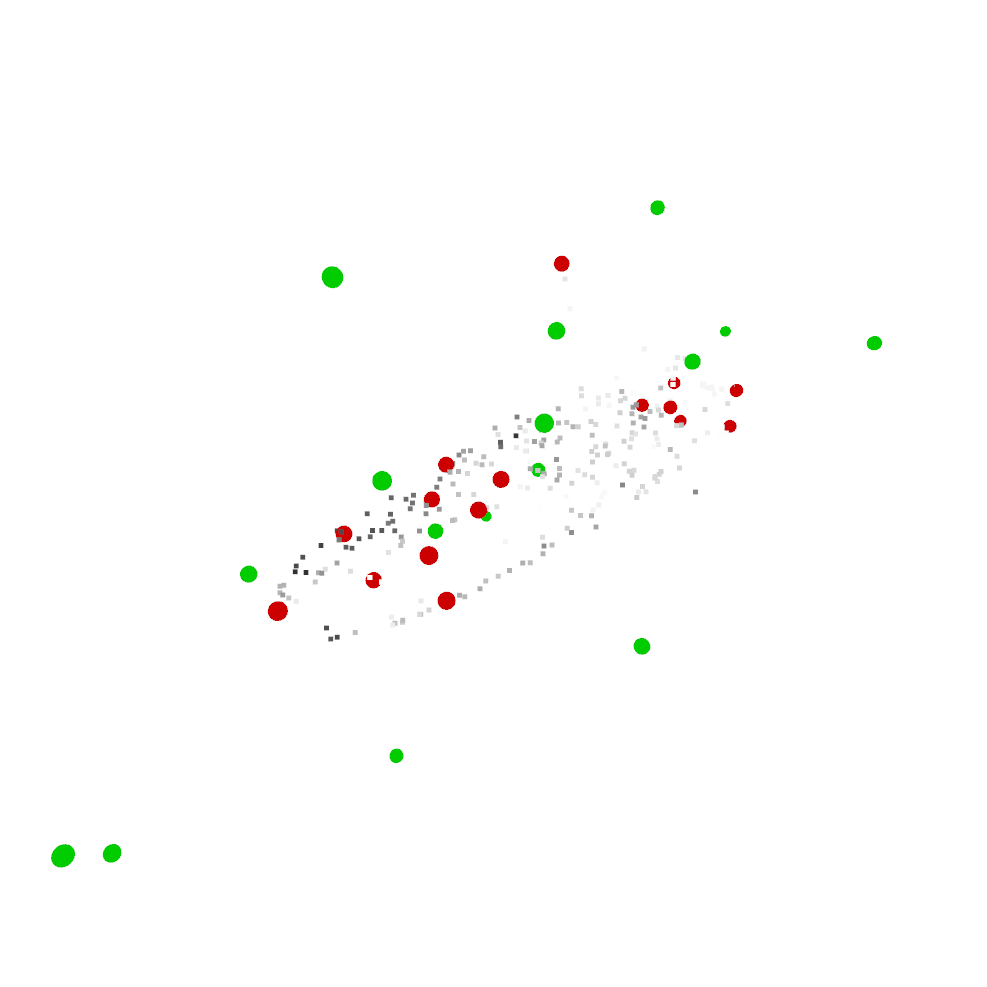} 
		\caption{Perturbed keypoints (green) and true keypoints (red) at $\sigma=0.8$.}
		\label{fig:vessel-fig}
	\end{subfigure}
	
	\caption{ShapeNet objects: bottle (\ref{fig:bottle}-\ref{fig:bottle-fig}), skateboard (\ref{fig:skateboard}-\ref{fig:skateboard-fig}), table (\ref{fig:table}-\ref{fig:table-fig}), vessel (\ref{fig:vessel}-\ref{fig:vessel-fig}).}
	\label{fig:corrector-analysis-extra-04}
\end{figure*}

 }

\begin{IEEEbiography}[{\includegraphics[width=1in,height=1.25in,clip,keepaspectratio]{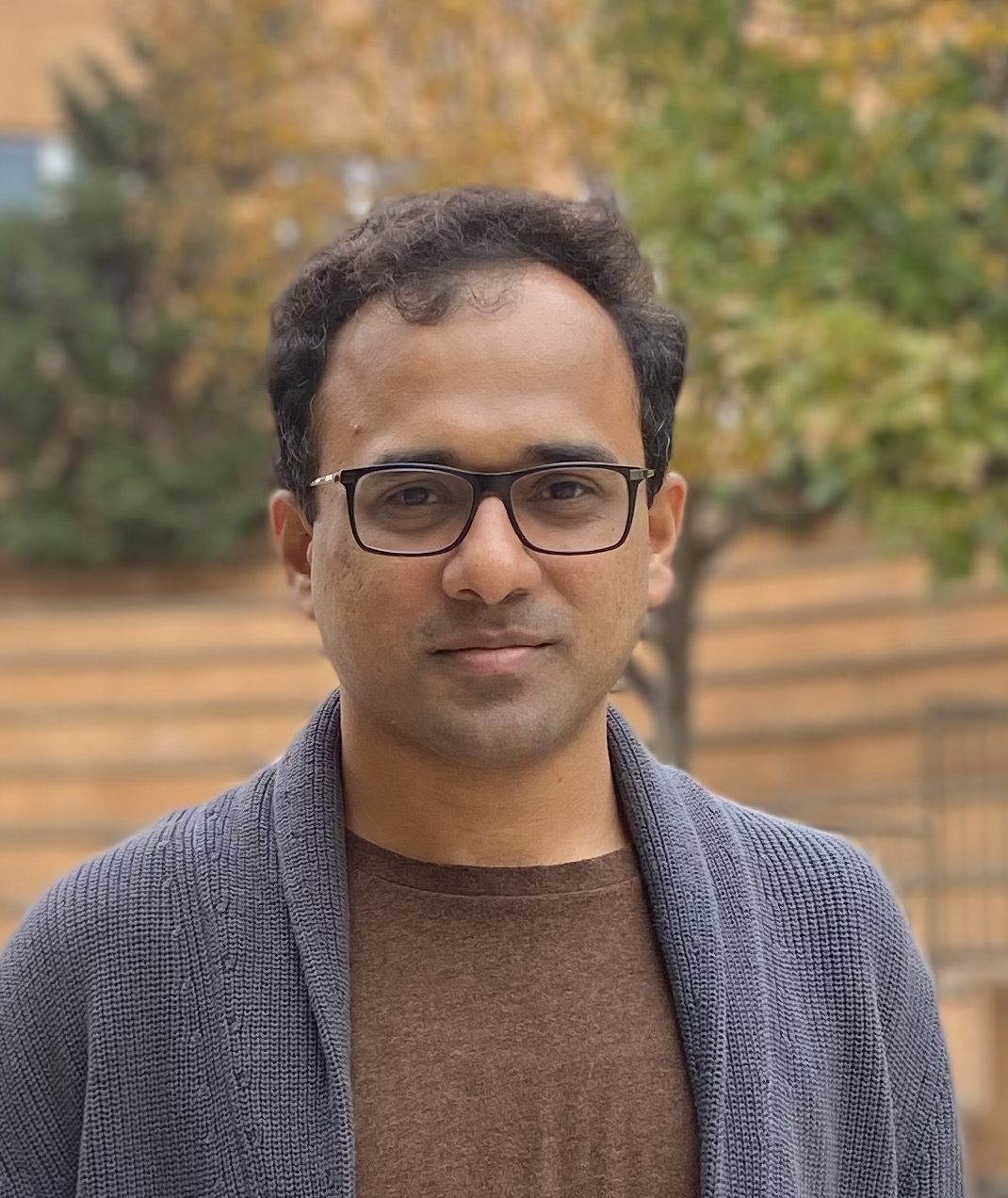}}]{Rajat Talak} is a Research Scientist in the Department of Aeronautics and Astronautics at the Massachusetts Institute of Technology (MIT). Prior to this, he was a Postdoctoral Associate in the Department of Aeronautics and Astronautics, at MIT, from 2020-2022. He received his Ph.D. from the Laboratory of Information and Decision Systems at MIT, in 2020. He holds a Master of Science degree from the Dept. of Electrical Communication Engineering, Indian Institute of Science, Bangalore, India, and a B.Tech from the Dept. of Electronics and Communication Engineering, National Institute of Technology - Surathkal, India.
His research interests are in the field of safe robot perception, 3D scene understanding, and networked autonomy. 
He is the recipient of ACM MobiHoc 2018 best paper award and gold medal for his MSc thesis. 
\end{IEEEbiography}
\begin{IEEEbiography}[{\includegraphics[width=1in,height=1.25in,clip,keepaspectratio]{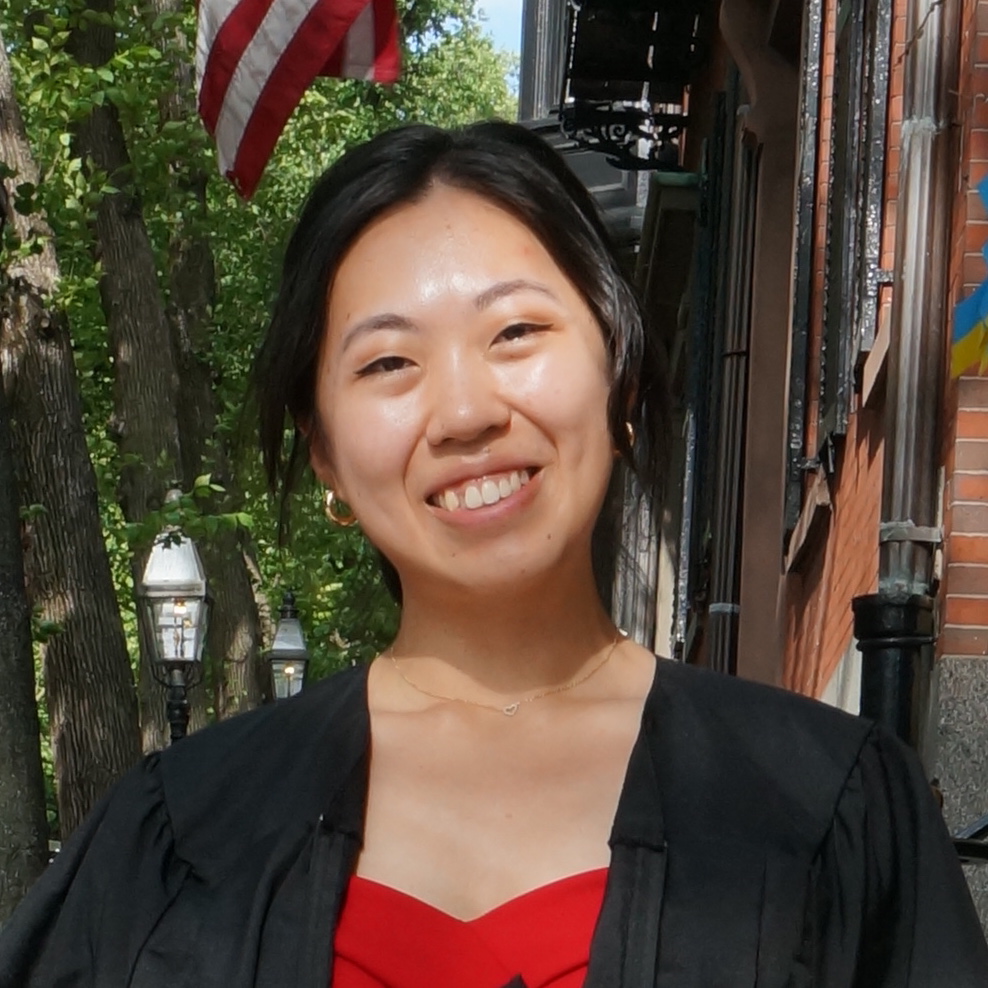}}]{Lisa R. Peng} received the B.S. and MEng degrees in electrical engineering and computer science from the Massachusetts Institute of Technology, in 2021 and 2022, respectively. Her research with SPARK (sensing, perception, autonomy, and robot kinetics) lab include self-supervised training and certification for 3D object pose estimation with applications in 3D scene graph construction and reinforcement learning.
	She is currently the Computer Vision Software Engineer with Ample, an EV battery swapping company based in San Francisco, California. She works on a variety of 2D and 3D perception problems for autonomous robotics.
\end{IEEEbiography} %
\begin{IEEEbiography}[{\includegraphics[width=1in,height=1.25in,clip,keepaspectratio]
		{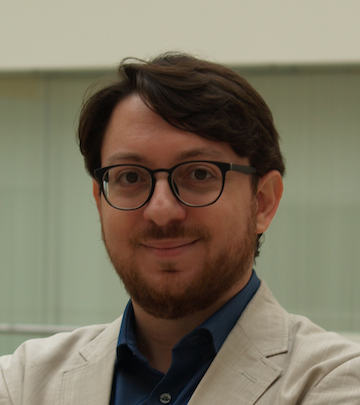}}]{Luca Carlone}
	is the Leonardo Career Development Associate Professor in the Department of Aeronautics and Astronautics at the Massachusetts Institute of Technology, and a Principal Investigator in the Laboratory for Information \& Decision Systems (LIDS). He has obtained a B.S. degree in mechatronics from the Polytechnic University of Turin, Italy, in 2006; an S.M. degree in mechatronics from the Polytechnic University of Turin, Italy, in 2008; an S.M. degree in automation engineering from the Polytechnic University of Milan, Italy, in 2008; and a Ph.D. degree in robotics also from the Polytechnic University of Turin in 2012. He joined LIDS as a postdoctoral associate (2015) and later as a Research Scientist (2016), after spending two years as a postdoctoral fellow at the Georgia Institute of Technology (2013-2015). His research interests include nonlinear estimation, numerical and distributed optimization, and probabilistic inference, applied to sensing, perception, and decision-making in single and multi-robot systems. His work includes seminal results on certifiably correct algorithms for localization and mapping, as well as approaches for visual-inertial navigation and distributed mapping. He is a recipient of the Best Student Paper Award at IROS 2021, the Best Paper Award in Robot Vision at ICRA 2020, a 2020 Honorable Mention from the IEEE Robotics and Automation Letters, a Track Best Paper award at the 2021 IEEE Aerospace Conference, the 2017 and 2022 Transactions on Robotics King-Sun Fu Memorial Best Paper Award, the Best Paper Award at WAFR 2016, the Best Student Paper Award at the 2018 Symposium on VLSI Circuits, and he was best paper finalist at RSS 2015, RSS 2021, and WACV 2023.   He is also a recipient of the AIAA Aeronautics and Astronautics Advising Award (2022), the NSF CAREER Award (2021), the RSS Early Career Award (2020), the Google Daydream Award (2019), the Amazon Research Award (2020, 2022), and the MIT AeroAstro Vickie Kerrebrock Faculty Award (2020). He is an IEEE senior member and an AIAA associate fellow. 
\end{IEEEbiography} 

\end{document}